\newif\ifreview
\reviewfalse
\ifreview
\documentclass[review,1p,times]{elsarticle}
\else
\documentclass[final,3p,times,twocolumn]{elsarticle}
\fi

\usepackage{amssymb}
\usepackage{amsmath}
\usepackage{multirow}
\usepackage{lineno}
\newcommand{\R}{\mathbb{R}}
\usepackage{graphicx}
\usepackage{booktabs}
\usepackage{xcolor}
\usepackage{subcaption}
\usepackage{graphicx}
\usepackage{tikz}
\usepackage{pgfplots}
\usepackage[percent]{overpic}
\pgfplotsset{compat=1.18}

\usepackage{algorithm}
\usepackage{algpseudocode}

\newif\ifshowchanges
\showchangesfalse  % Show red text % Set to \showchangesfalse to hide red text

\usepackage{hyperref}
\usepackage{xurl}
\hypersetup{
  breaklinks=true,
  colorlinks=true,
  urlcolor=Blue,
  linkcolor=Blue,
  citecolor=Blue
}

\DeclareMathOperator{\PoD}{PoD}

\DeclareMathOperator{\KL}{KL}
\newcommand*{\CoV}{\mathrm{CoV}}

\begin{document}
\begin{frontmatter}

%% Title, authors and addresses

%% use the tnoteref command within \title for footnotes;
%% use the tnotetext command for theassociated footnote;
%% use the fnref command within \author or \address for footnotes;
%% use the fntext command for theassociated footnote;
%% use the corref command within \author for corresponding author footnotes;
%% use the cortext command for theassociated footnote;
%% use the ead command for the email address,
%% and the form \ead[url] for the home page:
%% \title{Title\tnoteref{label1}}
%% \tnotetext[label1]{}
%% \author{Name\corref{cor1}\fnref{label2}}
%% \ead{email address}
%% \ead[url]{home page}
%% \fntext[label2]{}
%% \cortext[cor1]{}
%% \affiliation{organization={},
%%             addressline={},
%%             city={},
%%             postcode={},
%%             state={},
%%             country={}}
%% \fntext[label3]{}

% \title{A Statistical Approach for Evaluation of Automated Visual Crack Detection Using Probability of Detection}
\title{Evaluation of AI-based Visual Crack Detection in Steel Bridges Using Probability of Detection}

%% use optional labels to link authors explicitly to addresses:
\author[label1,label2]{Andrii Kompanets\corref{cor1}}
\author[label2,label3]{Finn Sherry\corref{cor1}}
\author[label2,label3]{Remco Duits}
\author[label1,label2]{Davide Leonetti}
\author[label1]{H.H. (Bert) Snijder}

\cortext[cor1]{Corresponding authors. Email: a.kompanets@tue.nl, f.m.sherry@tue.nl}

\affiliation[label1]{organization={Eindhoven University of Technology, Department of the Built Environment},
            % addressline={},
            city={Eindhoven},
            % postcode={5600 MB},
            % state={},
            country={The Netherlands}}

\affiliation[label2]{organization={Eindhoven Artificial Intelligence Systems Institute, Eindhoven University of Technology},
            % addressline={},
            city={Eindhoven},
            % postcode={},
            % state={},
            country={The Netherlands}}

\affiliation[label3]{organization={Eindhoven University of Technology, Department of Mathematics and Computer Science},
            % addressline={},
            city={Eindhoven},
            % postcode={},
            % state={},
            country={The Netherlands}}

% \author{}

% \affiliation{organization={},%Department and Organization
%             addressline={}, 
%             city={},
%             postcode={}, 
%             state={},
%             country={}}

\begin{abstract}
Bridge structures are regularly inspected for structural damage such as cracks and corrosion in order to ensure public safety and reduce maintenance costs.
Much research has been done on automating this process using computer vision methods, which are often evaluated and compared using metrics such as intersection over union, mean average precision, etc. However, predicting the actual effectiveness of an inspection method within the field of structural engineering from these metrics remains challenging. To enable the systematic use of these increasingly popular methods in engineering practice, evaluating the performance of these methods in a way that is compatible with standard engineering approaches is therefore an urgent necessity.
We present a new statistical evaluation framework to allow the comparison of computer vision methods with conventional visual inspection for crack detection in steel bridges.
The framework is based on probability of detection curves and can account for the influence of image resolution.
We apply this evaluation method to the real-world ``Cracks in Steel Bridges'' dataset \cite{CSB_dataset}, which contains annotated images of cracks in bridge structures.
The quantification of the probability of detection and its uncertainty enables a practical assessment of the effect of automated methods for damage detection in structural reliability analyses. 
In turn, this enables the wide-spread use of automated (AI-based) damage detection in safety critical applications. 
This evaluation method provides evidence that the proposed computer vision approach approach is robust for the crack detection task and can have a high added value as an addition to conventional visual inspection methods.
\end{abstract}

% %%Graphical abstract
% \begin{graphicalabstract}
% %\includegraphics{grabs}
% \end{graphicalabstract}

%%Research highlights
% \begin{highlights}
% \item Research highlight 1
% \item Research highlight 2
% \end{highlights}

\begin{keyword}
%% keywords here, in the form: keyword \sep keyword
crack detection \sep steel bridges \sep nondestructive evaluation \sep probability of detection
%% PACS codes here, in the form: \PACS code \sep code

%% MSC codes here, in the form: \MSC code \sep code
%% or \MSC[2008] code \sep code (2000 is the default)

\end{keyword}

\end{frontmatter}
\ifreview
\linenumbers
\fi

\section{Introduction}\label{sec:Introduction}

In Europe, bridges have aged considerably \cite{brady2009cost345}, and a large proportion of bridges are older than 50 years \cite{bridges2004d1}. 
For example about 80~\!\% of steel bridges and 20~\!\% of concrete bridges in the Netherlands
have less than a third of their design life remaining \cite{Rijkswaterstaat2022ReportCondition}.
This poses an issue because, with time, the bridge structures tend to deteriorate due to various factors such as fatigue and corrosion. Prolonged exploitation of the bridge increases the chances of its structural failure or, in the worst case, complete collapse of the bridge. 
This explains the recent surge in research activities aimed at improving the monitoring and inspection of the structural health of bridges \cite{shahrivar2025ai_shm,isac2025SHM_trends}. 
Special attention has been paid to the improvement of deep learning-based methods for visual detection of fatigue cracks, as has
been reviewed in \cite{ali2022AI_inspection_trends}, as the visual inspection is the simplest and most widely used method and fatigue cracks are the most threatening type of damage to the structural integrity of the bridge. 

Conventional visual inspection is carried out by trained inspectors who visually examine the bridge structure. 
In some cases, special equipment, such as climbing ropes, elevator machines, magnifying lenses, and spotlights, is necessary to allow the inspector to get close to the bridge and to detect damage, such as cracks or corrosion, more accurately.
Automated methods for visual bridge inspection have been proposed to improve and support conventional inspection methods, by increasing their speed, reducing their cost, and improving the consistency of these methods by minimising the chance of human error. 
These emerging methods usually involve a robotic system, e.g.\ an unmanned aerial vehicle or a crawling robot collecting images of the bridge structure. Those pictures are subsequently analysed by a Computer Vision (CV) algorithm. 

Several reviews of articles studying automated visual inspections of damage in bridges have been published \cite{amirkhani2024ReviewDLInspect,panigati2025ReviewDronesInspect}.
From this we can see the general progress over time of CV methods for crack detection.
Progress has been made by (1) using more advanced general-purpose CV algorithms, (2) tailoring those algorithms for specific tasks, e.g.\ crack detection in steel bridges, concrete, or asphalt (3) improving the quality and size of the datasets used to train the CV algorithms, and (4) improving the computing capabilities of the hardware used for the tasks. 
In order to deploy these methods in real-world scenarios, e.g.\ in supporting engineers during routine inspections of structures, however, their maturity must be properly evaluated.

In the race to develop better CV methods, they are often compared with each other using benchmark datasets.
However, only a few studies have compared computer vision methods for automatic crack detection with conventional inspection methods, such as visual inspections performed by a trained inspector. 
Examples of such studies include \cite{wojcikAIvsMAN12020}, which applied the DeepCrack algorithm \cite{liu2019deepcrack} to images of concrete bridges manually annotated by human inspectors. They found that while the accuracy, a standard metric used in the field of CV, was high ($\sim 90~\%$), the real ability of the algorithm to accurately detect cracks was poor, since it incorrectly classified most surface irregularities as cracks. 

In another case study reported in \cite{KimAIvsMAN12022}, a few concrete bridge piers were physically inspected by a human inspector and by an unmanned aerial vehicle with an automated crack detection system using CV methods. 
Diametrically opposite to the previous study, here the CV algorithm outperformed the human inspector and was able to detect cracks that were missed by the human. The use of a different CV algorithm may partially explain the divergence in the conclusions of \cite{wojcikAIvsMAN12020} and \cite{KimAIvsMAN12022}. 

Note that automated visual crack detection should not only be considered as a replacement for conventional inspection. 
In \cite{kim2025exploratory}, the authors considered CV crack detection as a way to augment the performance of a human inspector, rather than to replace the human inspector. 
In this case study, an inspector was able to detect 90~\% of cracks when assisted by CV and only 60~\% without CV assistance.

To the best of our knowledge, up to now visual inspection and CV inspection have been compared only in case studies. 
Despite their relevance, case studies tend to be extremely time-consuming and they generalise poorly, as a consequence of small sample sizes and a high susceptibility to researcher bias.
The fact that there are so few publications that compare CV methods to conventional visual inspection and that all existing results come from case studies represents a major gap in the field. 

One of the main obstacles standing in the way of such comparisons is the fact that the metrics for evaluating conventional visual inspection are different from those typically used to evaluate CV algorithms. 
Conventional crack detection method performance is commonly quantified using a probability of detection (PoD) curve, which relates the probability of detecting a crack to its characteristic size \cite{ENIQ_POD_BestPractices_2011,MILHDBK1823A,DNVGL_RP_C210_2015}.
PoD curves are used within a structural reliability assessment to quantify the beneficial effect of an inspection in terms of structural reliability, which is related to the probability of failure \cite{Leonetti2021InfluenceModels,Maljaars2014ProbabilisticLocations,Hashemi2019ValueBridges}.
Without such a curve, it can be particularly difficult to judge the true performance of a CV method for inspection, compare it with manual visual inspection, determine whether it can be used in real applications, and consider it quantitatively in a structural reliability assessment.

Generally, the performance of CV methods for damage detection depends to a large degree on different inspection conditions, such as camera sensor and optics, distance from the camera to the damage, and other variables that affect the visibility of damage in the images, such as lighting conditions and incidence. 
To the best of our knowledge, there is no existing general methodology to model the dependence of the performance of CV methods for automated crack detection on these highly relevant additional variables. 

\subsection{Contributions}\label{sec:Introduction:contributions}
Given the aforementioned context, we identify the following three open questions in the field of automated visual inspection:
\begin{itemize}
\item How can an automated CV crack detection method be appropriately compared to conventional crack detection methods?
\item Are CV methods for crack detection robust enough for deployment in real bridge inspection applications?
\item How do camera specifications and the distance from the camera to the crack affect the performance of CV methods for automated crack detection?
\end{itemize}
The main contribution of this paper is to answer the aforementioned questions using the example of our CV method for crack detection in steel bridges, as follows:
\begin{itemize}
\item We introduce a framework for comparing the crack detection performance of CV methods to conventional inspection approaches, described in Sec. \ref{sec:Effect on crack length update after inspection} and Algorithm (Alg.)~\ref{alg:missed_crack_distribution}. For this, we first adapt the use of PoD curves from conventional approaches to automated visual methods for crack detection in Sec.~\ref{sec: Probability of detection}.
\item We show, using our newly proposed comparison framework (Alg.~\ref{alg:missed_crack_distribution}), that our CV method can outperform existing estimates (Campbell et al.~\cite{Campbell2019pod}, DNVGL recommendation \cite{DNVGL_RP_C210_2015}) of conventional inspection performance for cracks of length less than 150 mm. For longer cracks, our CV method still outperforms the DNVGL recommendation \cite{DNVGL_RP_C210_2015}.
\item We analyse how the performance of our CV method is affected by the camera specifications and distance between camera and the crack. 
For this analysis, we use both a nonparametric (Sec.~\ref{sec:ResolutionEffect of Resolution on PoD:Resolution effect with binning}) and a parametric (Sec.~\ref{sec:Effect of Resolution on PoD:pod surface}) method. 
In both cases we observe that the PoD increases as the resolution improves.   
We show that an automated inspection, applying our CV method to images collected by a camera with instantaneous field of view (see Section \ref{sec:Effect of Resolution on PoD}) equal to $1.25 \times 10^{-4}$ rad/pix from a distance of up to 3 metres, has a higher PoD than conventional visual inspection and has comparable performance for longer cracks. 
\end{itemize}

\subsection{Structure of the article}
\label{sec:Introduction:Structure of the Article}
The paper is structured as follows.
Section~\ref{sec:Data} describes the data that was used in this study.
In Section~\ref{sec:Deep learning method for crack detection}, we present a CV method, based on a neural network, for the detection of cracks in images of steel bridges. 
Further, in Section~\ref{sec: Probability of detection} we propose a methodology to estimate the PoD curve of an automated visual crack detection approach. The PoD curve estimated for our CV method is then quantitatively compared with the conventional inspection method in Section~\ref{sec:Effect on crack length update after inspection}. In Section~\ref{sec:Effect of Resolution on PoD}, we propose two methods to explicitly take into account the effect of resolution on the probability of detection.
Finally, Section~\ref{sec:Conclusions} concludes the work.

\section{Data}\label{sec:Data}
\begin{figure*}
\centering
\begin{subfigure}[b]{0.33\textwidth}
\includegraphics[width=1\textwidth]{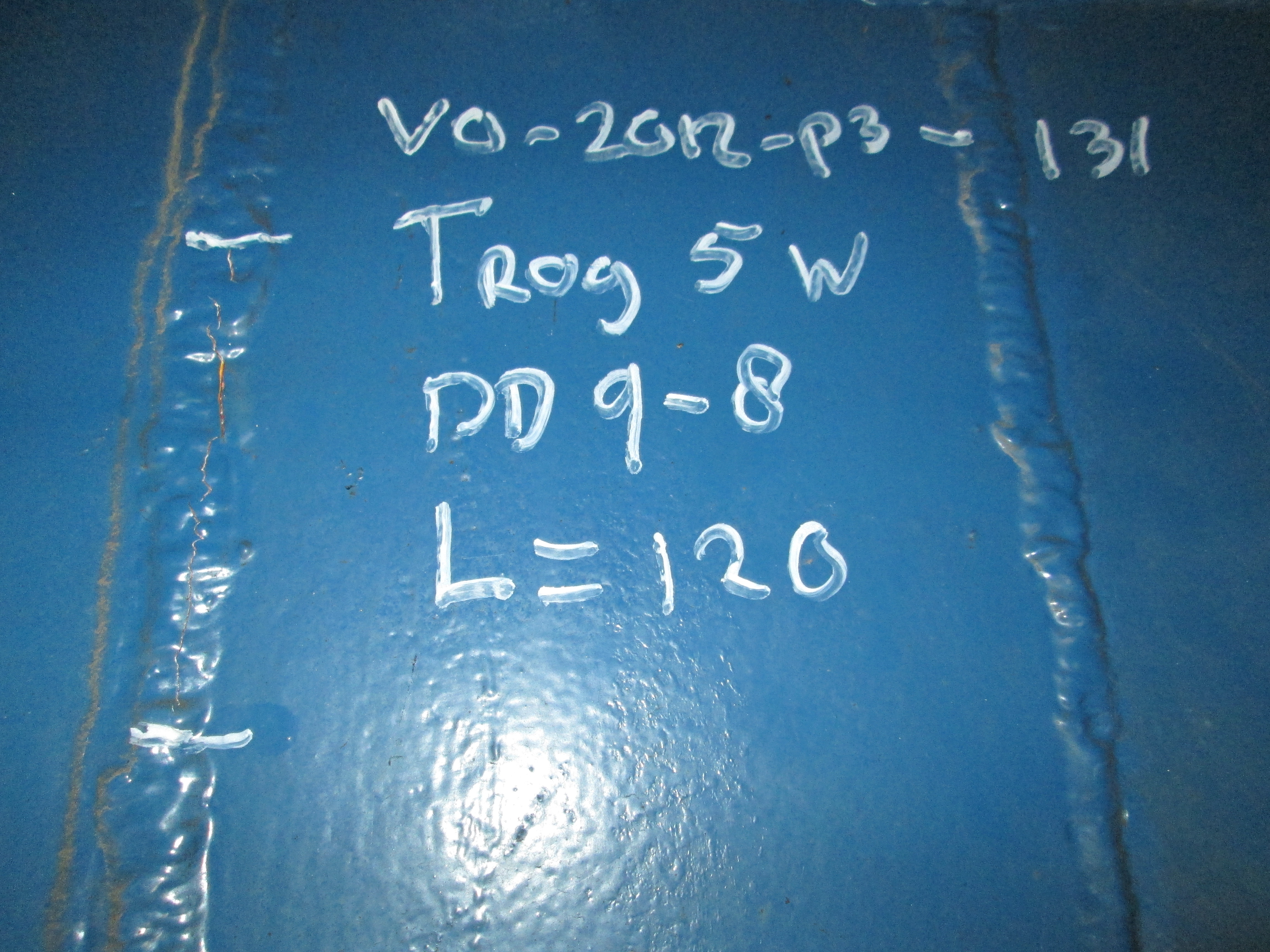}
\caption{}
\label{fig: Dataset examples/PoD}
\end{subfigure}\hfill
\begin{subfigure}[b]{0.33\textwidth}
\includegraphics[width=1\textwidth]{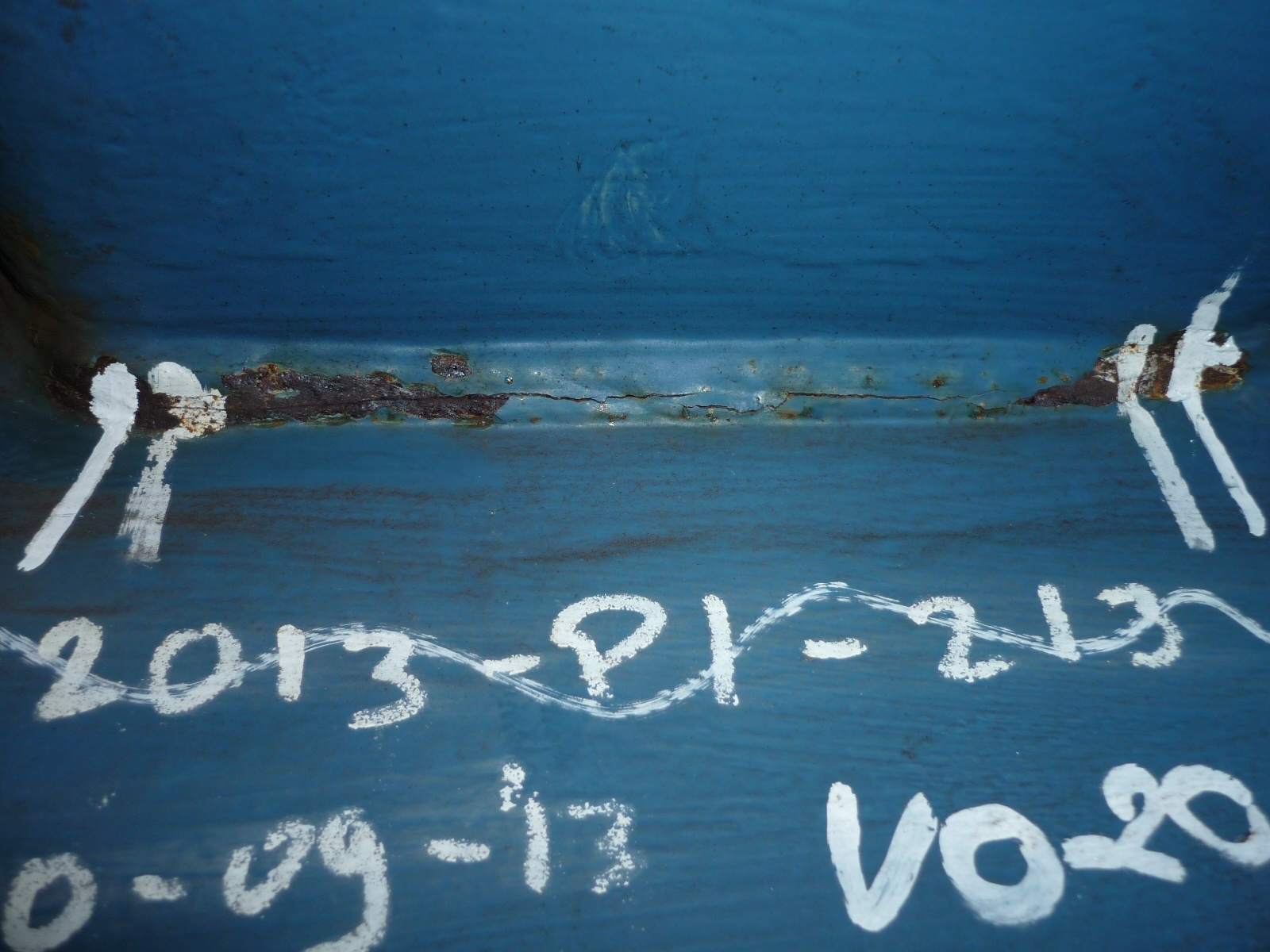}
\caption{}
\label{fig: Dataset examples/No PoD}
\end{subfigure}\hfill
\begin{subfigure}[b]{0.33\textwidth}
\includegraphics[width=1\textwidth]{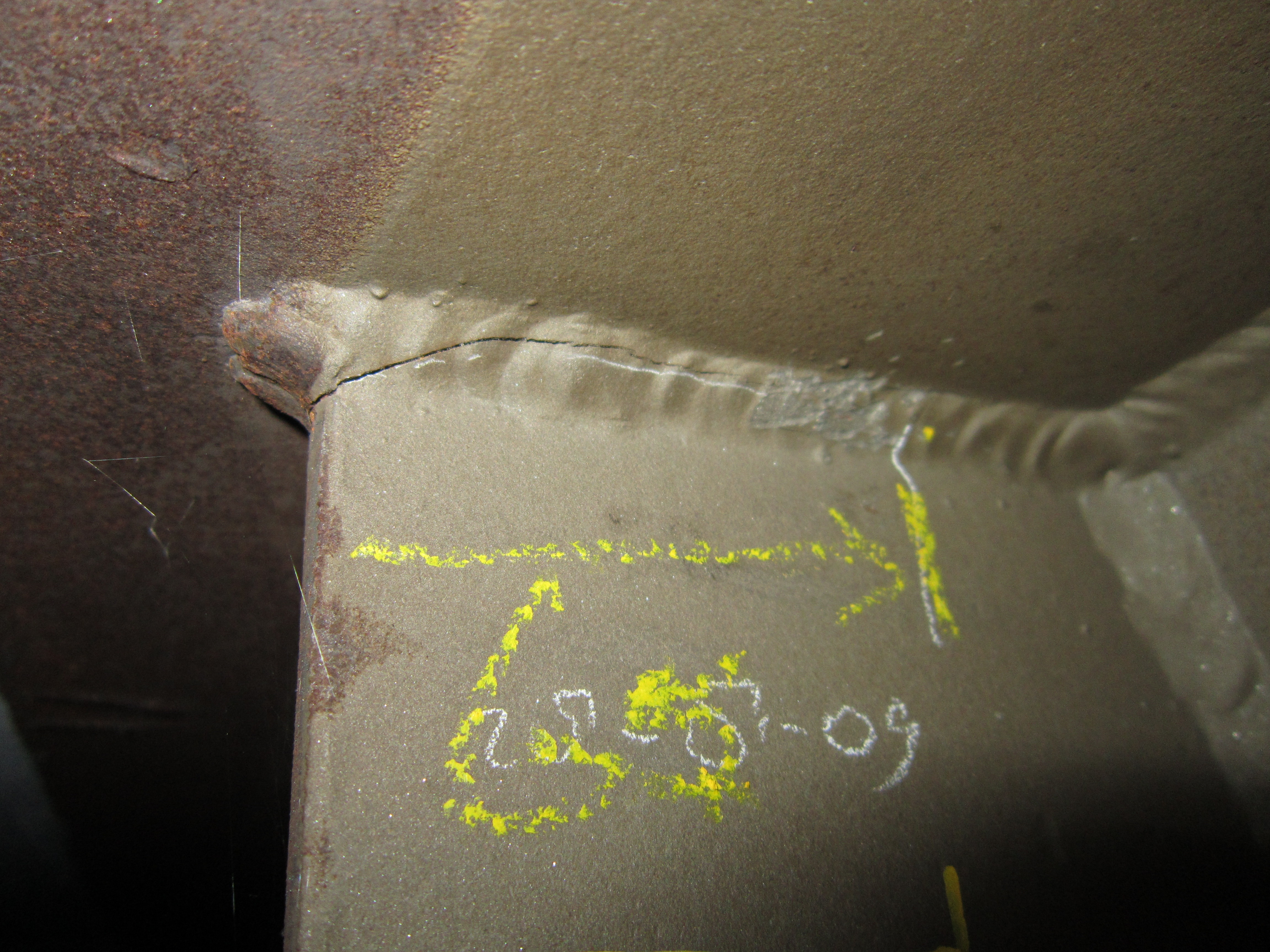}
\caption{}
\end{subfigure}\hfill
\caption{Examples of images used in this study, taken from the ``Cracks in Steel Bridges'' dataset \cite{CSB_dataset}.}
\label{fig: Crack/Dataset examples}
\end{figure*}

The analyses carried out in this paper are performed on images contained in the ``Cracks in Steel Bridges'' (CSB) \cite{CSB_dataset} dataset, which has also been used in previous works \cite{JP1, JP2}. 
The dataset consists of 6163 images collected over the years during routine conventional bridge inspections. Some examples are shown in Fig.~\ref{fig: Crack/Dataset examples}. The images have sizes ranging from a minimum of 600 × 450 pixels to a maximum of 4608 × 3456 pixels.
The images capture various structures of different bridges with structural damage, such as fatigue cracks and/or corrosion damage. 
As the images were taken while the bridge structures were in use, often dust, dirt, oil or water stains, and even insects are visible. In some images, artificial marks made by inspectors are present, indicating i.a.\ the location of the crack tips on the surface and the crack length. Bridge structure surfaces are painted mainly in blue, brown, white, or black. In some cases, the paint is intentionally removed at the tips of the cracks by the inspector during routine inspections to reveal the true length of the crack.
The images were captured at different angles relative to the captured structure surfaces, with different lighting conditions, and at different distances ranging from 0.5 to 5 metres. 
We have used these images first to train a CV method for automatic crack detection and then to assess the performance of this method using PoD estimation.

\subsection{PoD estimation dataset}\label{sec:Data:Data for PoD estimation}
To determine the PoD curve of an inspection method, one needs to use a dataset containing cracks of known lengths.
This information was not recorded for the available images. However, in some of the images, inspection markings made on the bridge surface indicating the length of the crack are visible. An example of such an image is shown in Fig.~\ref{fig: Dataset examples/PoD}. The dataset used for the PoD estimation consists of 239 manually selected images in which the crack length is identifiable from these markings. Notably, not all images with markings contain the crack length, cf.~Fig.~\ref{fig: Dataset examples/No PoD}.
Such images were not included in the PoD estimation dataset. 

\begin{figure}
    \centering
    \begin{subfigure}[c]{0.49\textwidth}
        \includegraphics[width=\textwidth]{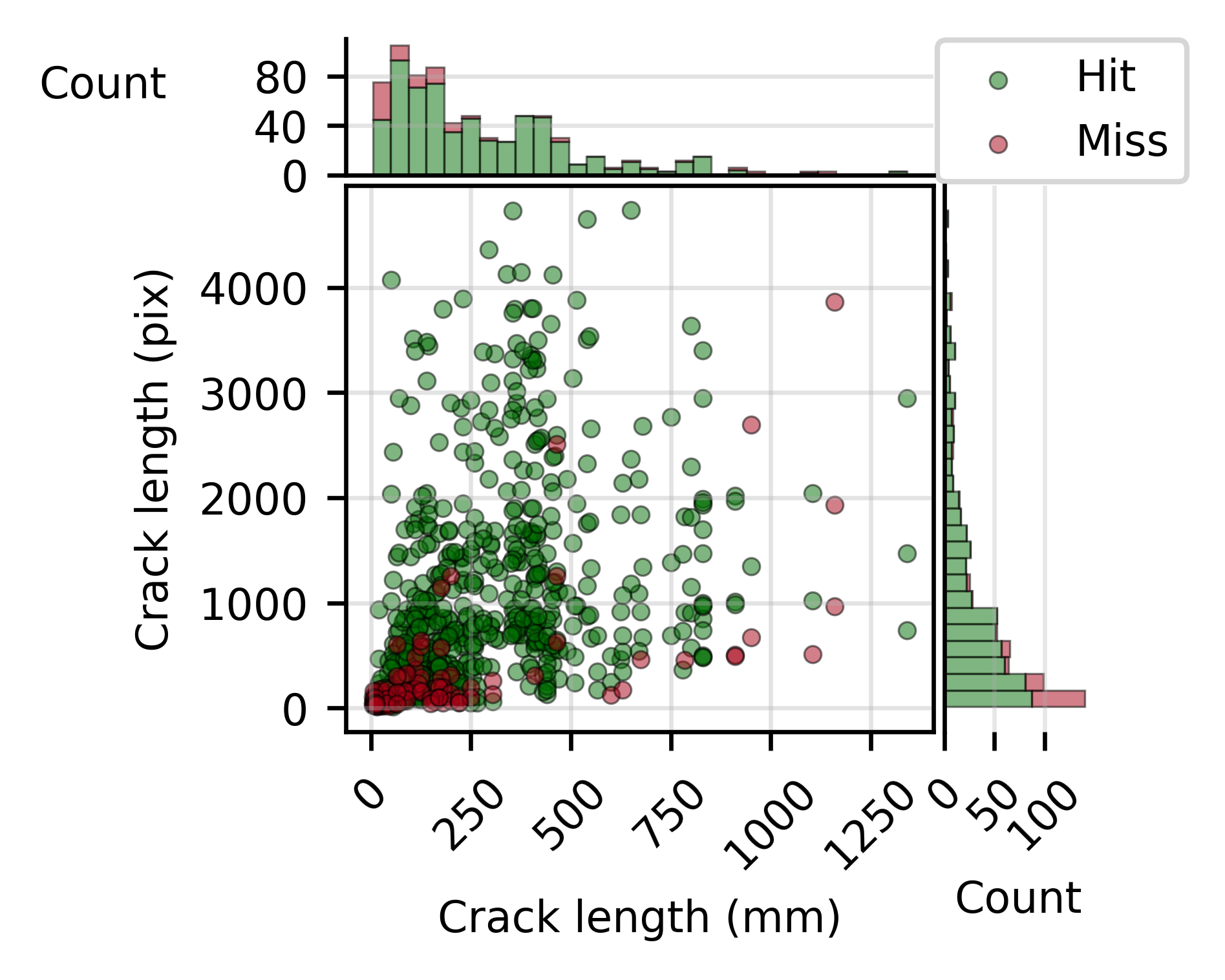}
        \caption{}
        \label{fig:Data statistics:mm vs pix}
    \end{subfigure}
    \hfill
    \begin{subfigure}[c]{0.49\textwidth}
        \includegraphics[width=\textwidth]{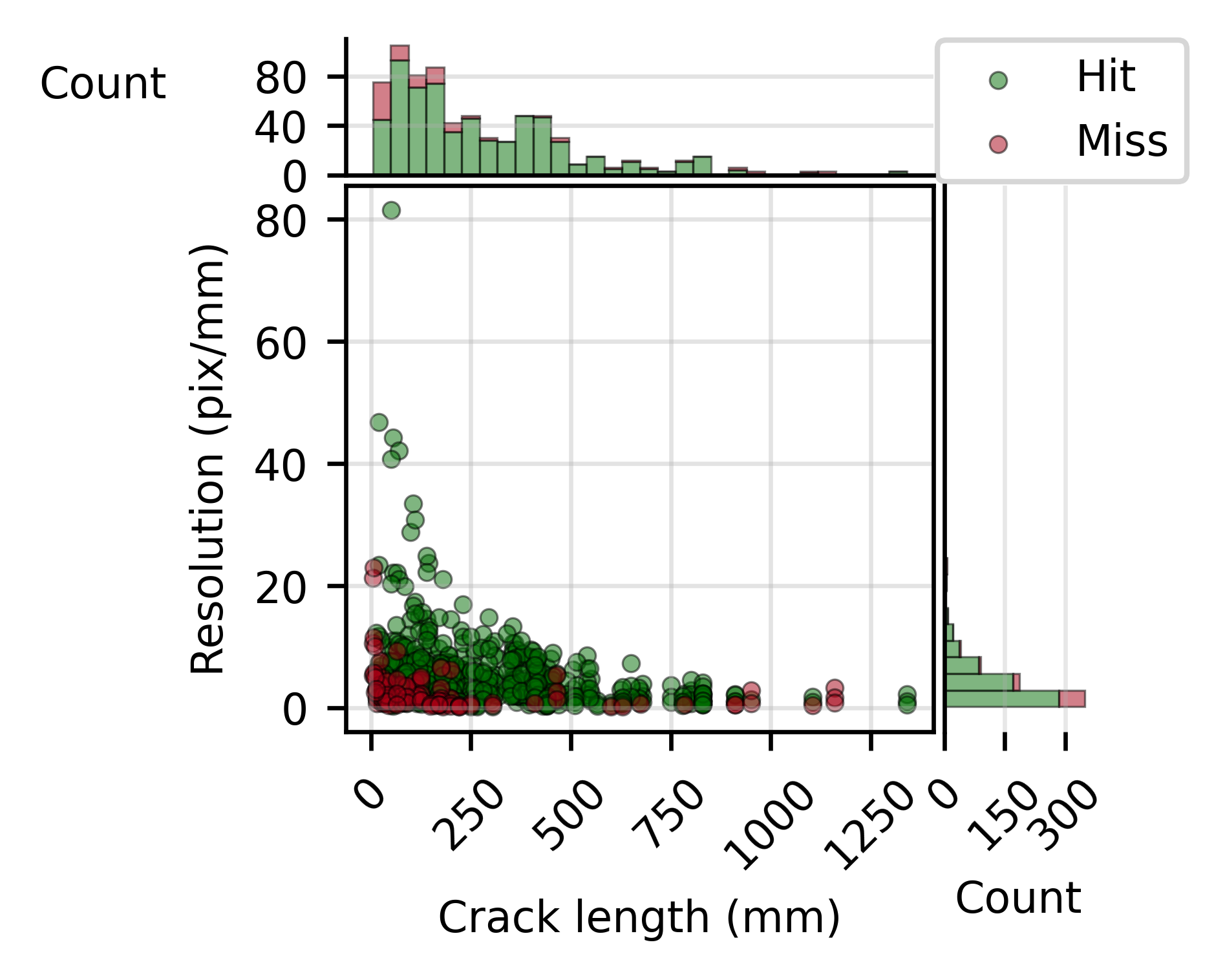}
        \caption{}
        \label{fig:Data statistics:pix/mm vs mm}
    \end{subfigure}
    \caption{Distribution of physical (mm) and digital (pix) crack lengths, and resolution over the dataset used for PoD estimation.}
    \label{fig:PoD data statistics}
\end{figure}

For each image in the PoD estimation dataset, the digital crack length in pixels is additionally recorded. The digital crack length is determined by manually selecting points on the crack and connecting these by straight lines. For straight cracks, only the end points are used; for particularly curvy cracks, an additional point is added at the bend. The length of these lines in pixels is then computed using the Pythagorean theorem. 
Fig.~\ref{fig:Data statistics:mm vs pix} shows the digital crack length in pixels against the physical crack length in millimetres for each image; by dividing the digital length by the physical length, we obtain the resolution in pixels per millimetre, which is plotted against the physical crack length in Fig.~\ref{fig:Data statistics:pix/mm vs mm}. In both figures, the marks correspond to images in which the crack was detected (hit, green) and missed (miss, red) by the crack detection CV method (see Sec.~\ref{sec:Deep learning method for crack detection}). In fact, the resolution is typically not uniform along the length of the crack if the camera is not facing straight onto the surface containing the crack.
Therefore, we work with an approximate average resolution over a single crack.

\subsection{Dataset for neural network training}\label{sec:Data:Data for neural network training}
To train our CV method for crack detection, we follow the approach in \cite{JP2}. 
The dataset described above was first manually annotated with bounding boxes to indicate cracks, and then split into train, validation, and test sets with 80/10/10~\% of the images, respectively. 
All the images used for PoD estimation were assigned to the test set; the other images were randomly assigned to the train, validation, and test sets. This ensures that the PoD estimation is performed on images that were not seen by the neural network during training, thereby reducing the risk that overfitting affects the results.

\section{CV method for crack detection}\label{sec:Deep learning method for crack detection}
As a CV method for crack detection in images, we used the Faster-RCNN neural network \cite{Ren2017FasterRCNN}, implemented in OpenMMLAB \cite{mmdetection}, which was also applied in \cite{JP2}. We now give a brief overview of the method's architecture.

The Faster-RCNN takes images as input and outputs a set of bounding boxes, each indicating a crack with a certain confidence level. 
The images are first resampled by the model using a bicubic interpolation method, so that their biggest size equals 1500 pixels. 
Subsequently, these images are encoded in feature maps with lower spatial dimensionality by the model's backbone, which is based on a ResNext101 neural network \cite{xie2017ResNext} pretrained on the ImageNet dataset \cite{Deng2009ImageNet}. 
A set of regions is then generated. These regions have more extreme aspect ratios (0.33; 1; 3 compared to 0.5; 1; 2) and scales compared to the original Faster-RCNN \cite{Ren2017FasterRCNN}, in order to account for the geometry of the cracks.
Using the feature maps, the network assigns each region a score indicating how confident it is that the region contains a crack.
Finally, the network removes unlikely and overlapping proposals, and refines the remaining regions and their confidence scores.

The model was trained using an AdamW optimiser \cite{Loshchilov2017AdamW} for 40 epochs with an initial learning rate of $8 \times 10^{-5}$ and a cosine-annealing learning rate schedule \cite{loshchilov2016CosineAnnealing} starting at epoch 20. In addition, the first 500 training iterations were used for the learning rate warm-up. A weight decay of $1 \times 10^{-4}$ was applied for L2 regularisation. Data augmentation was performed during training by randomly flipping the images and the corresponding segmentations. For all other hyperparameters, we use the default values of the OpenMMLab implementation \cite{mmdetection}. The neural network training was conducted using a single NVIDIA A100 40GB GPU.

\begin{figure*}
    \centering
    \begin{subfigure}[c]{0.32\textwidth}
        \includegraphics[width=\textwidth]{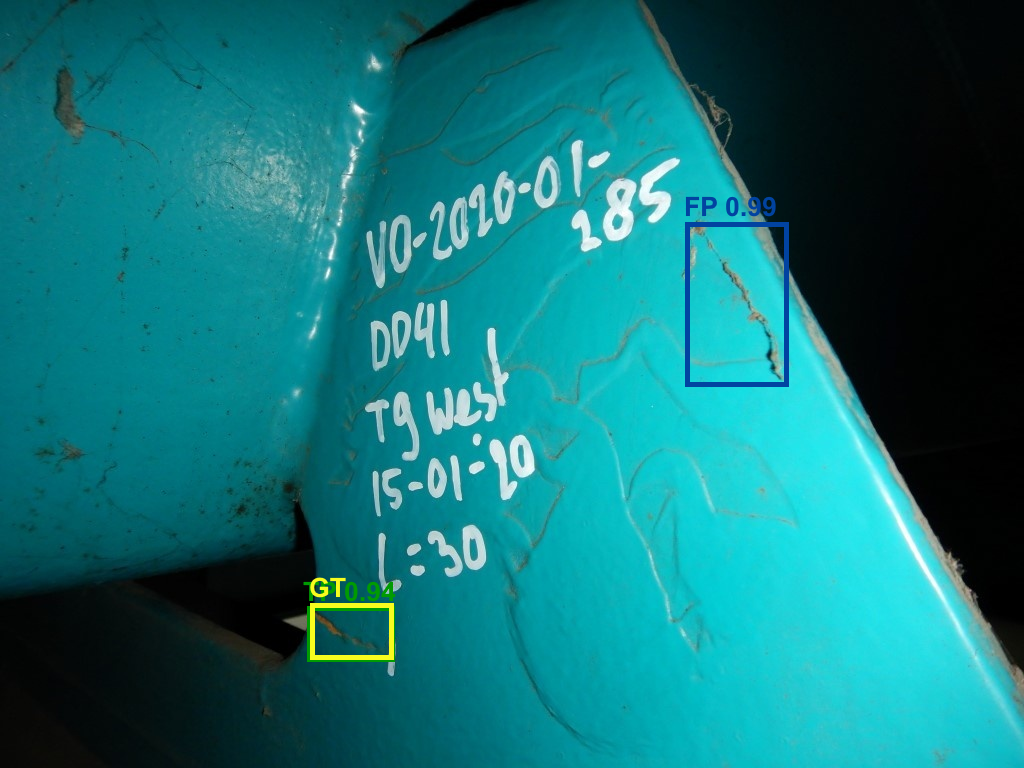}
        \caption{}
        \label{fig:NN_Outputs 1}
    \end{subfigure}
    \hspace{0.0em}
    \begin{subfigure}[c]{0.32\textwidth}
        \includegraphics[width=\textwidth]{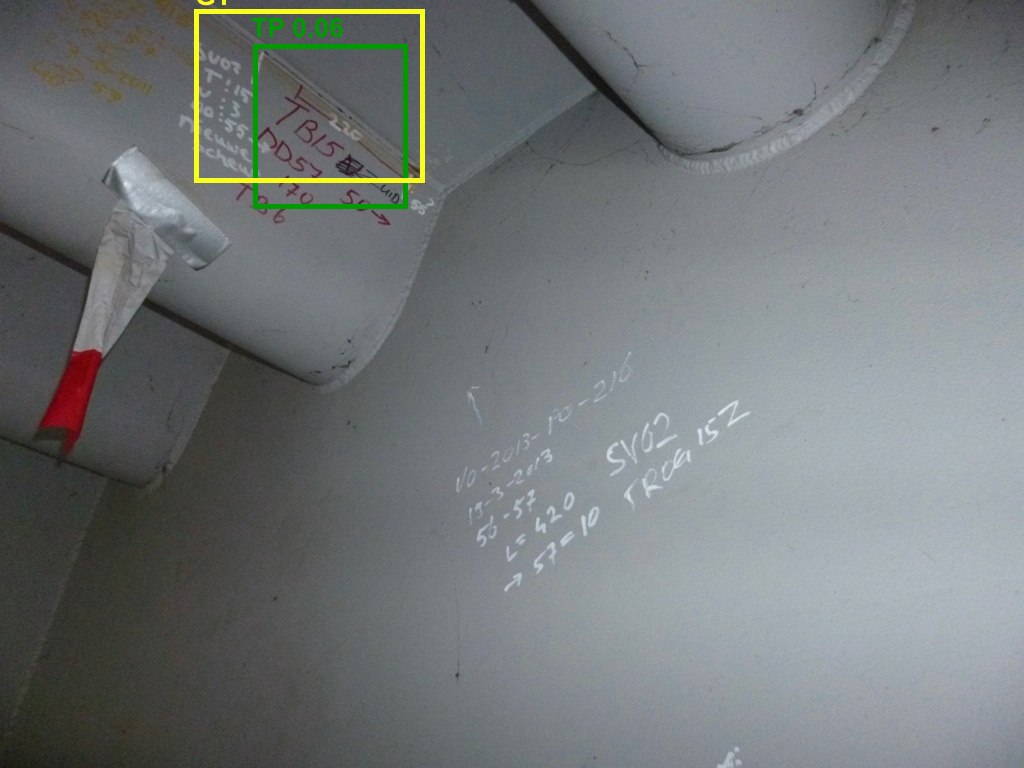}
        \caption{}
        \label{fig:NN_Outputs 2}
    \end{subfigure}
    \hspace{0.0em}
    \begin{subfigure}[c]{0.32\textwidth}
        \includegraphics[width=\textwidth]{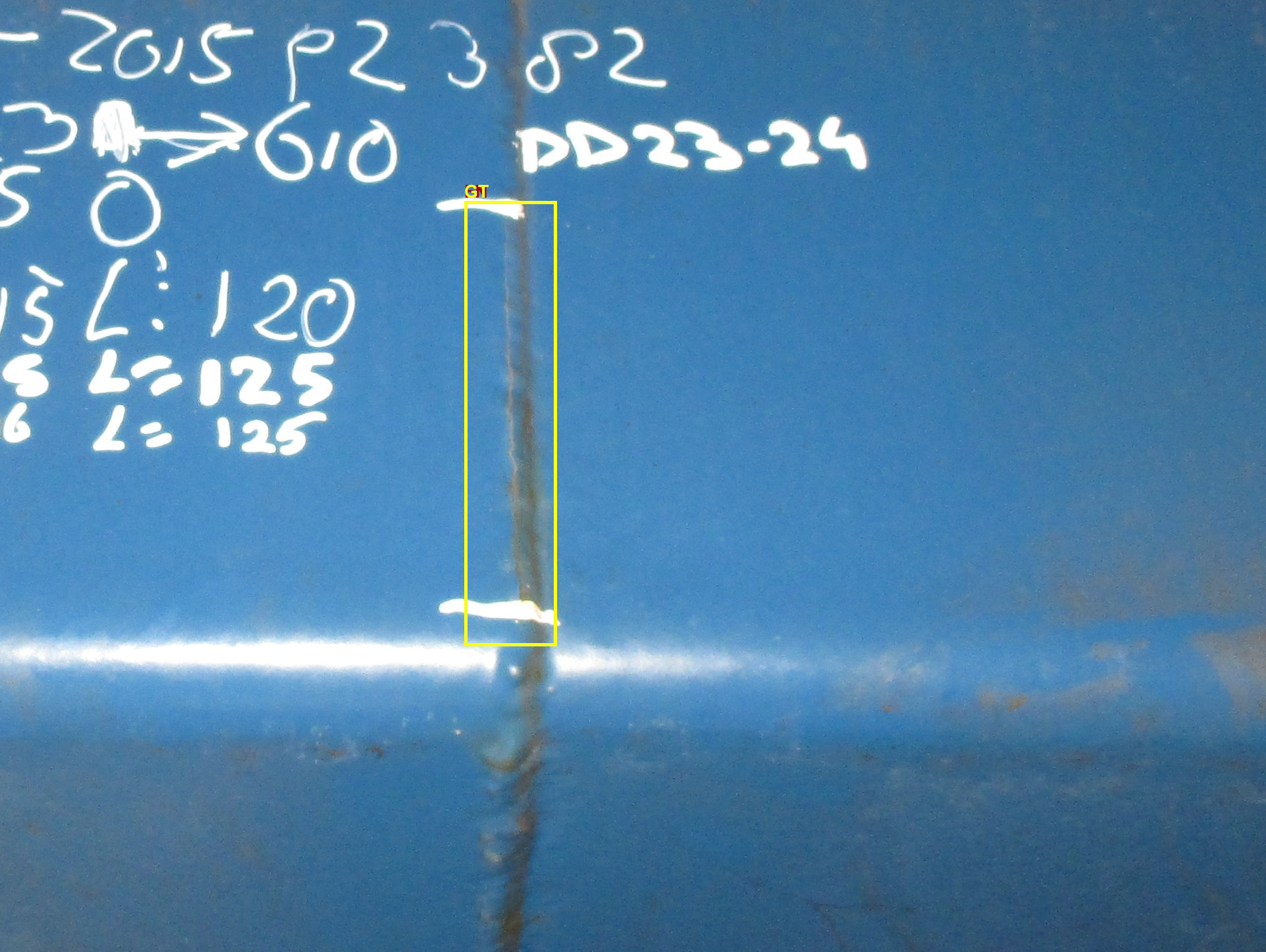}
        \caption{}
        \label{fig:NN_Outputs 3}
    \end{subfigure}
    \caption{Examples of the true positive (green), false negative (red), and false positive (blue) bounding boxes produced by the trained CV method. Annotated ground truth bounding boxes shown in yellow.}
    \label{fig:NN_Outputs}
\end{figure*}

Several examples of the output of the trained neural network are presented in Fig.~\ref{fig:NN_Outputs}, with indications of true positive, false negative, and false positive detections in green, red, and blue, respectively. Annotated ground truth bounding boxes are shown in yellow. Fig. \ref{fig:NN_Outputs 1} shows an example in which the network incorrectly identifies a dark line as a crack; this is the most common type of false positive.

We developed a set of rules to automatically classify bounding boxes as true positives or false positives by comparing the detections with ground-truth bounding boxes. The predicted boxes were considered in order of decreasing confidence and assigned to their best-overlapping ground-truth box. A prediction was scored as a true positive if it met any of the following three criteria: an intersection-over-union of at least 0.2, at least 80 \% of the ground-truth box covered by the prediction, or at least 80 \% of the predicted box covered by the ground-truth box (a prediction fully enclosed by the ground-truth box automatically satisfying the latter). This multiple-criterion rule accommodates the thin, elongated geometry of cracks, for which the intersection-over-union alone is overly strict. Each ground-truth box could be matched only once. Any remaining unmatched predictions were counted as false positives, and any unmatched ground-truth boxes as false negatives.

\section{Probability of detection curve}\label{sec: Probability of detection}
In the field of bridge inspection, it is standard practice to evaluate an inspection method using a PoD curve, which shows the probability that this inspection method will detect some damage, as a function of the damage characteristic size. A large variety of inspection methods have been evaluated using PoD curves, including visual inspection \cite{Campbell2019pod}, ultrasonic inspection, and liquid penetrant inspection \cite{righiniotis2006comparativePoDs}. 
There are established methodologies and standards for the estimation of PoD curves, most prominently MIL-HDBK-1823A \cite{MILHDBK1823A}. The methods for statistical analysis for PoD data can also be found in \cite{ENIQ_POD_BestPractices_2011,HSE2006ProbabilityCurves}. An example PoD curve can be seen in Fig.~\ref{fig:PoD, cb, comp.}.

The first step when estimating the PoD curve of an inspection method is to collect multiple measurements in samples with a known \emph{characteristic  size} of the damage, often denoted by $a$, in a realistic environment. Measurement data generally come in two forms. In the first form, often called ``$\hat{a}$ vs $a$'', one measures a continuous real-valued signal $\hat{a}$. In the second form, called ``hit/miss'', one only records whether the damage was detected (hit) or not detected (miss). In both cases, the ground truth size of the damage is also recorded. It is possible to transform a continuous detection signal $\hat{a}$ into a hit/miss record by choosing a threshold $\hat{a}_\mathrm{th}$, which means that the crack is considered to have been detected when $\hat{a} > \hat{a}_\mathrm{th}$.

\subsection{Methodology for PoD curve determination}\label{sec:PoD:PoD estimation method}
The recommended way to determine the PoD curve from hit/miss data is to fit a Generalised Linear Model (GLM) \cite{ENIQ_POD_BestPractices_2011}. Denoting the crack length by $a$ (in millimetres) and the probability of detection by $p$, the GLM takes the form:
\begin{equation}\label{eq:glm_general}
g(p) = \beta_0 + \beta_1 h(a)
\end{equation}
where $h(a) = a$ or $h(a) = \log(a)$, depending on which option allows a better fit of the PoD model, $\beta_0$ and $\beta_1$ are the model parameters and $g$ is the link function.
Using the logit link function, as is standard practice \cite{ENIQ_POD_BestPractices_2011}, we end up in the setting of logistic regression. Hence, the GLM in Eq.~\eqref{eq:glm_general} becomes
\begin{equation}\label{eq:logit_length}
\log\left(\frac{p}{1 - p}\right) = \beta_0 + \beta_1 h(a).
\end{equation}
When solving for $p$, the probability of detection of a crack of length $a$ is
\begin{equation}\label{eq:pod_length}
\PoD(a; \beta_0, \beta_1) \coloneqq p = \frac{\exp(\beta_0 + \beta_1 h(a))}{1 + \exp(\beta_0 + \beta_1 h(a))}.
\end{equation}
Parameters $\beta_0$ and $\beta_1$ are determined using the maximum likelihood method on a data set $\mathcal{S}$ of $n$ independent samples
\begin{equation}\label{eq:samples}
\mathcal{S} \coloneqq \{(a_i, Y_i)\}_{i = 1}^n,
\end{equation}
where $a_i$ is the true length of the crack and $Y_i \in \{0, 1\}$ indicates whether the crack is detected, $Y_i = 1$, or not detected , $Y_i = 0$. 
The likelihood of a pair of parameters $\beta_0, \beta_1$ is then given by 
\ifreview
\begin{equation}\label{eq:likelihood}
L(\beta_0, \beta_1; \mathcal{S}) = \prod_{i = 1}^n \PoD(a_i; \beta_0, \beta_1)^{Y_i} (1 - \PoD(a_i; \beta_0, \beta_1))^{1 - Y_i}.
\end{equation}
\else
\begin{multline}\label{eq:likelihood}
L(\beta_0, \beta_1; \mathcal{S}) \\= \prod_{i = 1}^n \PoD(a_i; \beta_0, \beta_1)^{Y_i} (1 - \PoD(a_i; \beta_0, \beta_1))^{1 - Y_i}.
\end{multline}
\fi
 Instead of optimising the likelihood directly, it is more convenient to maximise the log-likelihood, which is given by
 \ifreview
\begin{equation}\label{eq:log-likelihood}
\begin{split}
\ell(\beta_0, \beta_1; \mathcal{S}) & \coloneqq \log(L(\beta_0, \beta_1; \mathcal{S})) \\
& = \sum_{i = 1}^n \Big[ Y_i \log(\PoD(a_i; \beta_0, \beta_1)) + (1 - Y_i) \log(1 - \PoD(a_i; \beta_0, \beta_1)) \Big] \\
& = \sum_{i = 1}^n \Big[Y_i (\beta_0 + \beta_1 h(a_i)) - \log(1 + \exp(\beta_0 + \beta_1 h(a_i))) \Big].
\end{split}
\end{equation}
\else
\begin{equation}\label{eq:log-likelihood}
\begin{split}
\ell(\beta_0, \beta_1; \mathcal{S}) & \coloneqq \log(L(\beta_0, \beta_1; \mathcal{S})) \\
& = \sum_{i = 1}^n \Big[Y_i \log(\PoD(a_i; \beta_0, \beta_1)) \\
    & \qquad + (1 - Y_i) \log(1 - \PoD(a_i; \beta_0, \beta_1)) \Big] \\
& = \sum_{i = 1}^n \Big[Y_i (\beta_0 + \beta_1 h(a_i)) \\
    & \qquad - \log(1 + \exp(\beta_0 + \beta_1 h(a_i))) \Big].
\end{split}
\end{equation}
\fi
We optimise Eq.~\eqref{eq:log-likelihood} using the \texttt{LogisticRegression} method implemented in the Python library \texttt{scikit-learn} \cite{scikit-learn}. The PoD curve is then found by filling in the fitted parameters $\beta_0, \beta_1$ into Eq.~\eqref{eq:pod_length}; see Fig.~\ref{fig:PoD, cb, comp.} for some examples of PoD curves.

\subsubsection{Confidence bounds}\label{sec:PoD:confidence bounds}
We use the statistics of maximum likelihood estimators to compute confidence bounds on the PoD curve. Under the assumption that the data $\mathcal{S}$ are generated according to Eq.~\eqref{eq:pod_length} with some parameters $\beta_0$, $\beta_1$, with corresponding maximum likelihood estimates $\hat{\beta}_0$, $\hat{\beta}_1$, Wilk's likelihood ratio statistic, given by
\begin{equation}\label{eq:wilks_statistic_resolution}
W(\beta_0, \beta_1; \mathcal{S}) \coloneqq -2(\ell(\beta_0, \beta_1; \mathcal{S}) - \ell(\hat{\beta}_0, \hat{\beta}_1; \mathcal{S})),
\end{equation}
asymptotically follows a $\chi$-squared distribution with two degrees of freedom, since the model has two free parameters. We then define the 95~\% confidence parameter region as $B(\mathcal{S}) \coloneqq \{(\beta_0, \beta_1) \in \R^2 \mid W(\beta_0, \beta_1; \mathcal{S}) \leq \chi_{2,0.95}^2 \approx 5.99 \}$. In words, this means that we include parameters $(\beta_0, \beta_1)$ in our confidence region $B(\mathcal{S})$ if the probability of observing such a large value for Wilk's likelihood ratio is at least 5~\%, under the hypothesis that the data are distributed according to Eq.~\eqref{eq:pod_length} with those parameters. Each pair of parameters $(\beta_0, \beta_1) \in B(\mathcal{S})$ defines a PoD curve; the confidence bounds are found by taking the envelope of all such PoD curves, i.e.
\begin{equation}\label{eq:confidence_bounds_length}
\begin{aligned}
\PoD^{\textrm{upper}}(a; \mathcal{S}) & = \max_{(\beta_0, \beta_1) \in B(\mathcal{S})} \frac{\exp(\beta_0 + \beta_1 h(a))}{1 + \exp(\beta_0 + \beta_1 h(a))}, \textrm{ and} \\
\PoD^{\textrm{lower}}(a; \mathcal{S}) &= \min_{(\beta_0, \beta_1) \in B(\mathcal{S})} \frac{\exp(\beta_0 + \beta_1 h(a))}{1 + \exp(\beta_0 + \beta_1 h(a))}.
\end{aligned}
\end{equation}

\subsection{Results of PoD curve determination}\label{sec:PoD:PoD for CV crack detection}

\begin{figure*}
\centering
\includegraphics[width=\textwidth]{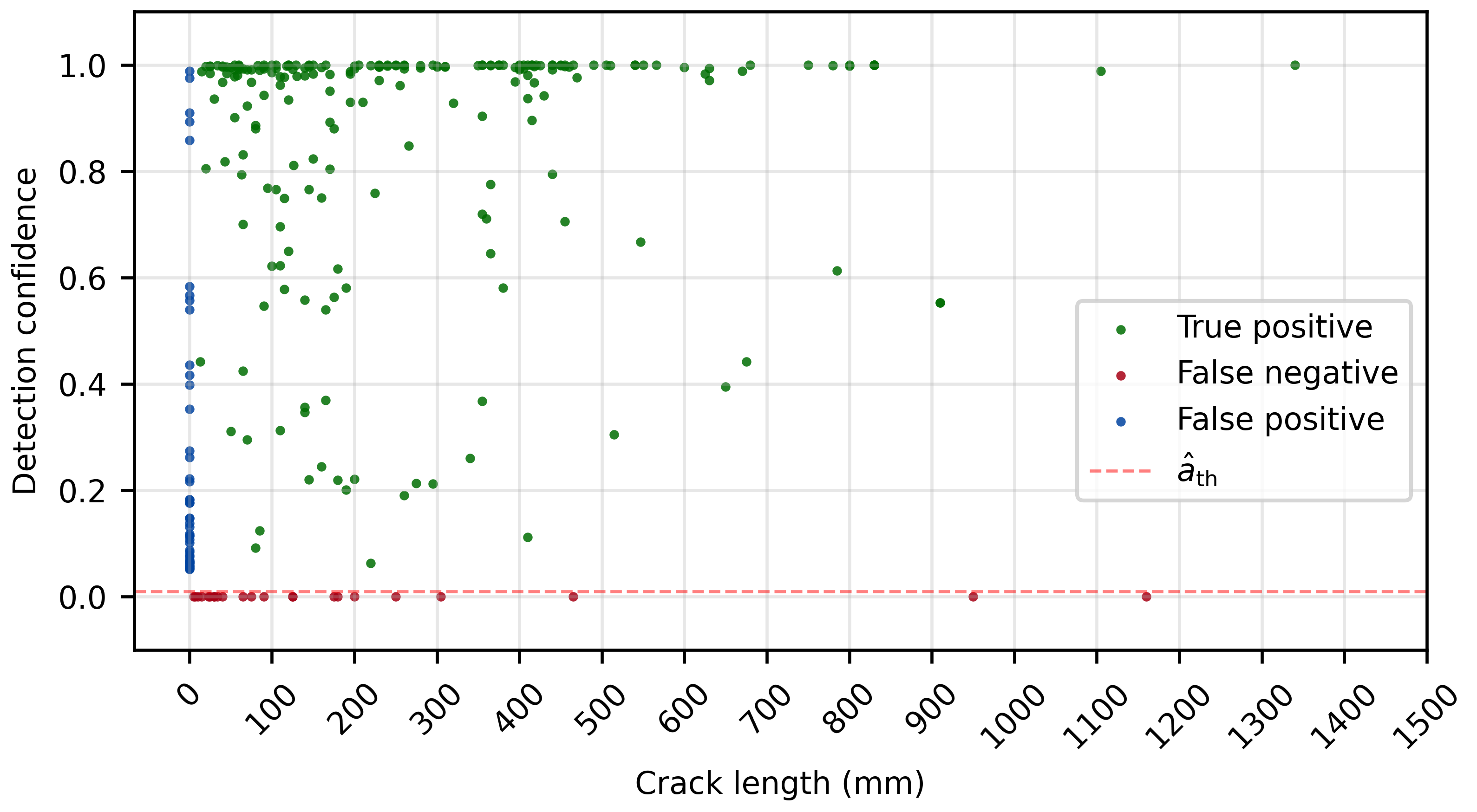}
\caption{Detection confidence of our CV method on the PoD estimation dataset.}
\label{fig:Detection results}
\end{figure*}

The CV method previously described in Sec.~\ref{sec:Deep learning method for crack detection} produces a detection confidence score between 0 and 1 for each image. Fig.~\ref{fig:Detection results} shows this score for every image in the dataset, as a function of the crack length. As in Fig.~\ref{fig:NN_Outputs}, the green, red, and blue marks correspond to true positives (hits), false negatives (misses), and false positives, respectively.
Note that if a linear model were fitted to these data, the errors would not follow a normal distribution. This violates one of the basic assumptions behind the methods for computing PoD curves in $a$ vs $\hat{a}$ \cite{MILHDBK1823A}.

\begin{figure*}
\centering
\includegraphics[width=0.7\textwidth]{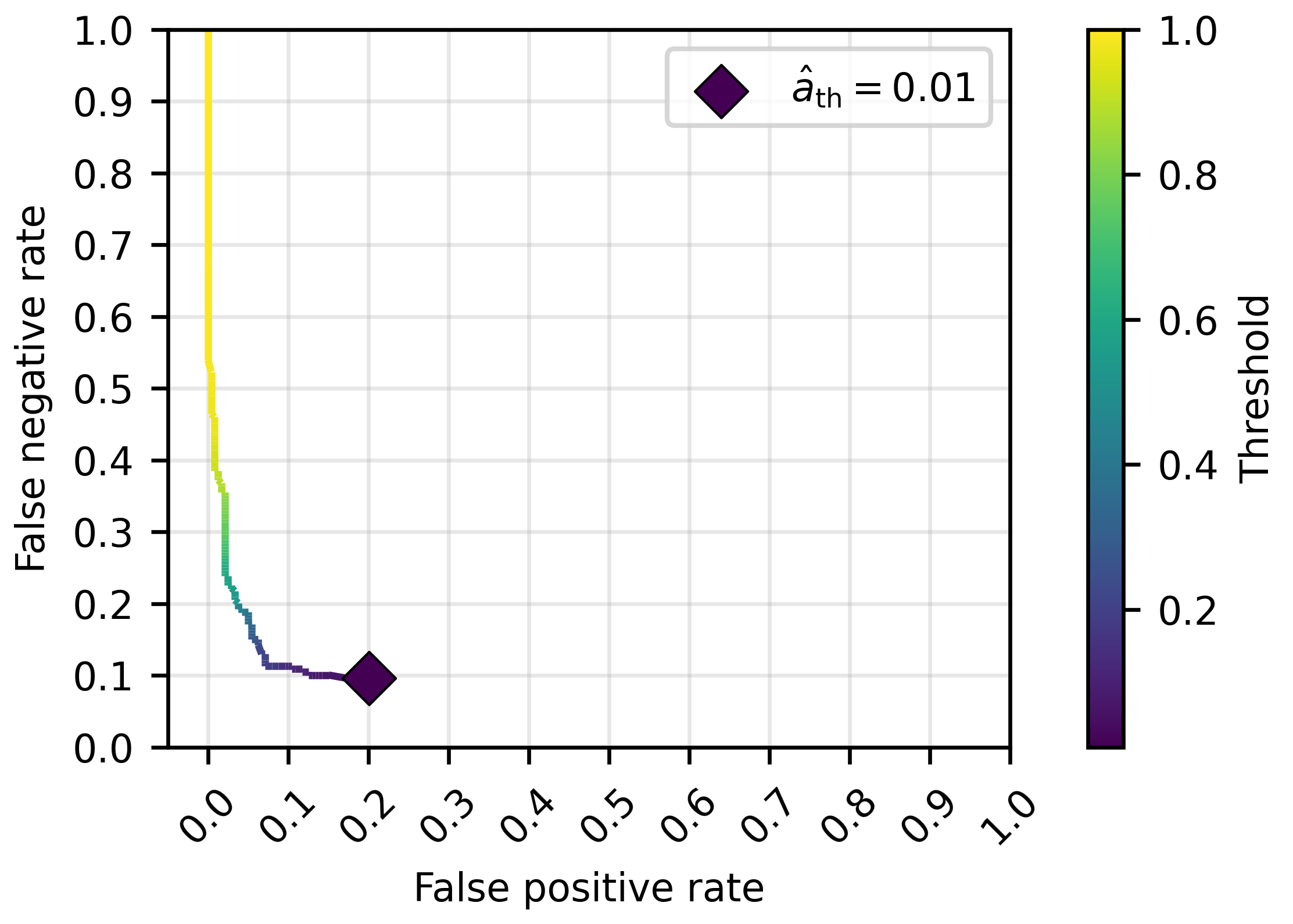}
\caption{Effect of the decision threshold $\hat{a}_\mathrm{th}$ on the tradeoff between false positive and false negative detection rates of our CV method.}
\label{fig:FP-FN tradeof}
\end{figure*}

As a consequence, we choose a threshold to obtain hit/miss data, in such a way as to balance the probability of detection with false positive errors. During bridge inspection, it is particularly important not to miss cracks, as undetected cracks may compromise the operational safety of the bridge, whereas false positives produced by a CV algorithm can be easily removed by human verification. As can be seen in Fig. \ref{fig:FP-FN tradeof},  at a very low value of $\hat{a}_\mathrm{th} = 0.01$, false positive detections occur in 20~\% of images, while false negative errors are minimised to 10\%. Thus, for the experiments in this study, we fix $\hat{a}_\mathrm{th} = 0.01$. Notably, a practicioner could choose a different threshold to change the tradeoff between false positive and false negative rate to better suit the requirements of the application.

We fitted the PoD model on the PoD estimation dataset, following the method described in Sec.~\ref{sec:PoD:PoD estimation method}, using the transformation $h(a) = \log(a)$ since this is commonly used and, as we show in Sec. \ref{sec:Effect of Resolution on PoD:pod surface}, it provides the best fit of the model to the data (cf.~Fig.~\ref{fig:selection_transforms_length_resolution}). The maximum likelihood estimates of the parameters are $\beta_0 = -1.890$ and $\beta_1 = 0.858$, so that the fitted PoD curve is
\begin{equation}\label{eq:pod_length_fitted}
\PoD_\textrm{CV}(a) = \frac{\exp(-1.890 + 0.858 \log(a))}{1 + \exp(-1.890 + 0.858 \log(a))}.
\end{equation}
As baselines, we use the PoD curve provided in the DNVGL recommendation~\cite{DNVGL_RP_C210_2015}, and the PoD curve estimated by Campbell et al.~\cite{Campbell2019pod}. It should be noted that both curves correspond to conventional visual inspection. 
The PoD curve in the DNVGL recommendation is estimated based on both data and expert opinion \cite{DNVGL_RP_C210_2015}, and is the accepted standard in the field. The PoD curve from the DNVGL recommendation is given by
\begin{equation}\label{eq:pod_dnvgl}
\PoD_\textrm{DNVGL}(a) = 1 - \frac{1}{1 + (a / 37.15)^{0.954}}.
\end{equation}
Campbell et al.~\cite{Campbell2019pod} derived their PoD curve using a methodology similar to the one used in this study. 
The PoD curve from Campbell et al.\ uses the GLM given in Eq.~\eqref{eq:pod_length}, with $h(a) = a$ and parameters $\beta_0 = -0.498$ and $\beta_1 = 0.0194$, so that
\begin{equation}\label{eq:pod_Campbell}
\PoD_\textrm{Campbell}(a) = \frac{\exp(-0.498 + 0.0194 a)}{1 + \exp(-0.498 + 0.0194 a)}. 
\end{equation}
It also should be mentioned that in the study by Campbell et al, the PoD was evaluated from the inspection of cracks with length up to $5 \frac{3}{8}$ inches, which is less than 150 mm. The inspection was typically performed at the distance of an arm's length \cite{Campbell2019pod}.

\begin{figure*}
    \centering
    \includegraphics[width=\textwidth]{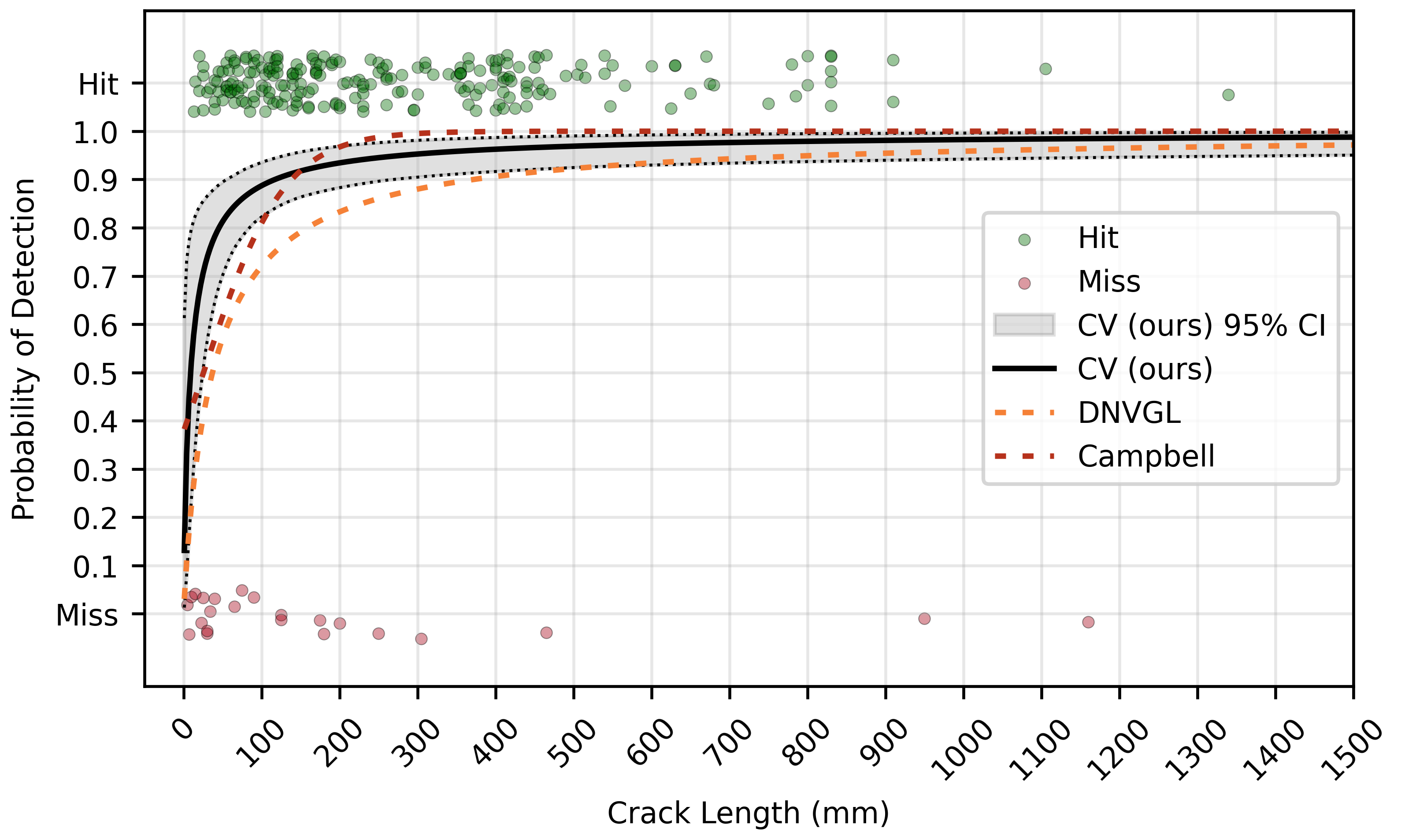}
    \caption{
    PoD model (Eq.~\eqref{eq:pod_length}, black) with 95~\% confidence intervals (gray) fitted to the hit/miss data generated by our CV method (cf.~Fig.~\ref{fig:Detection results}), compared to the PoD curves for conventional inspection methods by DNVGL (orange) and Campbell et al.\ (red). For better visualisation, the hit/miss data points are scattered around horizontal lines at 0 and 1.1.}
    \label{fig:PoD, cb, comp.}
\end{figure*}

The three PoD curves are shown in Fig.~\ref{fig:PoD, cb, comp.}. We see that the PoD of our CV method is superior to conventional visual inspection for cracks up to about 150 mm in length. 
For longer cracks, the detection capability of our CV method is slightly worse. 
This may be explained by the fact that it could be difficult to observe smaller cracks for a human inspector and the results may be affected by the experience of the inspector or inspection conditions. 
However, if automated inspection were to be conducted, it would be easier to get a good image of a small crack with a good camera and the results of the CV method would be expected to be more consistent. 
At the same time, a human inspector rarely misses a relatively large crack ($>$300 mm), while a CV method may wrongly classify it as background noise given that the image would inevitably include a large portion of the structure. Alternatively, it is possible that the PoD curve determined by Campbell et al.\ is not valid for such large cracks, since these were not present in the data set used for fitting the curve.

\section{Framework for comparison of PoD curves}\label{sec:Effect on crack length update after inspection}
We propose a framework to compare inspection methods based on their PoD curves using Monte Carlo simulation.
The approach is modelled on a structural reliability assessment procedure used to estimate the crack size distribution of a structure at a given moment during its useful life \cite{Leonetti2021InfluenceModels,Maljaars2014ProbabilisticLocations}. In this procedure, the crack size distribution is updated using the PoD curve when inspections are performed.
% The approach reflects the procedure used in a structural reliability assessment in which the crack size distribution at a given moment during the useful life of the structure is updated due to an inspection based on the PoD curve of the inspection method \cite{DNVGL_RP_C210_2015}. 
In a structural reliability assessment, the crack size distribution is obtained from a probabilistic fracture mechanics model; we instead calculate the influence of inspections for specified prior crack length distributions.

\subsection{Methodology for comparing PoD curves with Monte Carlo simulation}
% Bert: Apparently you are not satisfied with the visual comparison in Figure 6. Why not? Then that is the reason to develop a MC based comparison method. So, I am missing the reason why we propose a MC simulation to compare inspection methods based on their PoD curve.
We now describe our framework for comparing PoD curves, summarised in Alg.~\ref{alg:missed_crack_distribution}, in detail.
We first assume some prior log-normal distribution of the crack length. This represents the distribution of crack lengths in the structure before inspection.
Next, we simulate an inspection event, which is based on the Bayes theorem. 
Hence, we numerically estimate the posterior crack distribution after inspection by sampling a crack length from the prior log-normal distribution. 
This crack size is used to determine the associated probability of detection using the PoD curve. 
We uniformly sample a random number between 0 and 1, and mark the crack as detected if this number is less than the probability of detection.
If the sampled crack is missed by an inspection method, it is stored in the array of missed cracks, $\mathcal{M}$. 
For each inspection method, we continue to sample cracks from the prior distribution until a statistically sufficient number, namely $10^6$, of missed cracks is stored for each inspection method. This number is estimated empirically, so that when repeating the same experiment with the same inputs, but with different random instantiations, the deviations in performance are at most $10^{-4}$. 
For our automated CV inspection method, we use the maximum likelihood PoD curve estimated in this study, given by Eq.~\eqref{eq:pod_length_fitted}, and for conventional visual inspection, we use the PoD curves provided by the DNVGL recommendation (Eq.~\eqref{eq:pod_dnvgl}) \cite{DNVGL_RP_C210_2015} and by Campbell et al.\ (Eq.~\eqref{eq:pod_Campbell}) \cite{Campbell2019pod}.

\begin{algorithm}
\caption{Monte Carlo estimation of posterior crack-length distributions after inspection}
\label{alg:missed_crack_distribution}
\begin{algorithmic}[1]
\Require Prior crack distribution $P_{\mathrm{prior}}(\mu_c,\CoV_c)$
\Require Probability of detection curve for the inspection method $\PoD_{\mathrm{CV}}$, $\PoD_{\mathrm{Campbell}}$, $\PoD_{\mathrm{DNVGL}}$
\Require Target number of missed cracks $N_{\mathrm{target}}$

\State Initialize empty lists:
$\mathcal{I} \gets \emptyset$,
$\mathcal{M}_{\mathrm{CV}} \gets \emptyset$,
$\mathcal{M}_{\mathrm{Campbell}} \gets \emptyset$,
$\mathcal{M}_{\mathrm{DNVGL}} \gets \emptyset$

\While{$\min\{|\mathcal{M}_{\mathrm{CV}}|, |\mathcal{M}_{\mathrm{Campbell}}|, |\mathcal{M}_{\mathrm{DNVGL}}|\} < N_{\mathrm{target}}$}

    \State Sample crack length $l \sim P_{\mathrm{prior}}(\mu_c,\CoV_c)$
    \State Append $l$ to $\mathcal{I}$.

    \State Sample $u \sim \mathcal{U}(0,1)$
    \For{each method $j \in \{\mathrm{CV}, \mathrm{Campbell}, \mathrm{DNVGL}\}$}
        \If{$u > \PoD_j(l)$}
            \State Append $l$ to $\mathcal{M}_j$
        \EndIf
    \EndFor
\EndWhile
\For{each PoD model $j \in \{\mathrm{CV},\mathrm{Campbell},\mathrm{DNVGL}\}$}
    \State Fit a lognormal distribution to the missed-crack sample set $\mathcal{M}_j$
    \State Compute the normalised undetected crack length $C_j$ according to Eq.~\eqref{eq: normalised total crack length}
    \State Compute the Kullback-Leibler divergence $\KL_j$ according to Eq.~\eqref{eq:KL}
\EndFor

\State \textbf{return} Pre- and post-inspection crack distribution histograms; $C$ and $\KL$ metrics
\end{algorithmic}
\end{algorithm}
\label{sec:Effect on crack length update after inspection:results}\definecolor{cCV}{rgb}{0, 0, 0}
\definecolor{cCampbell}{rgb}{1, 0, 0}
\definecolor{cDNVGL}{rgb}{1.0, 0.5, 0.0}
\pgfplotstableread{Figures/Update_Crack_Length/update_crack_length_C.dat}\datatable
\pgfplotsset{every axis plot/.append style={very thick}}
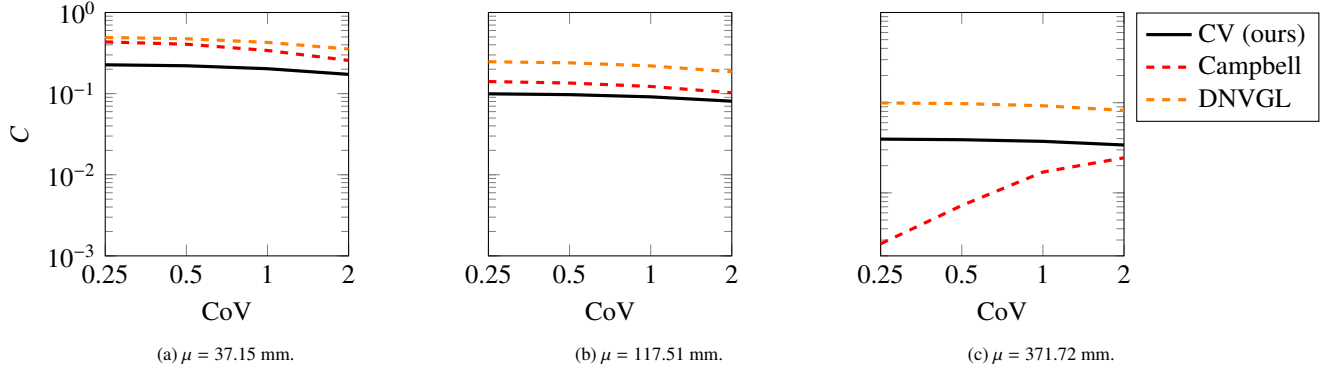
\begin{figure*}
\centering
\begin{subfigure}[t]{0.36\linewidth}
\begin{tikzpicture}[baseline=(current bounding box.south west)]
\begin{loglogaxis}[
    width=4.8cm,
    height=4.8cm,
    xlabel={$\CoV$},
    ylabel={$C$},
    xtick={0.25, 0.5, 1, 2},
    xticklabels={0.25, 0.5, 1, 2},
    xmin=0.25, xmax=2,
    ymin=0.001, ymax=1,
    legend pos=south west,
    legend cell align = {left},
]

\addplot[cCV]
table[
    x=CoV,
    y=CV,
    restrict expr to domain={\thisrow{fig}}{0:0},
] {\datatable};

\addplot[cCampbell,dashed]
table[
    x=CoV,
    y=Campbell,
    restrict expr to domain={\thisrow{fig}}{0:0},
] {\datatable};

\addplot[cDNVGL,dashed]
table[
    x=CoV,
    y=DNVGL,
    restrict expr to domain={\thisrow{fig}}{0:0},
] {\datatable};

\end{loglogaxis}
\end{tikzpicture}
\caption{$\mu = 37.15$ mm.}
\end{subfigure}
\begin{subfigure}[t]{0.31\linewidth}
\begin{tikzpicture}[baseline=(current bounding box.south)]
\begin{loglogaxis}[
    width=4.8cm,
    height=4.8cm,
    xlabel={$\CoV$},
    xtick={0.25, 0.5, 1, 2},
    xticklabels={0.25, 0.5, 1, 2},
    yticklabels=none,
    xmin=0.25, xmax=2,
    ymin=0.001, ymax=1,
]

\addplot[cCV]
table[
    x=CoV,
    y=CV,
    restrict expr to domain={\thisrow{fig}}{1:1},
] {\datatable};

\addplot[cCampbell,dashed]
table[
    x=CoV,
    y=Campbell,
    restrict expr to domain={\thisrow{fig}}{1:1},
] {\datatable};

\addplot[cDNVGL,dashed]
table[
    x=CoV,
    y=DNVGL,
    restrict expr to domain={\thisrow{fig}}{1:1},
] {\datatable};

\end{loglogaxis}
\end{tikzpicture}
\caption{$\mu = 117.51$ mm.}
\end{subfigure}
\begin{subfigure}[t]{0.31\linewidth}
\begin{tikzpicture}[baseline=(current bounding box.south)]
\begin{loglogaxis}[
    width=4.8cm,
    height=4.8cm,
    xlabel={$\CoV$},
    xtick={0.25, 0.5, 1, 2},
    xticklabels={0.25, 0.5, 1, 2},
    yticklabels=none,
    xmin=0.25, xmax=2,
    ymin=0.002, ymax=1,
    legend style={
        at={(1.05,1)},
        anchor=north west,
        cells={anchor=west}
    },
]

\addplot[cCV]
table[
    x=CoV,
    y=CV,
    restrict expr to domain={\thisrow{fig}}{2:2},
] {\datatable};

\addplot[cCampbell,dashed]
table[
    x=CoV,
    y=Campbell,
    restrict expr to domain={\thisrow{fig}}{2:2},
] {\datatable};

\addplot[cDNVGL,dashed]
table[
    x=CoV,
    y=DNVGL,
    restrict expr to domain={\thisrow{fig}}{2:2},
] {\datatable};

\addlegendentry{CV (ours)}
\addlegendentry{Campbell}
\addlegendentry{DNVGL}

\end{loglogaxis}
\end{tikzpicture}
\caption{$\mu = 371.72$ mm.}
\end{subfigure}
\caption{Results of computation of the normalised undetected crack length $C$ for our CV method and conventional visual inspection provided by Campbell et al.~\cite{Campbell2019pod} and by the DNVGL recommendation \cite{DNVGL_RP_C210_2015}}\label{fig:C graphs}
\end{figure*}

We consider the prior distributions of crack length to be a log-normal distribution. Within the sensitivity study, we consider 12 distributions, varying the mean $\mu$ (37.15~mm; 117.51~mm; 371.72~mm) and the coefficient of variation $\CoV$ (0.25; 0.5; 1; 2).
It should be noted that a random variable $X$ is log-normally distributed with mean $\mu$ and coefficient of variation $\CoV$ if $X = \exp(\mu_\mathcal{N} + \sigma_\mathcal{N} Z)$, where $Z$ has a standard normal distribution, and $\mu_\mathcal{N} = \log(\mu / \sqrt{1 + \CoV^2})$ and $\sigma_\mathcal{N}^2 = \log(1 + \CoV^2)$.
The mean $\mu$ is selected based on the PoD provided by the DNVGL recommendation. 
Specifically, crack lengths of 37.15~mm, 117.51~mm, and 371.72~mm correspond to PoD 50~\%, 75~\%, 90~\% in the PoD curve from the DNVGL recommendation, respectively. The coefficient of variation $\CoV$ is chosen to represent different levels of uncertainty in the prior distribution of the crack length, ranging from relatively concentrated crack populations ($\CoV = 0.25$) to highly dispersed populations ($\CoV = 2$). 
This allows for a sensitivity analysis of the inspection-method comparison framework with respect to the assumed prior crack size and variability.

Finally, we quantify the effect of the inspection with two metrics: (a) the normalised undetected crack length $C$, and (b) the Kullback-Leibler divergence evaluated between prior and posterior crack distributions $\KL$.
The normalised undetected crack length is computed as:
\begin{equation}
\label{eq: normalised total crack length}
C = \frac{\sum_{i \in \mathcal{M}} a_i}{\sum_{i \in \mathcal{I}} a_i},
\end{equation}
where $\mathcal{M}$ is the set of missed cracks and $\mathcal{I}$ is the set of sampled cracks from the prior distribution. 
A value of $C = 1$ indicates that no crack has been detected. As $C$ approaches 0, the total length of missed cracks decreases; better inspection methods will typically have a smaller $C$. However, $C$ does not describe how the missed crack lengths are distributed. For example, the same value of $C$ may result from missing many small cracks or from missing a smaller number of longer cracks, although these cases may have different implications for structural reliability.

\pgfplotstableread{Figures/Update_Crack_Length/update_crack_length_KL.dat}\datatable
\pgfplotsset{every axis plot/.append style={very thick}}
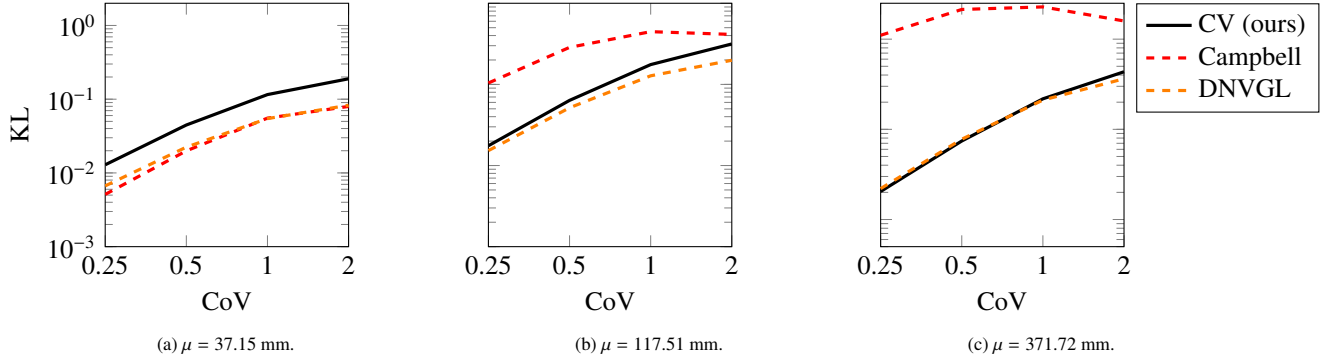
\begin{figure*}
\centering
\begin{subfigure}[t]{0.36\linewidth}
\begin{tikzpicture}[baseline=(current bounding box.south west)]
\begin{loglogaxis}[
    width=4.8cm,
    height=4.8cm,
    xlabel={$\CoV$},
    ylabel={$\KL$},
    xtick={0.25, 0.5, 1, 2},
    xticklabels={0.25, 0.5, 1, 2},
    xmin=0.25, xmax=2,
    ymin=0.001, ymax=2,
    legend pos=south east,
]

\addplot[cCV]
table[
    x=CoV,
    y=CV,
    restrict expr to domain={\thisrow{fig}}{0:0},
] {\datatable};

\addplot[cCampbell,dashed]
table[
    x=CoV,
    y=Campbell,
    restrict expr to domain={\thisrow{fig}}{0:0},
] {\datatable};

\addplot[cDNVGL,dashed]
table[
    x=CoV,
    y=DNVGL,
    restrict expr to domain={\thisrow{fig}}{0:0},
] {\datatable};

\end{loglogaxis}
\end{tikzpicture}
\caption{$\mu = 37.15$ mm.}
\end{subfigure}
\begin{subfigure}[t]{0.31\linewidth}
\begin{tikzpicture}[baseline=(current bounding box.south)]
\begin{loglogaxis}[
    width=4.8cm,
    height=4.8cm,
    xlabel={$\CoV$},
    xtick={0.25, 0.5, 1, 2},
    xticklabels={0.25, 0.5, 1, 2},
    yticklabels=none,
    xmin=0.25, xmax=2,
    ymin=0.001, ymax=1,
]

\addplot[cCV]
table[
    x=CoV,
    y=CV,
    restrict expr to domain={\thisrow{fig}}{1:1},
] {\datatable};

\addplot[cCampbell,dashed]
table[
    x=CoV,
    y=Campbell,
    restrict expr to domain={\thisrow{fig}}{1:1},
] {\datatable};

\addplot[cDNVGL,dashed]
table[
    x=CoV,
    y=DNVGL,
    restrict expr to domain={\thisrow{fig}}{1:1},
] {\datatable};

\end{loglogaxis}
\end{tikzpicture}
\caption{$\mu = 117.51$ mm.}
\end{subfigure}
\begin{subfigure}[t]{0.31\linewidth}
\begin{tikzpicture}[baseline=(current bounding box.south)]
\begin{loglogaxis}[
    width=4.8cm,
    height=4.8cm,
    xlabel={$\CoV$},
    xtick={0.25, 0.5, 1, 2},
    xticklabels={0.25, 0.5, 1, 2},
    yticklabels=none,
    xmin=0.25, xmax=2,
    ymin=0.005, ymax=2.5,
    legend style={
        at={(1.05,1)},
        anchor=north west,
        cells={anchor=west}
    },
]

\addplot[cCV]
table[
    x=CoV,
    y=CV,
    restrict expr to domain={\thisrow{fig}}{2:2},
] {\datatable};

\addplot[cCampbell,dashed]
table[
    x=CoV,
    y=Campbell,
    restrict expr to domain={\thisrow{fig}}{2:2},
] {\datatable};

\addplot[cDNVGL,dashed]
table[
    x=CoV,
    y=DNVGL,
    restrict expr to domain={\thisrow{fig}}{2:2},
] {\datatable};

\addlegendentry{CV (ours)}
\addlegendentry{Campbell}
\addlegendentry{DNVGL}

\end{loglogaxis}
\end{tikzpicture}
\caption{$\mu = 371.72$ mm.}
\end{subfigure}
\caption{Results of computation of the Kullback-Leibler divergence $\KL$ for our CV method and conventional visual inspection provided by Campbell et al.~\cite{Campbell2019pod} and by the DNVGL recommendation \cite{DNVGL_RP_C210_2015}}\label{fig:KL graphs}
\end{figure*}

The Kullback-Leibler (KL) divergence is computed as:
\ifreview
\begin{equation}\label{eq:KL}
\KL(P_\mathrm{post} \,||\, P_\mathrm{prior}) = \int_0^\infty P_\mathrm{post}(a) \log(P_\mathrm{post}(a) / P_\mathrm{prior}(a)) \, {\rm d} a.
\end{equation}
\else
\begin{multline}\label{eq:KL}
\KL(P_\mathrm{post} \,||\, P_\mathrm{prior}) \\ = \int_0^\infty P_\mathrm{post}(a) \log(P_\mathrm{post}(a) / P_\mathrm{prior}(a)) \, {\rm d} a.
\end{multline}
\fi
The KL divergence measures a generalised notion of distance between the distribution of crack lengths after inspection (posterior $P_\mathrm{post}$) relative to the initial distribution of crack lengths (prior $P_\mathrm{prior}$). 
The posterior $P_\mathrm{post}$ is determined by fitting a log-normal distribution on the sample of missed cracks.
The KL divergence is a standard way of measuring the difference between prior and posterior distributions \cite{Kullback1951KLDiv}. 
A higher value of KL would generally indicate a stronger effect of the inspection method and a shift of the crack distribution to the lower crack length.  However, the KL divergence is computed between normalised probability distributions and therefore does not account for the total amount of missed crack damage. As a result, an inspection method may produce a large shift in the crack-size distribution while still leaving a substantial total crack length undetected.

For this reason, both metrics are required. The normalised undetected crack length $C$ evaluates the residual amount of undetected damage, while the Kullback-Leibler divergence $\KL$ evaluates the change in the statistical distribution of the remaining cracks. Used together, they provide a more complete assessment of inspection performance than either metric alone.

\subsection{Results of comparing PoD curves with Monte Carlo simulation}
The quantitative results of our proposed Monte Carlo evaluation are presented visually in Fig.~\ref{fig:C graphs}~and~\ref{fig:KL graphs}, and numerically in Table~\ref{tab:crack_update}. It can be observed that for smaller cracks with $\mu = 37.15$ mm and $\mu = 117.51$ mm, the values of the normalised undetected crack length $C$ are lower for the CV method than for either of the conventional inspection methods, regardless of the $\CoV$ of the prior distribution. 
This is in line with what is shown in Fig.~\ref{fig:PoD, cb, comp.}, where the PoD for the CV method is higher than for Campbell et al.\ and DNVGL when detecting cracks up to approximately 150~mm in length. 
For the prior crack distribution with  $\mu = 371.72$, the normalised crack length $C$ of the CV method lies between those of Campbell et al.\ and DNVGL. 
For short cracks, the CV method also produces a higher $\KL$ divergence, corresponding to a larger change in the shape of the crack length distribution. For long cracks, the visual inspection of Campbell et al.\ instead 
This again repeats the pattern from Fig.~\ref{fig:PoD, cb, comp.}, where for cracks above 150~mm in length, the PoD curve of our CV method is between the PoD curves for manual visual inspection as proposed in \cite{Campbell2019pod, DNVGL_RP_C210_2015}.

\begin{figure*}
\centering
\begin{subfigure}{0.49\textwidth}
    \includegraphics[width=\textwidth]{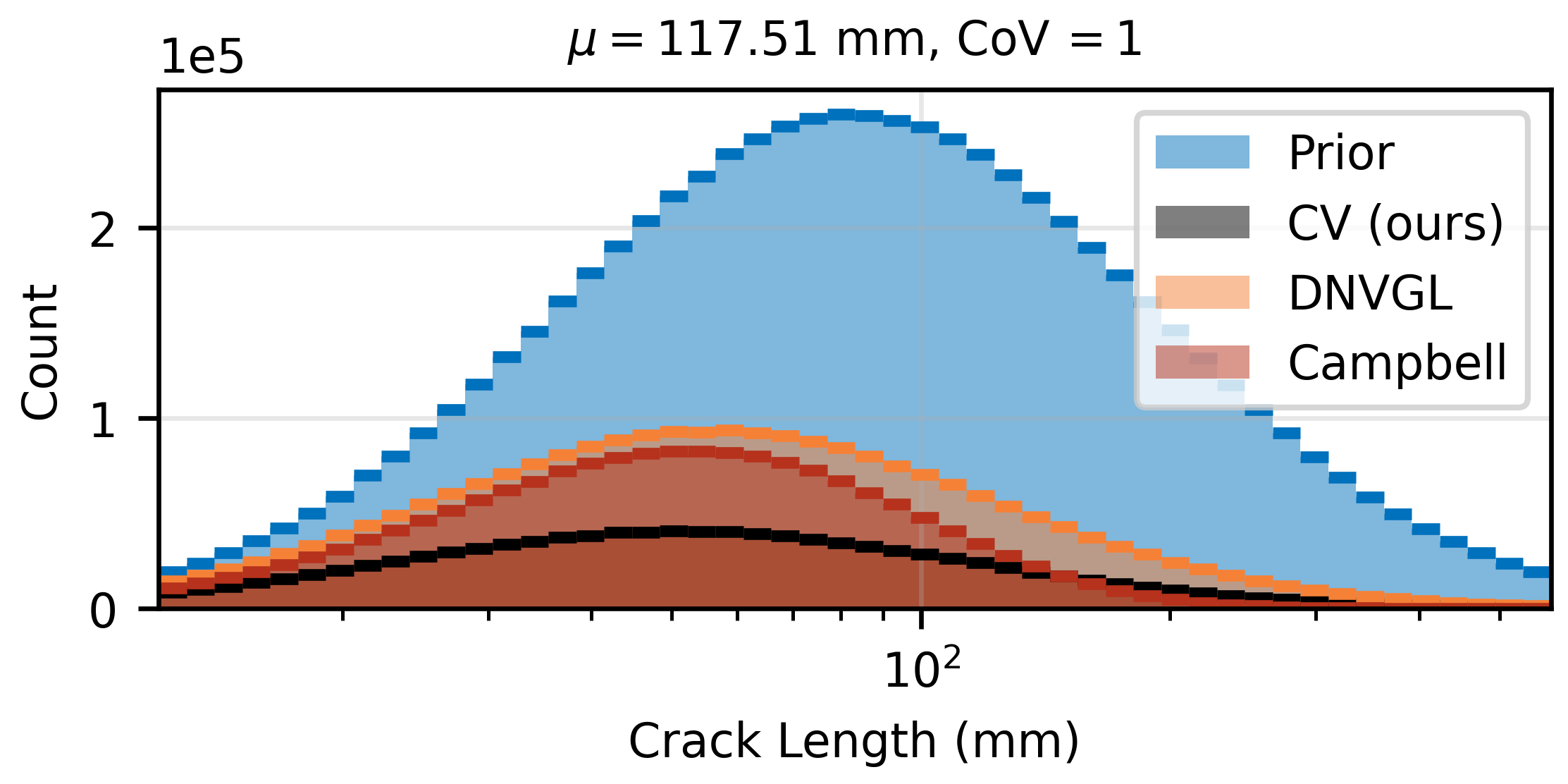}
    \caption{}
    \label{fig:Histograms:mu 117}
\end{subfigure}
\begin{subfigure}{0.49\textwidth}
    \includegraphics[width=\textwidth]{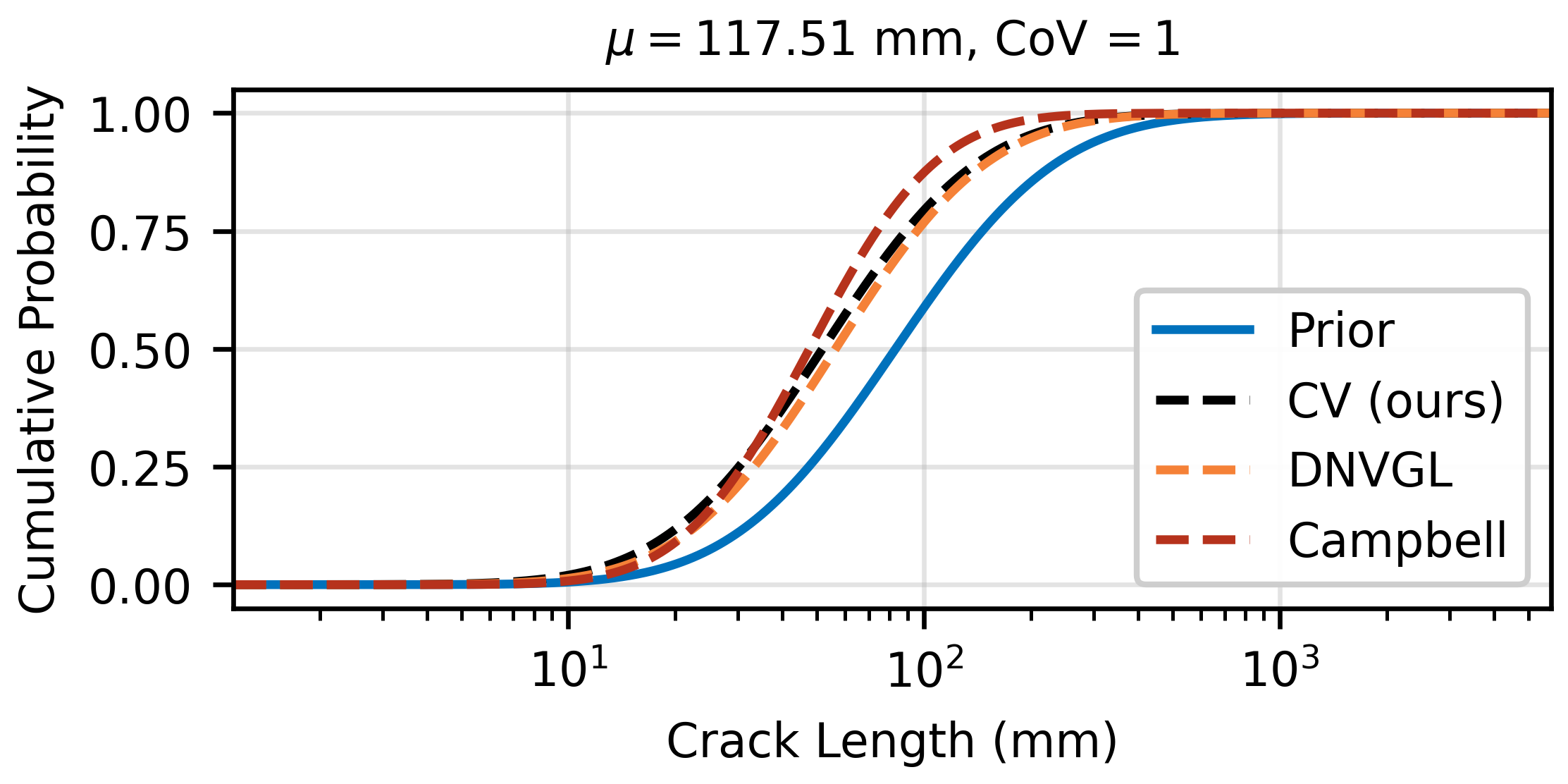}
    \caption{}
    \label{fig:Histograms:cdf mu 117}
\end{subfigure}
\begin{subfigure}{0.49\textwidth}
    \includegraphics[width=\textwidth]{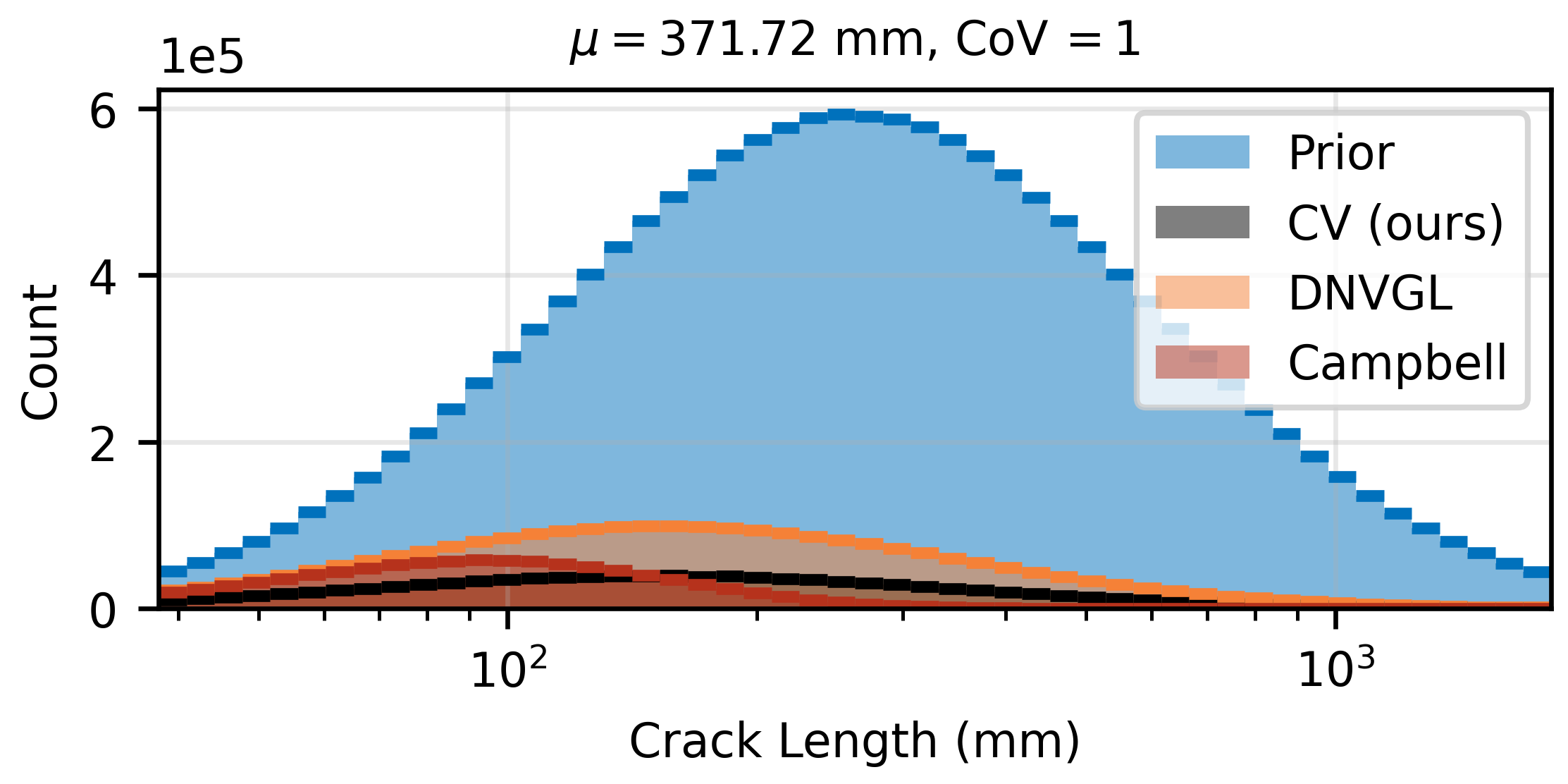}
    \caption{}
    \label{fig:Histograms:mu 371}
\end{subfigure}
\begin{subfigure}{0.49\textwidth}
    \includegraphics[width=\textwidth]{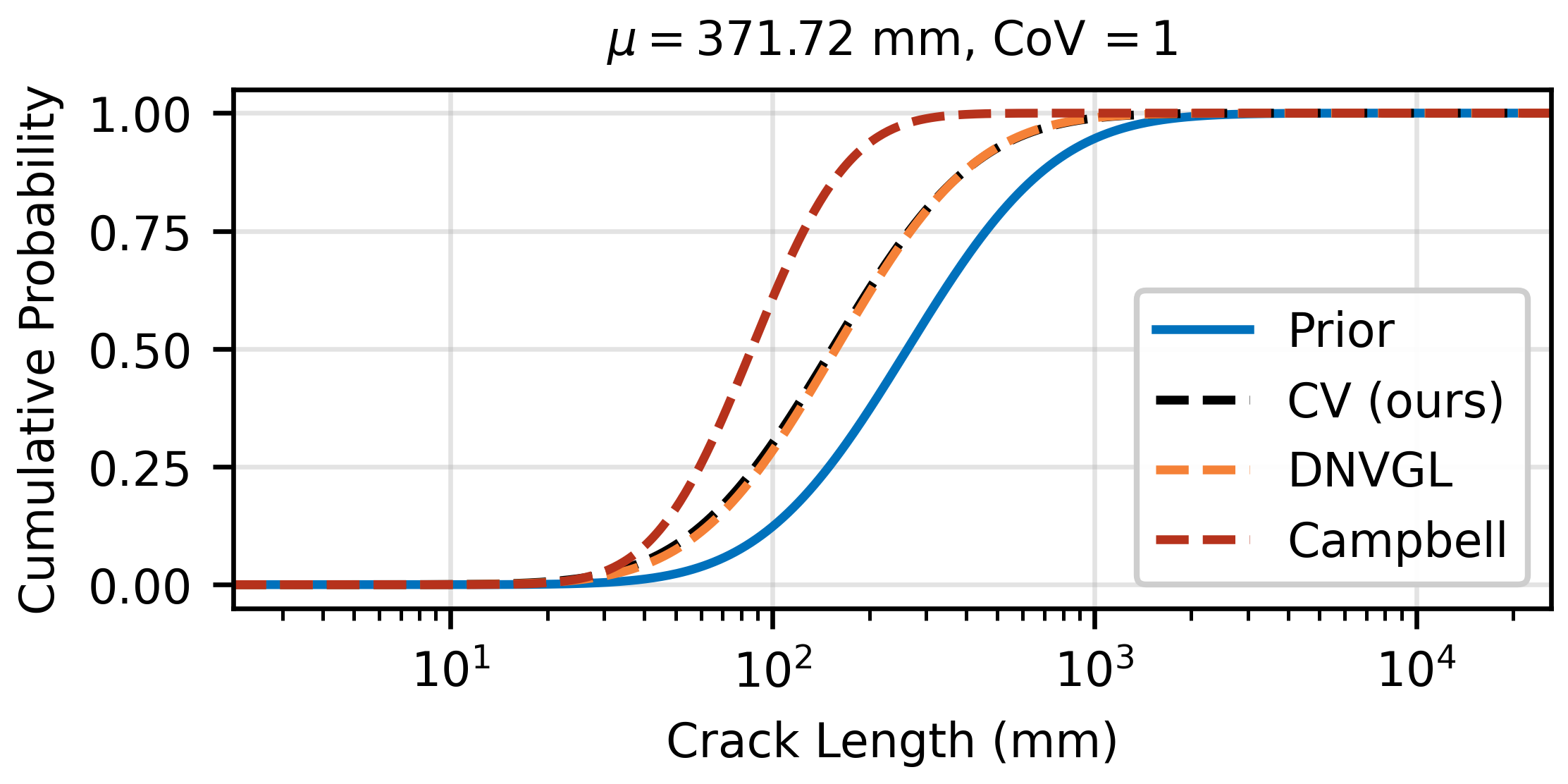}
    \caption{}
    \label{fig:Histograms:cdf mu 371}
\end{subfigure}
\caption{Left: (unnormalised) histograms of prior and posterior crack distributions, obtained using Alg.~\ref{alg:missed_crack_distribution}.
Right: cumulative distributions of crack lengths.
}\label{fig:Histograms}
\end{figure*}

Fig.~\ref{fig:Histograms} further visualises these results using histograms of the posterior crack distribution for a prior with $\mu = 117.51$; $\CoV = 1$ and $\mu = 371.72$; $\CoV = 1$. The use of absolute histogram counts helps explain the need for both metrics. The reduction in counts reflects the amount of cracks that remains undetected, which is measured by $C$, whereas the shift in the distribution reflects the change in the posterior crack-size distribution, which is measured by $\KL$. For instance, the CV method produces lower counts than the DNVGL recommendation, indicating a smaller amount of missed damage. At the same time, its posterior distribution remains closer to the prior, particularly in terms of mean crack length, which can lead to a lower $\KL$. Therefore, the figure shows that $C$ and $\KL$ capture complementary aspects of the inspection outcome. 

Overall, we see that the proposed CV method fairly uniformly outperforms the conventional visual inspection according to the DNVGL baseline, while the inspection method by Campbell et al.\ tends to perform better than our CV method on long cracks. This is in line with the simulation results for the other prior distributions, which can be found in \ref{app:A}.

\begin{table*}
\centering
\caption{Results of computation of normalised undetected crack length $C$ and Kullback-Leibler divergence $\KL$ for our CV method and conventional visual inspection with PoD curves provided by Campbell et al.~\cite{Campbell2019pod} and by the DNVGL recommendation \cite{DNVGL_RP_C210_2015}}
\label{tab:crack_update}
\begin{tabular}{cc|ccc|ccc}\toprule
$\mu$ (mm) & $\CoV$ &  $C_\mathrm{CV}$&  $C_\mathrm{Campbell}$&  $C_\mathrm{DNVGL}$ &  $\KL_{\mathrm{CV}}$& $\KL_{\mathrm{Campbell}}$&$\KL_{\mathrm{DNVGL}}$ \\ \midrule 
\multirow{4}{*}{37.15} & 0.25 & 0.227 & 0.434 & 0.493 & 0.013 & 0.005 & 0.007 \\
 & 0.5 & 0.221 & 0.406 & 0.474 & 0.045 & 0.020 & 0.022 \\
 & 1 & 0.203 & 0.340 & 0.428 & 0.115 & 0.055 & 0.055 \\
 & 2 & 0.173 & 0.256 & 0.355 & 0.189 & 0.080 & 0.083 \\
\hline
\multirow{4}{*}{117.51} & 0.25 & 0.099 & 0.141 & 0.247 & 0.018 & 0.104 & 0.015 \\
 & 0.5 & 0.097 & 0.135 & 0.240 & 0.064 & 0.286 & 0.052 \\
 & 1 & 0.091 & 0.123 & 0.220 & 0.175 & 0.447 & 0.128 \\
 & 2 & 0.081 & 0.103 & 0.186 & 0.315 & 0.413 & 0.198 \\
\hline
\multirow{4}{*}{371.72} & 0.25 & 0.039 & 0.003 & 0.099 & 0.020 & 1.103 & 0.022 \\
 & 0.5 & 0.039 & 0.007 & 0.098 & 0.074 & 2.134 & 0.078 \\
 & 1 & 0.037 & 0.017 & 0.092 & 0.218 & 2.278 & 0.211 \\
 & 2 & 0.034 & 0.024 & 0.082 & 0.435 & 1.591 & 0.364 \\ \bottomrule 
\end{tabular}
\end{table*}

\section{Effect of resolution on PoD}\label{sec:Effect of Resolution on PoD}
To ensure the best inspection result, it is important not only to know the average performance of an inspection method, but also to understand the effect of different inspection variables on that performance. For automated CV crack detection, we expect that the resolution of the crack as visible in an image is the parameter that affects the result of inspections the most. The resolution of the cracks is designated as $r$, measured in pixels per millimetre (pix/mm).

\definecolor{myblueborder}{RGB}{30,90,200}
\definecolor{mybluefill}{RGB}{235,242,255}
\begin{figure*}
\centering
\begin{overpic}[width=\textwidth]{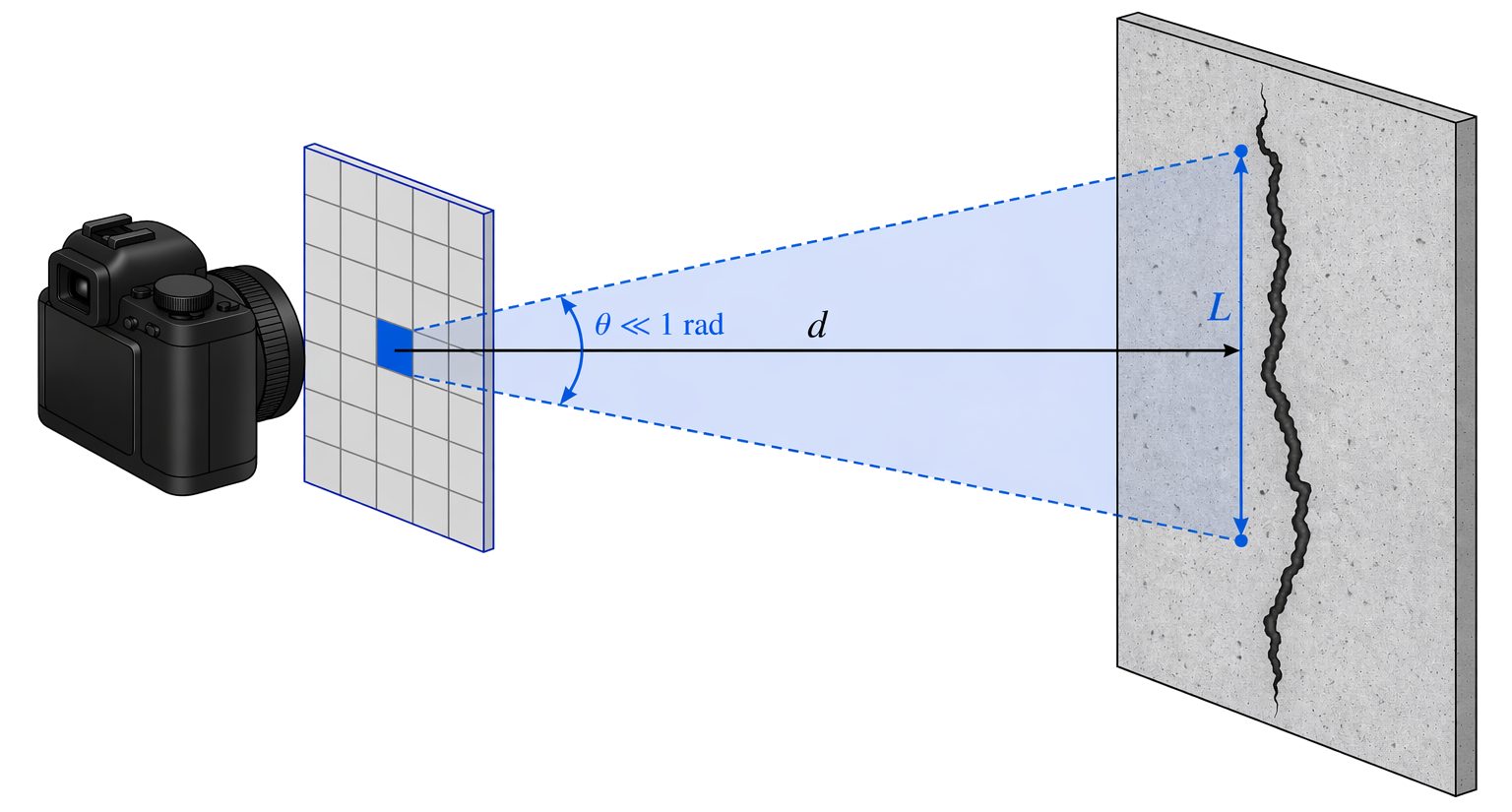}
    \put(30,15){
        \fcolorbox{myblueborder}{mybluefill}{
            \parbox{0.30\textwidth}{
                \centering
                \vspace{0.1mm}
                { $a = 2d\tan\!\left(\frac{\theta}{2}\right)$}\\[1mm]
                
                { $\tan\!\left(\frac{\theta}{2}\right)\approx \frac{\theta}{2},\ \text{if } \theta \ll 1~\mathrm{rad}$}\\[1mm]
                
                { $a \approx d\theta$}\\[1mm]
                
                { $r = \frac{1 \text{ pix}}{a \text{ mm}} \approx \frac{1}{d\theta}\ \mathrm{pix/mm}$}
                \vspace{1mm}
            }
        }
    }
\end{overpic}
\caption{The resolution depends inversely on the distance of the camera to the surface $d$ and the IFOV $\theta$.}
\label{fig:derivation_resolution}
\end{figure*}

\begin{figure*}
    \centering
        \includegraphics[width=\textwidth]{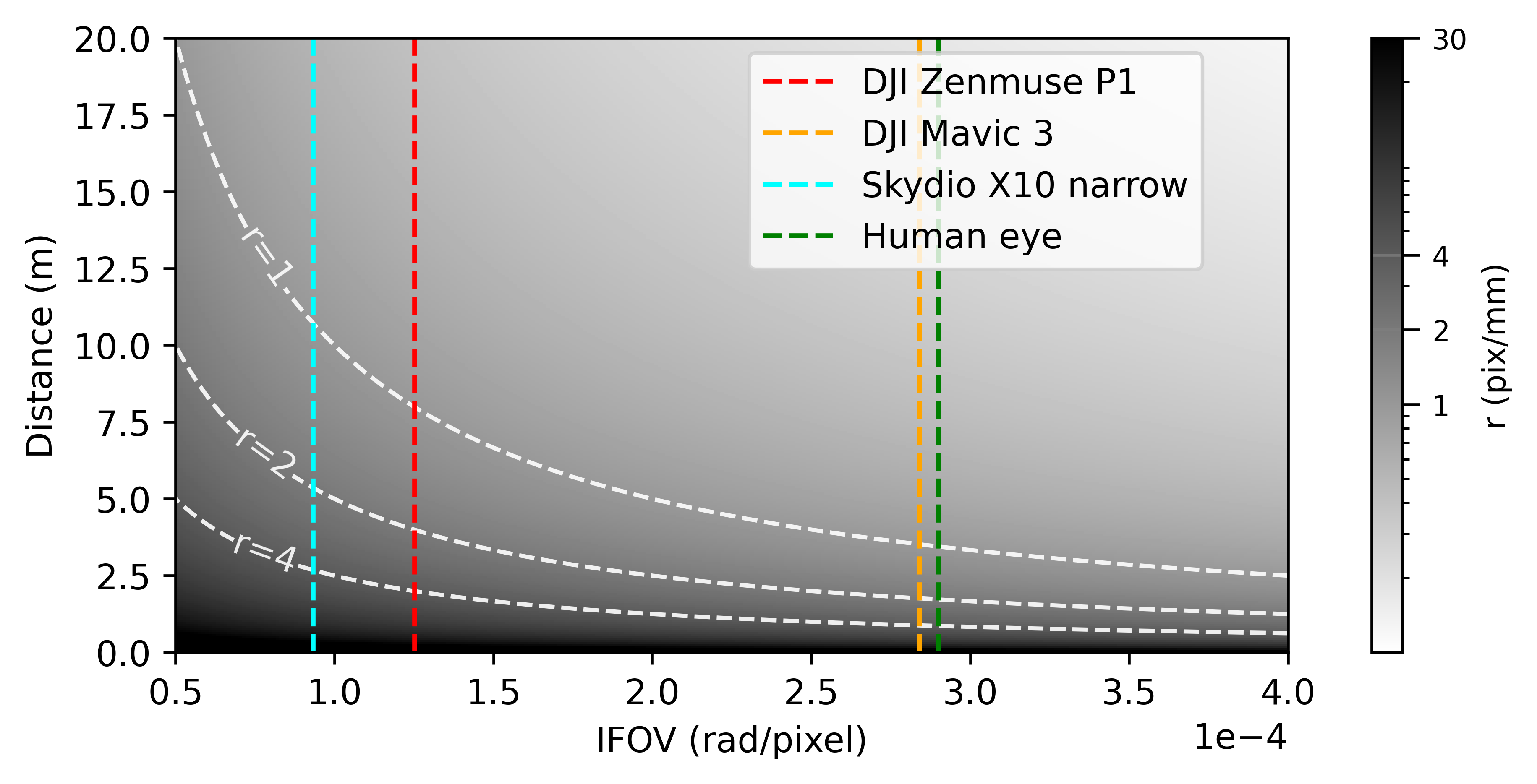}
        \caption{The resolution as a function of the IFOV of the camera and the distance between the camera and the crack. The IFOV of cameras commonly used for inspection of bridges with drones is provided for reference.}
    \label{fig:smooth resolution heatmap}
\end{figure*}

The resolution of the damage depends on two variables: (a) the instantaneous field of view (IFOV) of the camera $\theta$ [rad/pixel], which defines the field of view of a single pixel on the camera, and (b) the distance between the camera and the damage $d$ [mm].
As shown in Fig. \ref{fig:derivation_resolution}, it is possible to derive the relation between the resolution of the damage $r$ and the IFOV $\theta$ and the distance $d$:
\begin{equation}\label{sec:Resolution}
r = \frac{1}{\theta \cdot d}.
\end{equation}
To better illustrate the dependency of $r$ on $\theta$ and $d$, we show the resolution as a heatmap in Fig. \ref{fig:smooth resolution heatmap}, where the three vertical lines indicate the IFOV of three cameras available on typical drones typically used for bridge inspection: a DJI Mavic 3 drone with inbuilt camera; a DJI Matrice drone with a DJI Zenmuse P1 camera; a Skydio drone X10 with 64 MP narrow camera. For comparison, a rough approximation of the IFOV of a human eye is also indicated in Fig. \ref{fig:smooth resolution heatmap} based on \cite{kaiser2009IFOVHumaneye}.

\subsection{Methodology for including resolution in PoD}\label{sec:method_resolution_pod}
We propose two methods to evaluate the effect of the resolution on the PoD: the first is based on PoD curves fitting hit/miss data which are binned depending on the resolution of the images, and the second makes use of a parametric PoD surface model. 

\subsubsection{Binning approach}\label{sec:ResolutionEffect of Resolution on PoD:Resolution effect with binning}
To evaluate the effect of resolution on the PoD using a binning approach, we divide the images from the PoD estimation dataset into four resolution bins, with ranges of 0-1 pix/mm; 1-2 pix/mm; 2-4 pix/mm; 4-30 pix/mm. The boundaries between the bins are illustrated in Fig.~\ref{fig:smooth resolution heatmap}.
Next, we fit the PoD curve model from Eq.~\eqref{eq:pod_length} for each resolution bin, using inspection outcomes on images only from that specific resolution bin. We assume that the resulting PoD curve for each bin is applicable for a crack resolution in the middle of each bin.
According to \cite{ENIQ_POD_BestPractices_2011}, proper PoD estimation requires at least 60 samples; using only the PoD estimation dataset results in bins with too few samples. To ensure that at least 60 samples are available in each bin, we artificially increase the number of images by downsampling each of the 239 images from the PoD estimation dataset twice: by a factor of 2 and by a factor of 4. The downsampled images have the same physical length, but a smaller digital length, so that the resolution of each crack on the downsampled images is reduced. This is equivalent to taking a picture with a camera with less resolution. 

\subsubsection{Parametric approach}\label{sec:Effect of Resolution on PoD:pod surface}
We also introduce a parametric approach to explictly include the influence of resolution in the formulation of the PoD. 
This parametric approach relates to the binning approach from Sec.~\ref{sec:ResolutionEffect of Resolution on PoD:Resolution effect with binning} in the same way that the crack length PoD model in Eq.~\eqref{eq:pod_length} relates to the classical binomial model \cite{ENIQ_POD_BestPractices_2011}.
Even in the case of a single covariate (only crack length), the binomial method has numerous deficiencies. First and foremost, it requires an arbitrary choice of bins, and this selection can have a large influence, e.g.\ on the confidence bounds. Additionally, the binomial model is inefficient with data since the data points only influence the PoD curve within their own bins. For these reasons, it has been generally recommended to forgo the binomial model in favour of parametric models \cite{ENIQ_POD_BestPractices_2011}. To analyse the influence of resolution on PoD, we therefore extend our logistic model described in Sec.~\ref{sec: Probability of detection}, by adding the resolution $r$ in pix/mm as an additional independent variable:
\begin{equation}\label{eq:logit_length_resolution}
\log\left(\frac{p}{1 - p}\right) = \beta_0 + \beta_1 h_a(a) + \beta_2 h_r(r),
\end{equation}
with $h_r(r) = \log(r)$ or $h_r(r) = r$, so that
\begin{equation}\label{eq:pod_length_resolution}
\PoD(a, r; \beta_0, \beta_1, \beta_2) = \frac{\exp(\beta_0 + \beta_1 h_a(a) + \beta_2 h_r(r))}{1 + \exp(\beta_0 + \beta_1 h_a(a) + \beta_2 h_r(r))}.
\end{equation}
Hence, instead of a PoD curve, the model in Eq.~\eqref{eq:pod_length_resolution} gives a PoD surface, parametrised by the length of the crack $a$ and the resolution $r$.
Therefore, the dataset is now extended to include the resolution of the image:
\begin{equation}\label{eq:samples_resolution}
\mathcal{S} \coloneqq \{(a_i, r_i, Y_i)\}_{i = 1}^n.
\end{equation}
Note that the PoD curve model in Eq.~\eqref{eq:pod_length} is included in the PoD surface model in Eq.~\eqref{eq:pod_length_resolution} by setting $\beta_2 = 0$. The PoD surface model therefore generalises the PoD curve model, enabling quantitative comparison of both models.
The model parameters $\beta_0$, $\beta_1$, and $\beta_2$ are estimated using the maximum likelihood method. 
The log-likelihood is a simple modification of Eq.~\eqref{eq:log-likelihood}:
\ifreview
\begin{equation}\label{eq:log_likelihood_resolution}
\ell(\beta_0, \beta_1, \beta_2; \mathcal{S}) = \sum_{i = 1}^n \Big[Y_i (\beta_0 + \beta_1 h_a(a_i) + \beta_2 h_r(r_i)) - \log(1 + \exp(\beta_0 + \beta_1 h_a(a_i) + \beta_2 h_r(r_i))) \Big].
\end{equation}
\else
\begin{equation}\label{eq:log_likelihood_resolution}
\begin{split}
& \ell(\beta_0, \beta_1, \beta_2; \mathcal{S}) \\
& \quad = \sum_{i = 1}^n \Big[Y_i (\beta_0 + \beta_1 h_a(a_i) + \beta_2 h_r(r_i)) \\
& \qquad - \log(1 + \exp(\beta_0 + \beta_1 h_a(a_i) + \beta_2 h_r(r_i))) \Big].
\end{split}
\end{equation}
\fi

\begin{figure*}
\centering
\includegraphics[width=0.8\textwidth]{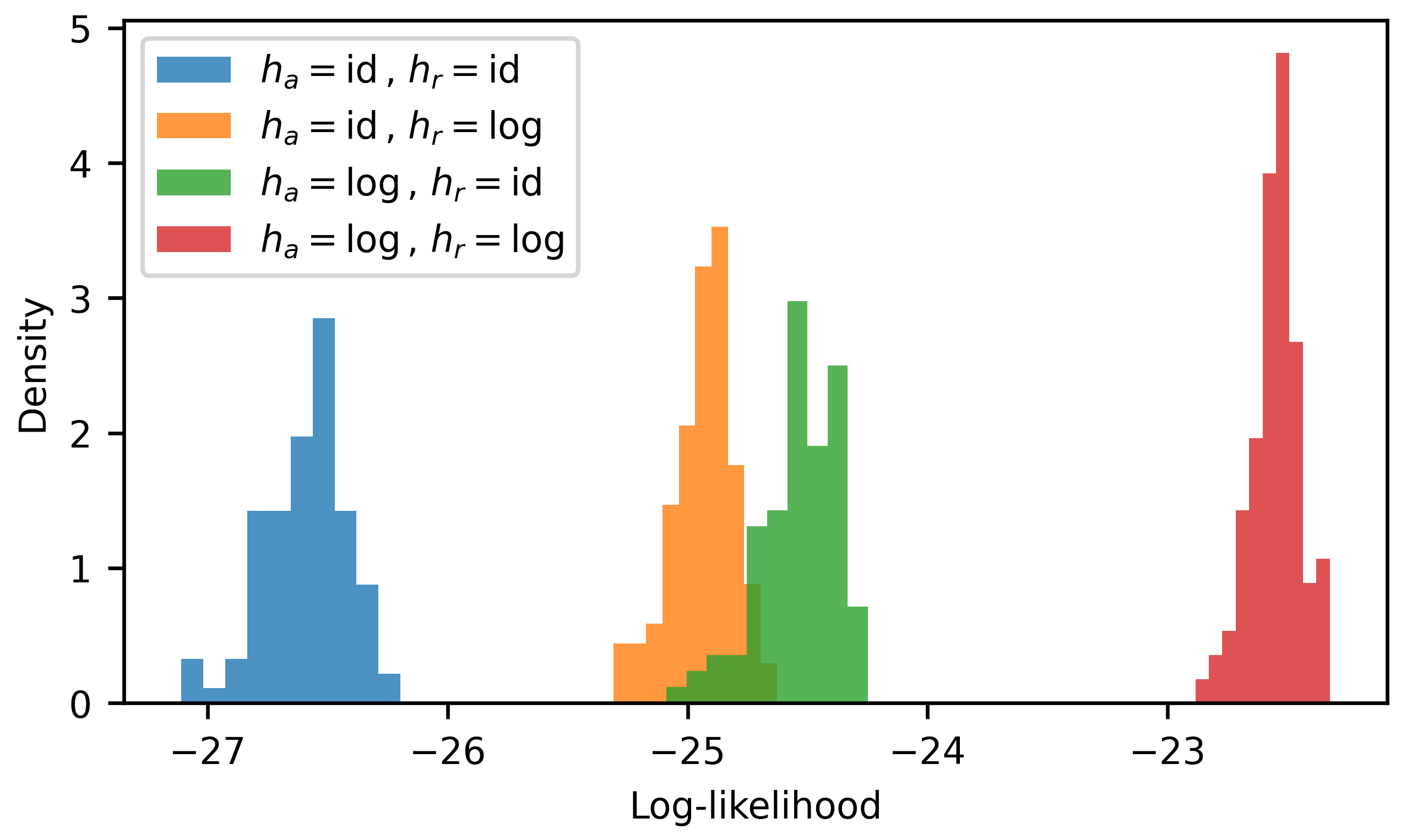}
\caption{Distribution of log-likelihoods on validation sets determined using cross-validation for different choices of transformations $h_a$ and $h_r$.
}\label{fig:selection_transforms_length_resolution}
\end{figure*}
Transformations $h_a$ and $h_r$ are selected using cross-validation.
First, 100 random shuffles of the data are produced. Subsequently, each shuffle is divided into 10 folds. For each of the folds, the logistic model in Eq.~\eqref{eq:pod_length_resolution} is fitted to the remaining 9 of the 10 folds, and the log-likelihood is calculated on the left-out fold. This is repeated for each combination of $h_a = \log, \operatorname{id}$ and $h_r = \log, \operatorname{id}$, producing the empirical distributions shown in Fig.~\ref{fig:selection_transforms_length_resolution}. Since $h_a = \log = h_r$ clearly produces the highest log-likelihood, we will continue with this choice of transformations.

Next, the model is fitted on all the data and confidence regions are computed using the (asymptotic) statistics of the maximum likelihood estimators. Specifically, under the assumption that the data is generated by our model in Eq.~\eqref{eq:pod_length_resolution} with some parameters $\beta_0, \beta_1, \beta_2$, Wilk's likelihood ratio statistic, given by
\ifreview
\begin{equation}\label{eq:wilks_statistic_length_resolution}
W(\beta_0, \beta_1, \beta_2; \mathcal{S}) \coloneqq -2 (\ell(\beta_0, \beta_1, \beta_2; \mathcal{S}) - \ell(\hat{\beta}_0, \hat{\beta}_1, \hat{\beta}_2; \mathcal{S})),
\end{equation}
\else
\begin{multline}\label{eq:wilks_statistic_length_resolution}
W(\beta_0, \beta_1, \beta_2; \mathcal{S}) \\ \coloneqq -2 (\ell(\beta_0, \beta_1, \beta_2; \mathcal{S}) - \ell(\hat{\beta}_0, \hat{\beta}_1, \hat{\beta}_2; \mathcal{S})),
\end{multline}
\fi
with $\hat{\beta}_0, \hat{\beta}_1, \hat{\beta}_2$ the maximum likelihood estimates, asymptotically follows a $\chi$-squared distribution with three degrees of freedom, since our model has three free parameters. 
Then we define our 95~\% confidence region as $B(\mathcal{S}) \coloneqq \{(\beta_0, \beta_1, \beta_2) \in \R^3 \mid W(\beta_0, \beta_1, \beta_2) \leq \chi_{3, 0.95}^2 \approx 7.815\}$. In words this means that $(\beta_0, \beta_1, \beta_2)$ is included in $B(\mathcal{S})$ if the probability of observing such a large value of Wilk's statistic is at least 5~\%, under the hypothesis that the data is distributed according to Eq.~\eqref{eq:pod_length_resolution} with those parameters.

Since the model now produces a PoD surface instead of a PoD curve, we take constant resolution cross-sections to visualise this surface in Fig.~\ref{fig:pod_length_resolution}. Likewise, the confidence bounds are found by taking constant resolution cross-sections of the envelope of PoD surfaces generated by $B(\mathcal{S})$:
\ifreview
\begin{equation}\label{eq:confidence_bounds_length_resolution}
\begin{aligned}
\PoD^{\textrm{upper}}(a, r; \mathcal{S}) & = \max_{(\beta_0, \beta_1, \beta_2) \in B(\mathcal{S})} \frac{\exp(\beta_0 + \beta_1 h_a(a) + \beta_2 h_r(r))}{1 + \exp(\beta_0 + \beta_1 h_a(a) + \beta_2 h_r(r))}, \textrm{ and} \\
\PoD^{\textrm{lower}}(a, r; \mathcal{S}) & = \min_{(\beta_0, \beta_1, \beta_2) \in B(\mathcal{S})} \frac{\exp(\beta_0 + \beta_1 h_a(a) + \beta_2 h_r(r))}{1 + \exp(\beta_0 + \beta_1 h_a(a) + \beta_2 h_r(r))}.
\end{aligned}
\end{equation}
\else
\begin{equation}\label{eq:confidence_bounds_length_resolution}
\begin{aligned}
& \PoD^{\textrm{upper}}(a, r; \mathcal{S}) \\
& \quad = \max_{(\beta_0, \beta_1, \beta_2) \in B(\mathcal{S})} \frac{\exp(\beta_0 + \beta_1 h_a(a) + \beta_2 h_r(r))}{1 + \exp(\beta_0 + \beta_1 h_a(a) + \beta_2 h_r(r))}, \textrm{ and} \\
& \PoD^{\textrm{lower}}(a, r; \mathcal{S}) \\
& \quad = \min_{(\beta_0, \beta_1, \beta_2) \in B(\mathcal{S})} \frac{\exp(\beta_0 + \beta_1 h_a(a) + \beta_2 h_r(r))}{1 + \exp(\beta_0 + \beta_1 h_a(a) + \beta_2 h_r(r))}.
\end{aligned}
\end{equation}
\fi

Finally, we want to assess whether extending the PoD model to include the influence of resolution in addition to crack length is a sensible choice. We define the following hypotheses:
\begin{itemize}
\item[\hypertarget{h0}{$H_0$}:] The probability of detection for a crack of length $a$ at resolution $r$ is given by the PoD model in Eq.~\eqref{eq:pod_length_resolution} with parameters $\beta_0, \beta_1, 0$, for some $\beta_0, \beta_1 \in \R$.
\item[\hypertarget{ha}{$H_a$}:] The probability of detection for a crack of length $a$ at resolution $r$ is given by the PoD model in Eq.~\eqref{eq:pod_length_resolution} with parameters $\beta_0, \beta_1, \beta_2$, for some $\beta_0, \beta_1, \beta_2 \in \R$.
\end{itemize}

We define the log-likelihood ratio statistic
\begin{equation}\label{eq:statistic_length_resolution_vs_length}
\Lambda(\mathcal{S}) \coloneqq -2 (\ell(\hat{\beta}_0, \hat{\beta}_1, 0; \mathcal{S}) - \ell(\tilde{\beta}_0, \tilde{\beta}_1, \tilde{\beta}_2; \mathcal{S})),
\end{equation}
with $\hat{\beta}_0, \hat{\beta}_1$ the maximum likelihood estimates for the parameters in the model which includes only the crack length and $\tilde{\beta}_0, \tilde{\beta}_1, \tilde{\beta}_2$ the maximum likelihood estimates in the model that includes both the crack length and the resolution. Then, under the null-hypothesis $H_0$, $\Lambda(\mathcal{S})$ follows a $\chi$-squared distribution with one degree of freedom, since the extended model has one free parameter more than the null model. We reject the null-hypothesis at a 5~\% significance level if $\Lambda(\mathcal{S}) > \chi_{1, 0.95}^2 \approx 3.841$, i.e.\ if the probability under the null-hypothesis of observing a statistic at least so large is less than 5~\% for the considered dataset.

\subsection{Results of including resolution in PoD}\label{sec:results_resolution_pod}
\subsubsection{Binning approach}
\begin{figure*}
    \centering
    \includegraphics[width=\textwidth]{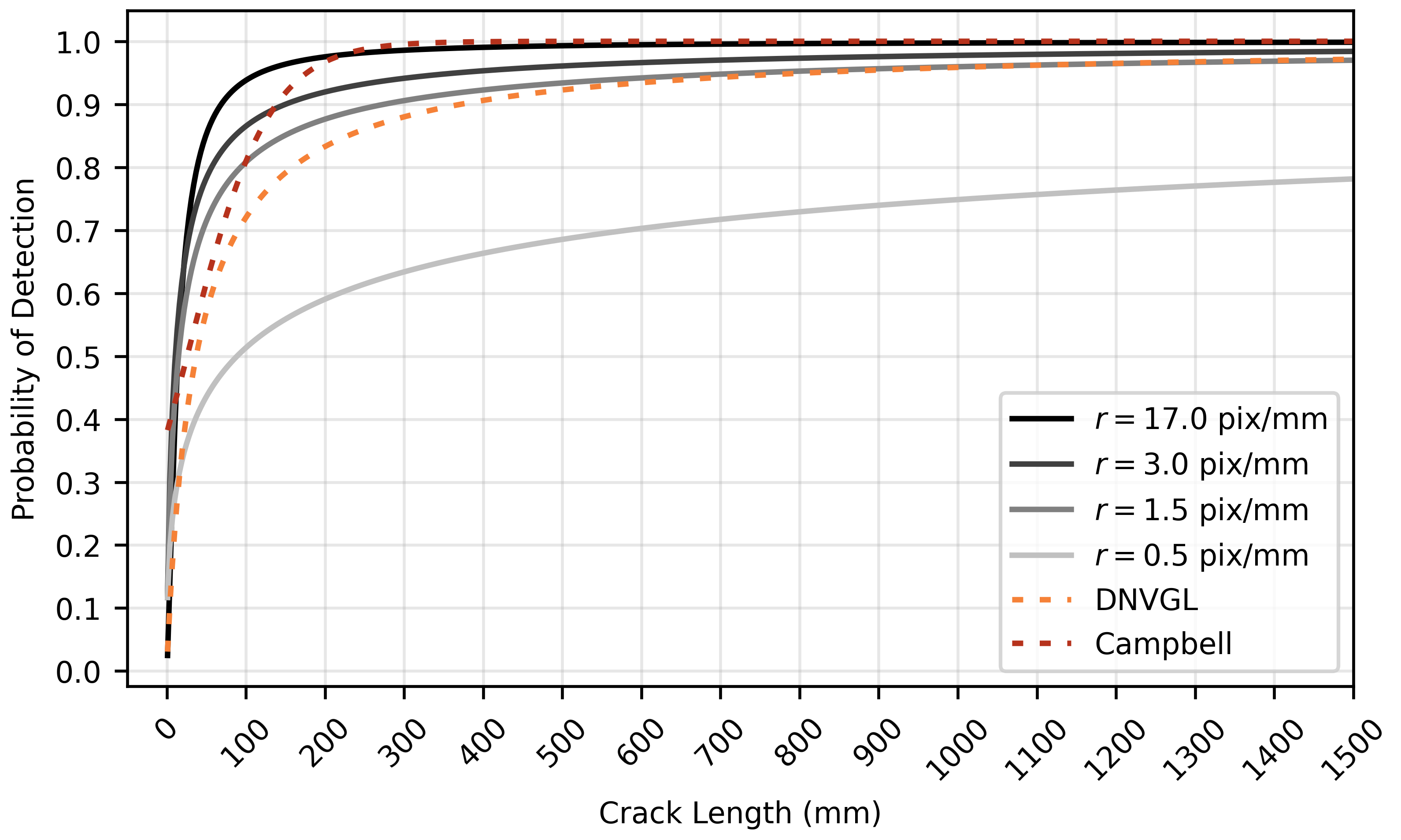}
    \caption{PoD curves calculated for 4 resolution bins, compared to the baselines of Campbell et al.~\cite{Campbell2019pod} and DNVGL \cite{DNVGL_RP_C210_2015}. Note that confidence bounds (Eq.~\eqref{eq:confidence_bounds_length}) are omitted for visual clarity.
    }
    \label{fig:Binned PoD}
\end{figure*}

\begin{table}
\centering
\caption{Parameters of the PoD curves fitted on each resolution bin}
\label{tab:binned_resolution_parameters}
\begin{tabular}{c|c|c}\toprule
$r$ (pix/mm)    & $\beta_0$ & $\beta_1$ \\ \midrule 
0-1             & $-2.022$  & $0.451$ \\
1-2             & $-2.024$  & $0.752$ \\
2-4             & $-1.977$  & $0.833$ \\
4-30            & $-3.673$  & $1.387$ \\ \bottomrule 
\end{tabular}
\end{table}
The PoD curves are shown in Fig.~\ref{fig:Binned PoD}, where we denote the bins by their central resolution, e.g.\ $r = 17$ pix/mm corresponds to the 4-30 pix/mm bin.
The confidence bounds, which can be computed as explained in Sec.~\ref{sec: Probability of detection}, have been omitted for visual clarity.\footnote{The confidence bounds, which do not have simple parametric forms, are available in the accompanying code \cite{Sherry2026PoDCurveComparison}.\label{ftn:confidence_bounds}}
% \url{https://github.com/finnsherry/PoDCurveComparison}
The PoD curve estimated on images of all resolutions, given by Eq.~\eqref{eq:pod_length_fitted}, aligns with the PoD curve for the 2-4 pix/mm bin. Fig.~\ref{fig:Binned PoD} additionally shows the PoD curves from Campbell et al.\ (Eq.~\eqref{eq:pod_Campbell}) \cite{Campbell2019pod} and the DNVGL recommmendation (Eq.~\eqref{eq:pod_dnvgl}) \cite{DNVGL_RP_C210_2015}. 
As one might expect, the detection capabilities of our CV method improve as resolution increases. For moderate resolutions $r > 1.5$ pix/mm, the probability of detection is above the PoD curve from the DNVGL recommendation over all crack lengths, and above the PoD curve estimated by Campbell et al.\ for cracks up to about 100 mm in length. 

\subsubsection{Parametric approach}
\begin{figure*}
\centering
\includegraphics[width=\textwidth]{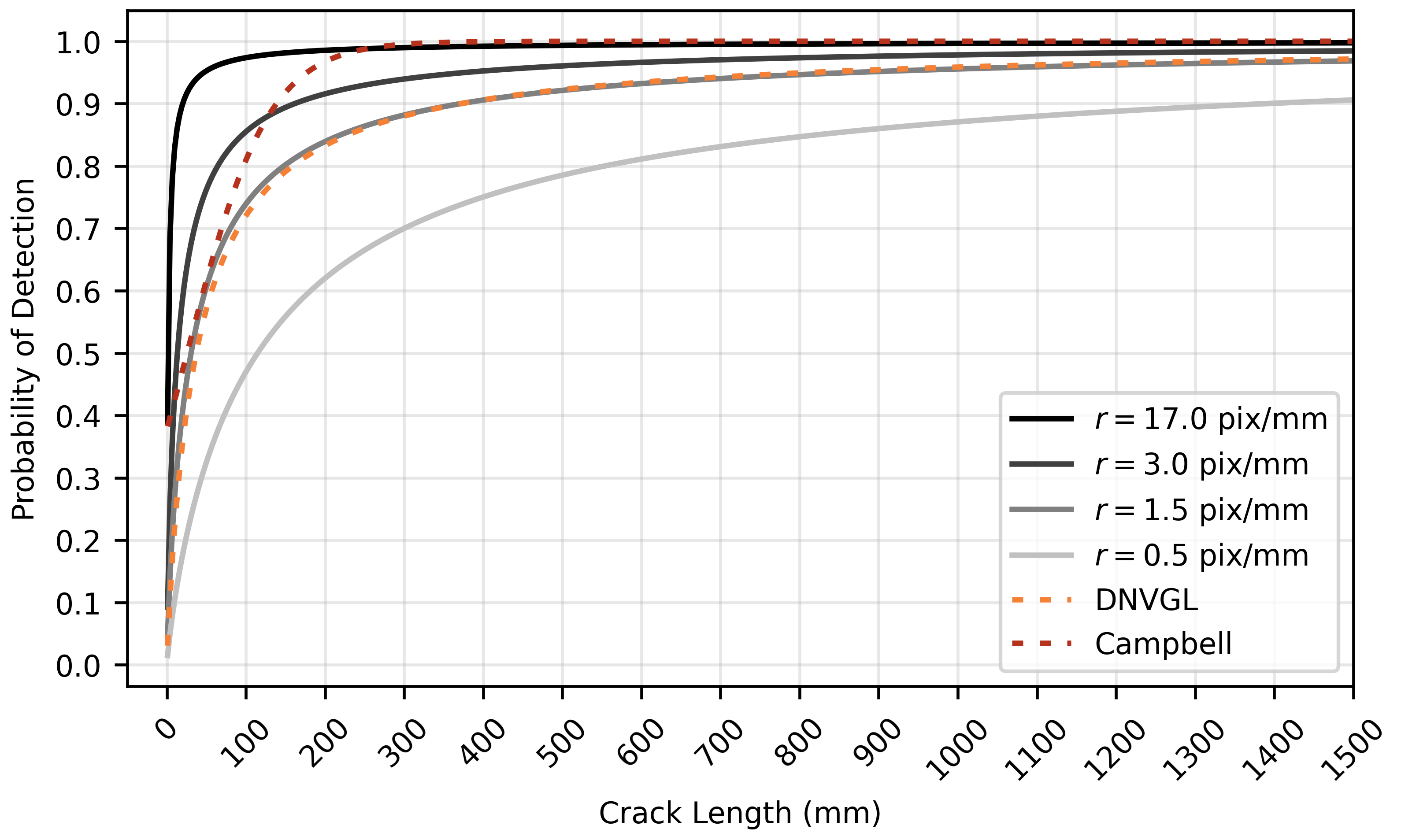}
\caption{Cross-sections of the PoD surface (Eq.~\eqref{eq:pod_length_resolution}) at 4 fixed resolutions (solid lines), compared to the baselines of Campbell et al.~\cite{Campbell2019pod} and DNVGL \cite{DNVGL_RP_C210_2015}. Note that confidence bounds (Eq.~\eqref{eq:confidence_bounds_length_resolution}) are omitted for visual clarity.}
\label{fig:pod_length_resolution}
\end{figure*}

We fitted the PoD surface as explained in Sec.~\ref{sec:Effect of Resolution on PoD:pod surface}. The maximum likelihood estimates of the parameters are $\beta_0 = -3.446$, $\beta_1 = 0.881$, and $\beta_2 = 1.058$, so that 
\ifreview
\begin{equation}\label{eq:pod_length_resolution_fitted}
\PoD_\textrm{CV}(a, r) = \frac{\exp(-3.446 + 0.881 \log(a) + 1.058 \log(r))}{1 + \exp(-3.446 + 0.881 \log(a) + 1.058 \log(r))}.
\end{equation}
\else
\begin{multline}\label{eq:pod_length_resolution_fitted}
\PoD_\textrm{CV}(a, r) \\ = \frac{\exp(-3.446 + 0.881 \log(a) + 1.058 \log(r))}{1 + \exp(-3.446 + 0.881 \log(a) + 1.058 \log(r))}.
\end{multline}
\fi
Fig.~\ref{fig:pod_length_resolution} shows the cross-sections for resolution $r = 0.5, 1.5, 3, 17$ pix/mm, compared to the PoD curve from the DNVGL recommendation \cite{DNVGL_RP_C210_2015} and the PoD curve estimated by Campbell et al. \cite{Campbell2019pod}. 
The confidence bounds, which can be computed as explained in Sec.~\ref{sec:Effect of Resolution on PoD:pod surface}, have again been omitted for visual clarity.\footref{ftn:confidence_bounds}
In line with the results from the binning approach (cf.~Fig.~\ref{fig:Binned PoD}), the probability of detection increases not only as the crack becomes longer, but also as the resolution of the image improves. 

We also use hypothesis testing to assess whether extending the PoD curve model to include the effect of resolution is supported by data. In our data, we observe that the log-likelihood ratio statistic, defined in Eq.~\eqref{eq:statistic_length_resolution_vs_length}, is $\Lambda(\mathcal{S}) = 69.66 > 3.84 \approx \chi_{1, 0.95}^2$, so that we reject the null-hypothesis \hyperlink{h0}{$H_0$} in favour of the alternative hypothesis \hyperlink{ha}{$H_a$}. In other words, we have found the improvement of the PoD surface model in Eq.~\eqref{eq:pod_length_resolution} compared to the PoD curve model in Eq.~\eqref{eq:pod_length} to be statistically significant at a 5~\% significance level for the considered dataset.

\subsubsection{Comparison of approaches}
\begin{figure*}
\centering
\includegraphics[width=\textwidth]{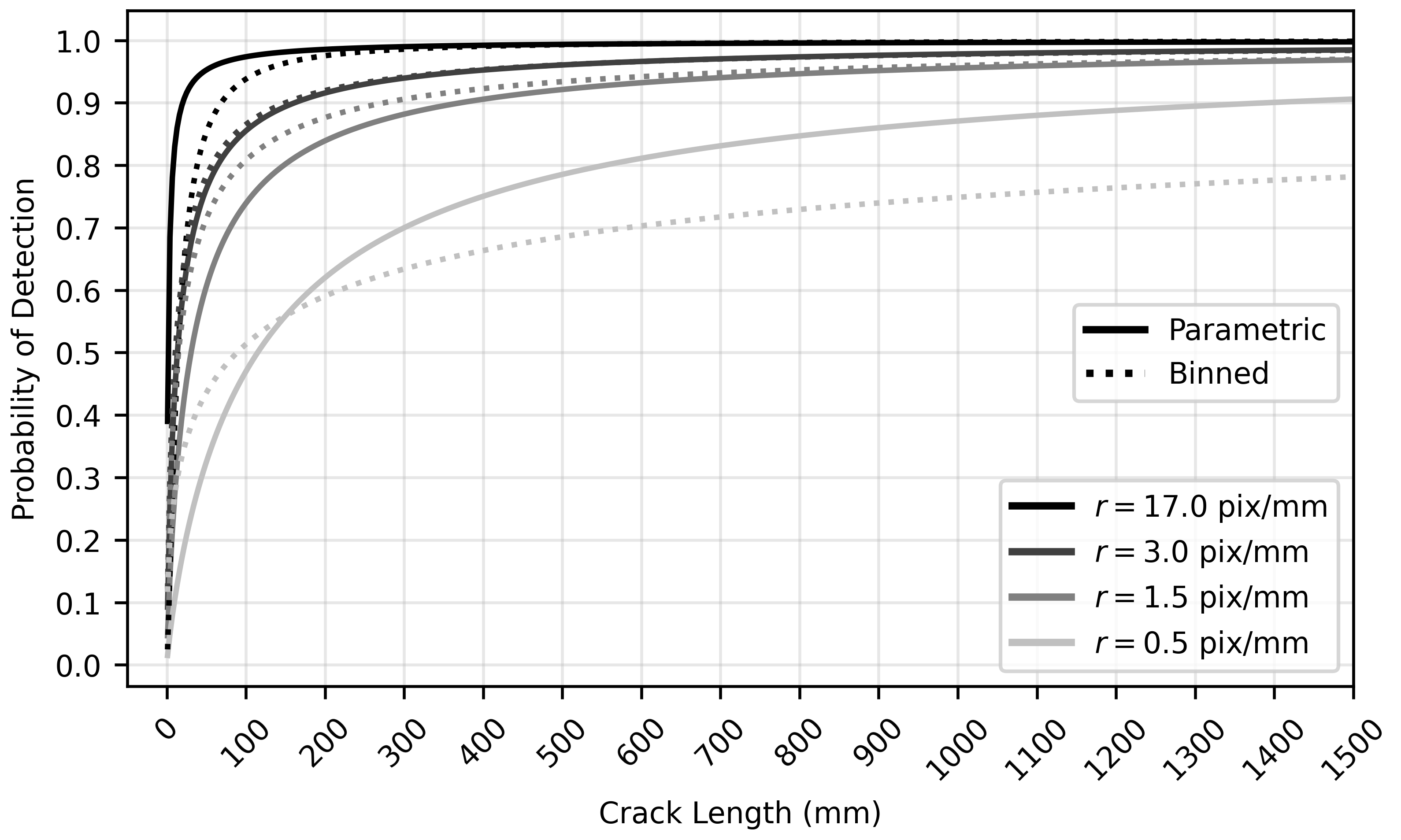}
\caption{Fixed resolution cross-sections of the PoD surface (Eq.~\eqref{eq:pod_length_resolution}) (solid lines), and the PoD curves fitted on each resolution bin (dotted lines).}
\label{fig:comparison_resolution_methods}
\end{figure*}

We now compare the binning approach to the parametric approach. In Fig.~\ref{fig:comparison_resolution_methods}, the dotted lines give the PoD curves found using the binning method, and the solid lines give cross-sections of the parametric PoD surface, with resolution in the middle of each of the bins. 
We quantify the performance of the two approaches by computing the mean squared error (MSE), where the error is simply the difference between the observed outcome (hit $= 1$, miss $= 0$) and the probability of detection. We find that the binned approach gives an MSE of $0.097$, while the parametric approach has an MSE of $0.101$.

It is possible that the reduction in error of the binned approach compared to the parametric approach is due to the greater flexibility afforded by the binned approach.
It must be noted, however, that this flexibility comes at the cost of the effective addition of 5 parameters, since it uses 4 PoD curves with 2 parameters each instead of 1 PoD surface with 3 parameters. These additional parameters can harm the robustness and generalisation of the model. Since the error reduction is minor ($\sim 4~\%$), we therefore perform the remaining experiments with the PoD surface model.

\subsection{Numerical evaluation of resolution dependence of crack detection performance}\label{sec:Effect of Resolution on PoD:Numerical comparison of PoD curves}
To further study the effect of the resolution on the inspection performance, we utilise the approach and the metrics proposed in Sec.~\ref{sec:Effect on crack length update after inspection}.
We take constant resolution, $r = 0.5, 1.5, 3, 17$ pix/mm, cross-sections of the PoD surface given in Eq.~\eqref{eq:pod_length_resolution_fitted}. Such a cross-section can be interpreted to represent the PoD curve of the automated CV inspection method with a fixed camera and distance to the crack surface. We also again include the PoD curves from the DNVGL recommendation (Eq.~\eqref{eq:pod_dnvgl}) \cite{DNVGL_RP_C210_2015} and determined by Campbell et al.\ (Eq.~\eqref{eq:pod_Campbell}) \cite{Campbell2019pod}.

Tables \ref{tab:crack_update_C} and \ref{tab:crack_update_KL} summarise how the normalised undetected crack length $C$ and the Kullback-Leibler divergence $\KL$, given by Eqs.~\eqref{eq: normalised total crack length} and \eqref{eq:KL}, respectively, depend on the prior mean $\mu$ and coefficient of variation $\CoV$. The same results are shown visually in Figs. \ref{fig:C_resolution} and \ref{fig:KL_resolution}.
A clear trend can be seen that with the improvement of resolution, the PoD performance improves. 
Furthermore, the results show that for low crack resolution up to $r = 1.5$ pix/mm, which corresponds to inspection with the camera having IFOV $\theta = 1.25 \times 10^{-4}$ rad/pix from a distance of 7 metres, the performance of our CV crack detection method is consistently worse than conventional visual inspection, regardless of the prior crack distribution. We can also conclude that for our CV method to outperform the conventional inspection the resolution of the images collected during inspection should be more than 3 pix/mm, which corresponds to distances closer than 3 metres for camera with IFOV $\theta = 1.25 \times 10^{-4}$ rad/pix. 

Note that the comparison framework could easily be extended to also take into account resolution by additionally sampling the resolution from some prior distribution. However, since the PoD curves by Campbell et al.~\cite{Campbell2019pod} and from the DNVGL recommendation \cite{DNVGL_RP_C210_2015} do not include resolution, this would not be useful for this experiment.

\definecolor{cCVr1}{rgb}{0, 0, 0}
\definecolor{cCVr2}{rgb}{0.25, 0.25, 0.25}
\definecolor{cCVr3}{rgb}{0.5, 0.5, 0.5}
\definecolor{cCVr4}{rgb}{0.75, 0.75, 0.75}
\definecolor{cCampbell}{rgb}{1, 0, 0}
\definecolor{cDNVGL}{rgb}{1.0, 0.5, 0.0}
\pgfplotsset{every axis plot/.append style={very thick}}
\pgfplotstableread{Figures/update_resolution_C.dat}\datatable

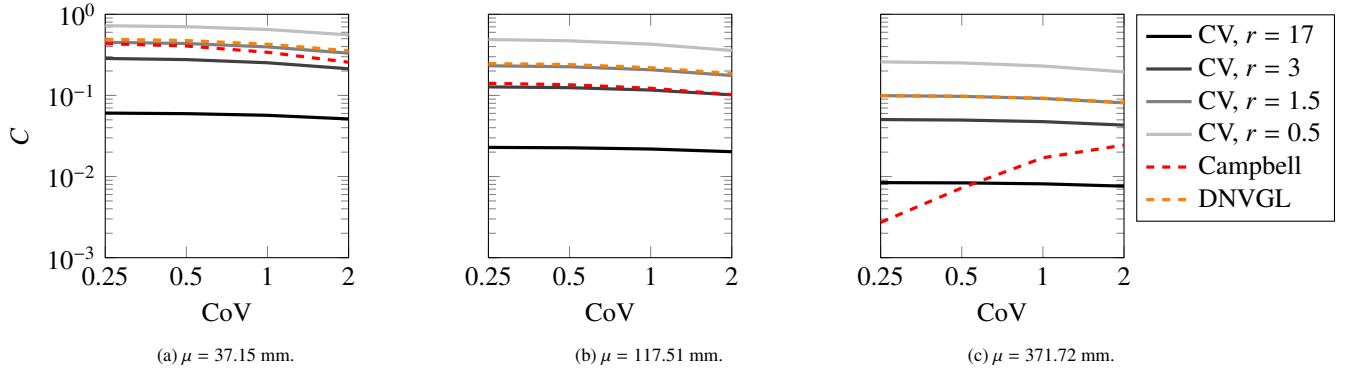
\begin{figure*}
\centering
\begin{subfigure}[t]{0.36\linewidth}
\begin{tikzpicture}[baseline=(current bounding box.south west)]
\begin{loglogaxis}[
    width=4.8cm,
    height=4.8cm,
    xlabel={$\CoV$},
    ylabel={$C$},
    xtick={0.25, 0.5, 1, 2},
    xticklabels={0.25, 0.5, 1, 2},
    xmin=0.25, xmax=2,
    ymin=0.001, ymax=1,
    legend pos=south west,
]

\addplot[cCVr1]
table[
    x=CoV,
    y=CV_17_0,
    restrict expr to domain={\thisrow{fig}}{0:0},
] {\datatable};

\addplot[cCVr2]
table[
    x=CoV,
    y=CV_3_0,
    restrict expr to domain={\thisrow{fig}}{0:0},
] {\datatable};

\addplot[cCVr3]
table[
    x=CoV,
    y=CV_1_5,
    restrict expr to domain={\thisrow{fig}}{0:0},
] {\datatable};

\addplot[cCVr4]
table[
    x=CoV,
    y=CV_0_5,
    restrict expr to domain={\thisrow{fig}}{0:0},
] {\datatable};

\addplot[cCampbell,dashed]
table[
    x=CoV,
    y=Campbell,
    restrict expr to domain={\thisrow{fig}}{0:0},
] {\datatable};

\addplot[cDNVGL,dashed]
table[
    x=CoV,
    y=DNVGL,
    restrict expr to domain={\thisrow{fig}}{0:0},
] {\datatable};

\end{loglogaxis}
\end{tikzpicture}
\caption{$\mu = 37.15$ mm.}
\end{subfigure}
\begin{subfigure}[t]{0.31\linewidth}
\begin{tikzpicture}[baseline=(current bounding box.south)]
\begin{loglogaxis}[
    width=4.8cm,
    height=4.8cm,
    xlabel={$\CoV$},
    xtick={0.25, 0.5, 1, 2},
    xticklabels={0.25, 0.5, 1, 2},
    yticklabels=none,
    xmin=0.25, xmax=2,
    ymin=0.001, ymax=1,
]

\addplot[cCVr1]
table[
    x=CoV,
    y=CV_17_0,
    restrict expr to domain={\thisrow{fig}}{1:1},
] {\datatable};

\addplot[cCVr2]
table[
    x=CoV,
    y=CV_3_0,
    restrict expr to domain={\thisrow{fig}}{1:1},
] {\datatable};

\addplot[cCVr3]
table[
    x=CoV,
    y=CV_1_5,
    restrict expr to domain={\thisrow{fig}}{1:1},
] {\datatable};

\addplot[cCVr4]
table[
    x=CoV,
    y=CV_0_5,
    restrict expr to domain={\thisrow{fig}}{1:1},
] {\datatable};

\addplot[cCampbell,dashed]
table[
    x=CoV,
    y=Campbell,
    restrict expr to domain={\thisrow{fig}}{1:1},
] {\datatable};

\addplot[cDNVGL,dashed]
table[
    x=CoV,
    y=DNVGL,
    restrict expr to domain={\thisrow{fig}}{1:1},
] {\datatable};

\end{loglogaxis}
\end{tikzpicture}
\caption{$\mu = 117.51$ mm.}
\end{subfigure}
\begin{subfigure}[t]{0.31\linewidth}
\begin{tikzpicture}[baseline=(current bounding box.south)]
\begin{loglogaxis}[
    width=4.8cm,
    height=4.8cm,
    xlabel={$\CoV$},
    xtick={0.25, 0.5, 1, 2},
    xticklabels={0.25, 0.5, 1, 2},
    yticklabels=none,
    xmin=0.25, xmax=2,
    ymin=0.001, ymax=1,
    legend style={
        at={(1.05,1)},
        anchor=north west,
        cells={anchor=west}
    },
]

\addplot[cCVr1]
table[
    x=CoV,
    y=CV_17_0,
    restrict expr to domain={\thisrow{fig}}{2:2},
] {\datatable};

\addplot[cCVr2]
table[
    x=CoV,
    y=CV_3_0,
    restrict expr to domain={\thisrow{fig}}{2:2},
] {\datatable};

\addplot[cCVr3]
table[
    x=CoV,
    y=CV_1_5,
    restrict expr to domain={\thisrow{fig}}{2:2},
] {\datatable};

\addplot[cCVr4]
table[
    x=CoV,
    y=CV_0_5,
    restrict expr to domain={\thisrow{fig}}{2:2},
] {\datatable};

\addplot[cCampbell,dashed]
table[
    x=CoV,
    y=Campbell,
    restrict expr to domain={\thisrow{fig}}{2:2},
] {\datatable};

\addplot[cDNVGL,dashed]
table[
    x=CoV,
    y=DNVGL,
    restrict expr to domain={\thisrow{fig}}{2:2},
] {\datatable};

\addlegendentry{CV, $r = 17$}
\addlegendentry{CV, $r = 3$}
\addlegendentry{CV, $r = 1.5$}
\addlegendentry{CV, $r = 0.5$}
\addlegendentry{Campbell}
\addlegendentry{DNVGL}

\end{loglogaxis}
\end{tikzpicture}
\caption{$\mu = 371.72$ mm.}
\end{subfigure}
\caption{Results of computation of normalised undetected crack length $C$ for our CV method and conventional visual inspection provided by Campbell et al.~\cite{Campbell2019pod} and by the DNVGL recommendation \cite{DNVGL_RP_C210_2015}}\label{fig:C_resolution}
\end{figure*}

\pgfplotstableread{Figures/update_resolution_KL.dat}\datatable
\begin{figure*}
\centering
\begin{subfigure}[t]{0.36\linewidth}
\begin{tikzpicture}[baseline=(current bounding box.south west)]
\begin{loglogaxis}[
    width=4.8cm,
    height=4.8cm,
    xlabel={$\CoV$},
    ylabel={$\KL$},
    xtick={0.25, 0.5, 1, 2},
    xticklabels={0.25, 0.5, 1, 2},
    xmin=0.25, xmax=2,
    ymin=0.001, ymax=2.5,
    legend pos=south west,
]

\addplot[cCVr1]
table[
    x=CoV,
    y=CV_17_0,
    restrict expr to domain={\thisrow{fig}}{0:0},
] {\datatable};

\addplot[cCVr2]
table[
    x=CoV,
    y=CV_3_0,
    restrict expr to domain={\thisrow{fig}}{0:0},
] {\datatable};

\addplot[cCVr3]
table[
    x=CoV,
    y=CV_1_5,
    restrict expr to domain={\thisrow{fig}}{0:0},
] {\datatable};

\addplot[cCVr4]
table[
    x=CoV,
    y=CV_0_5,
    restrict expr to domain={\thisrow{fig}}{0:0},
] {\datatable};

\addplot[cCampbell,dashed]
table[
    x=CoV,
    y=Campbell,
    restrict expr to domain={\thisrow{fig}}{0:0},
] {\datatable};

\addplot[cDNVGL,dashed]
table[
    x=CoV,
    y=DNVGL,
    restrict expr to domain={\thisrow{fig}}{0:0},
] {\datatable};

\end{loglogaxis}
\end{tikzpicture}
\caption{$\mu = 37.15$ mm.}
\end{subfigure}
\begin{subfigure}[t]{0.31\linewidth}
\begin{tikzpicture}[baseline=(current bounding box.south)]
\begin{loglogaxis}[
    width=4.8cm,
    height=4.8cm,
    xlabel={$\CoV$},
    xtick={0.25, 0.5, 1, 2},
    xticklabels={0.25, 0.5, 1, 2},
    yticklabels=none,
    xmin=0.25, xmax=2,
    ymin=0.001, ymax=2.5,
]

\addplot[cCVr1]
table[
    x=CoV,
    y=CV_17_0,
    restrict expr to domain={\thisrow{fig}}{1:1},
] {\datatable};

\addplot[cCVr2]
table[
    x=CoV,
    y=CV_3_0,
    restrict expr to domain={\thisrow{fig}}{1:1},
] {\datatable};

\addplot[cCVr3]
table[
    x=CoV,
    y=CV_1_5,
    restrict expr to domain={\thisrow{fig}}{1:1},
] {\datatable};

\addplot[cCVr4]
table[
    x=CoV,
    y=CV_0_5,
    restrict expr to domain={\thisrow{fig}}{1:1},
] {\datatable};

\addplot[cCampbell,dashed]
table[
    x=CoV,
    y=Campbell,
    restrict expr to domain={\thisrow{fig}}{1:1},
] {\datatable};

\addplot[cDNVGL,dashed]
table[
    x=CoV,
    y=DNVGL,
    restrict expr to domain={\thisrow{fig}}{1:1},
] {\datatable};

\end{loglogaxis}
\end{tikzpicture}
\caption{$\mu = 117.51$ mm.}
\end{subfigure}
\begin{subfigure}[t]{0.31\linewidth}
\begin{tikzpicture}[baseline=(current bounding box.south)]
\begin{loglogaxis}[
    width=4.8cm,
    height=4.8cm,
    xlabel={$\CoV$},
    xtick={0.25, 0.5, 1, 2},
    xticklabels={0.25, 0.5, 1, 2},
    yticklabels=none,
    xmin=0.25, xmax=2,
    ymin=0.001, ymax=2.5,
    legend style={
        at={(1.05,1)},
        anchor=north west,
        cells={anchor=west}
    },
]

\addplot[cCVr1]
table[
    x=CoV,
    y=CV_17_0,
    restrict expr to domain={\thisrow{fig}}{2:2},
] {\datatable};

\addplot[cCVr2]
table[
    x=CoV,
    y=CV_3_0,
    restrict expr to domain={\thisrow{fig}}{2:2},
] {\datatable};

\addplot[cCVr3]
table[
    x=CoV,
    y=CV_1_5,
    restrict expr to domain={\thisrow{fig}}{2:2},
] {\datatable};

\addplot[cCVr4]
table[
    x=CoV,
    y=CV_0_5,
    restrict expr to domain={\thisrow{fig}}{2:2},
] {\datatable};

\addplot[cCampbell,dashed]
table[
    x=CoV,
    y=Campbell,
    restrict expr to domain={\thisrow{fig}}{2:2},
] {\datatable};

\addplot[cDNVGL,dashed]
table[
    x=CoV,
    y=DNVGL,
    restrict expr to domain={\thisrow{fig}}{2:2},
] {\datatable};

\addlegendentry{CV, $r = 17$}
\addlegendentry{CV, $r = 3$}
\addlegendentry{CV, $r = 1.5$}
\addlegendentry{CV, $r = 0.5$}
\addlegendentry{Campbell}
\addlegendentry{DNVGL}

\end{loglogaxis}
\end{tikzpicture}
\caption{$\mu = 371.72$ mm.}
\end{subfigure}
\caption{Results of computation of the Kullback-Leibler divergence $\KL$ for our CV method and conventional visual inspection provided by Campbell et al.~\cite{Campbell2019pod} and by the DNVGL recommendation \cite{DNVGL_RP_C210_2015}}\label{fig:KL_resolution}
\end{figure*}
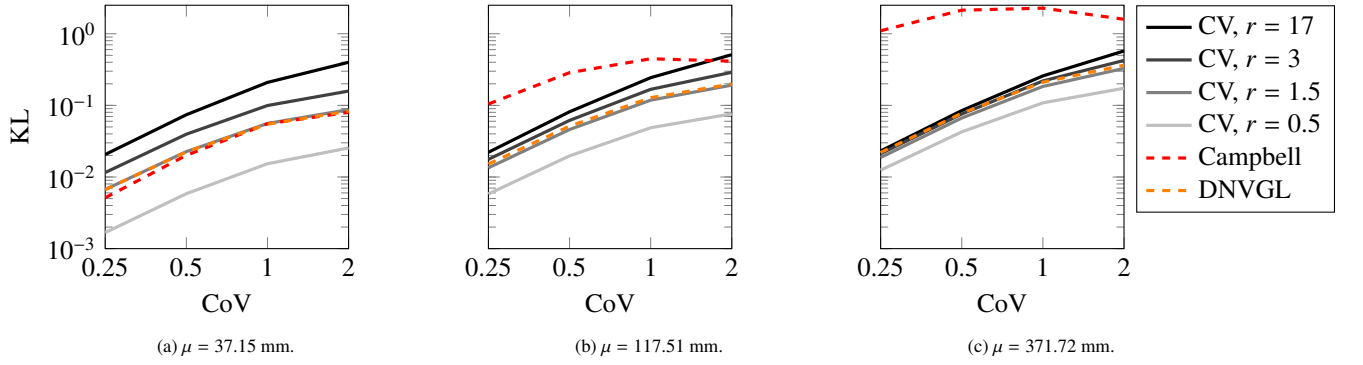

\begin{table*}
\centering
\caption{Results of computation of the normalised undetected crack length $C$ for the CV method at different resolutions and conventional visual inspection with PoDs provided by Campbell et al.~\cite{Campbell2019pod} and by the DNVGL recommendation \cite{DNVGL_RP_C210_2015}}
\label{tab:crack_update_C}
\begin{tabular}{cc|cccc|cc}
\toprule
$\mu$ (mm) & $\CoV$ & \multicolumn{4}{c|}{$C_{\mathrm{CV}}$} & $C_{\mathrm{Campbell}}$ & $C_{\mathrm{DNVGL}}$ \\ \cmidrule(lr){3-6}
& & $r=17$ & $r=3$ & $r=1.5$ & $r=0.5$ & & \\
\midrule
\multirow{4}{*}{37.15} & 0.25 & 0.061 & 0.285 & 0.452 & 0.723 & 0.434 & 0.493 \\
 & 0.5 & 0.060 & 0.276 & 0.436 & 0.703 & 0.406 & 0.474 \\
 & 1 & 0.057 & 0.253 & 0.396 & 0.650 & 0.340 & 0.428 \\
 & 2 & 0.051 & 0.213 & 0.331 & 0.557 & 0.256 & 0.355 \\
\hline
\multirow{4}{*}{117.51} & 0.25 & 0.023 & 0.127 & 0.232 & 0.489 & 0.141 & 0.247 \\
 & 0.5 & 0.023 & 0.125 & 0.225 & 0.471 & 0.135 & 0.240 \\
 & 1 & 0.022 & 0.116 & 0.207 & 0.428 & 0.122 & 0.220 \\
 & 2 & 0.020 & 0.102 & 0.176 & 0.358 & 0.102 & 0.186 \\
\hline
\multirow{4}{*}{371.72} & 0.25 & 0.008 & 0.050 & 0.099 & 0.259 & 0.003 & 0.099 \\
 & 0.5 & 0.008 & 0.050 & 0.097 & 0.251 & 0.007 & 0.098 \\
 & 1 & 0.008 & 0.048 & 0.092 & 0.230 & 0.017 & 0.092 \\
 & 2 & 0.008 & 0.043 & 0.081 & 0.195 & 0.024 & 0.082 \\
\bottomrule
\end{tabular}
\end{table*}

\begin{table*}
\centering
\caption{Results of computation of the Kullback-Leibler divergence $\KL$ for the CV method at different resolutions and conventional visual inspection with PoDs provided by Campbell et al.~\cite{Campbell2019pod} and by the DNVGL recommendation \cite{DNVGL_RP_C210_2015}.}
\label{tab:crack_update_KL}
\begin{tabular}{cc|cccc|cc}
\toprule
$\mu$ (mm) & $\CoV$ & \multicolumn{4}{c|}{$\KL_{\mathrm{CV}}$} & $\KL_{\mathrm{Campbell}}$ & $\KL_{\mathrm{DNVGL}}$ \\ \cmidrule(lr){3-6}
& & $r=17$ & $r=3$ & $r=1.5$ & $r=0.5$ & & \\
\midrule
\multirow{4}{*}{37.15} & 0.25 & 0.021 & 0.012 & 0.007 & 0.002 & 0.005 & 0.007 \\
 & 0.5 & 0.074 & 0.040 & 0.023 & 0.006 & 0.020 & 0.022 \\
 & 1 & 0.209 & 0.099 & 0.056 & 0.015 & 0.055 & 0.055 \\
 & 2 & 0.400 & 0.158 & 0.087 & 0.025 & 0.080 & 0.083 \\
\hline
\multirow{4}{*}{117.51} & 0.25 & 0.022 & 0.018 & 0.013 & 0.006 & 0.104 & 0.015 \\
 & 0.5 & 0.081 & 0.062 & 0.046 & 0.020 & 0.286 & 0.051 \\
 & 1 & 0.244 & 0.167 & 0.118 & 0.049 & 0.447 & 0.128 \\
 & 2 & 0.510 & 0.290 & 0.192 & 0.076 & 0.414 & 0.199 \\
\hline
\multirow{4}{*}{371.72} & 0.25 & 0.023 & 0.021 & 0.019 & 0.012 & 1.100 & 0.022 \\
 & 0.5 & 0.085 & 0.076 & 0.067 & 0.043 & 2.132 & 0.078 \\
 & 1 & 0.258 & 0.219 & 0.184 & 0.108 & 2.276 & 0.211 \\
 & 2 & 0.575 & 0.424 & 0.328 & 0.173 & 1.593 & 0.363 \\
\bottomrule
\end{tabular}
\end{table*}

\section{Conclusions}\label{sec:Conclusions}
In this paper, we proposed a new framework, namely Alg.~\ref{alg:missed_crack_distribution}, for the evaluation of computer vision methods for the detection of cracks using PoD curves (Fig.~\ref{fig:PoD, cb, comp.}). 
Unlike standard evaluation metrics such as F1-score, intersection over union, mean average precision, etc., which are commonly used to evaluate the performance of computer vision methods, this approach allows the comparison of our computer vision method to conventional visual inspection of bridges performed by a human inspector. The typical evaluation results (histograms, posterior crack length distributions) of the approach, comparing our CV method to Campbell's method \cite{Campbell2019pod} and the DNVGL recommendation \cite{DNVGL_RP_C210_2015},  are shown in Fig.~\ref{fig:Histograms}. 
The proposed evaluation and comparison provides a better insight into the true performance of the computer vision method and can be used to assess the reliability of the method in real applications. 
In general, the approach is suitable for determining PoD curves for any automated approach (AI-based) for damage detection. As our widely applicable evaluation method can be integrated into  general structural reliability frameworks, we release the comparison code \cite{Sherry2026PoDCurveComparison}.
% at \url{https://github.com/finnsherry/PoDCurveComparison}. 

Based on existing PoD curves for conventional visual inspection, our computer vision method trained on a selected subset of images of the ``Cracks in Steel Bridges'' dataset \cite{CSB_dataset}, is capable of outperforming conventional visual inspection to detect cracks of length smaller than 150 mm. 
For larger cracks, the performance of our computer vision method lies between different estimates of conventional visual inspection performance, namely by Campbell et al.~\cite{Campbell2019pod}, and by the DNVGL recommendation \cite{DNVGL_RP_C210_2015}.
These findings are also supported numerically, see Figs.~\ref{fig:C graphs}~and~\ref{fig:KL graphs}. Moreover, the results are consistent across a wide range of prior crack length distributions, as can be seen in \ref{app:A}.
Hence, automated visual inspection with our computer vision and conventional visual inspection are complementary and give the best performance when applied in conjunction. 

The resolution of the images plays an important role in the detectability of cracks by our computer vision method.
We analysed this effect using two approaches: a nonparametric approach (Sec.~\ref{sec:ResolutionEffect of Resolution on PoD:Resolution effect with binning}) and a parametric one (Sec.~\ref{sec:Effect of Resolution on PoD:pod surface}).
Using our proposed inspection comparison method (Alg.~\ref{alg:missed_crack_distribution}), we find that our computer vision method gives better performance than conventional visual inspection on the entire range of crack lengths, assuming that the images were collected by a camera with IFOV less than 0.0001 rad/pix and from a distance less than 3 m.

Overall, we have established a statistical evaluation framework to allow the comparison of CV crack detection methods with conventional inspection methods. Although beyond the scope of this article, it would be interesting to compare other CV methods using this evaluation approach in the future.

\section*{Acknowledgments}
The authors would like to thank the Dutch bridge infrastructure owners ProRail and Rijks\-waterstaat, and the company Nebest, for their support in acquiring information on cracks in steel bridges. 
The research is primarily funded by the Eindhoven Artificial Intelligence Systems Institute (EAISI), and partly by the Dutch Foundation of Science NWO (Geometric Learning for Image Analysis, VI.C 202-031).

{\small
\bibliographystyle{elsarticle-num}
\bibliography{references} 

@techreport{Rijkswaterstaat2022ReportCondition,
  title         = {Staat van de Infrastructuur Rijkswaterstaat},
  institution   = {Rijkswaterstaat},
  year          = {2022},
  url           = {https://www.rijksoverheid.nl/documenten/rapporten/2023/12/04/bijlage-3-staat-van-de-infrastructuur-2023},
  note          = {accessed: 2024-03-06},
}

@techreport{brady2009cost345,
  title         = {Cost 345. Procedures required for the assessment of highway structures. Final report},
  author        = {Brady, Kenneth C. and O’Reilly, Miles and Bevc, Lojze and Znidarič, Aleš and O’Brien, Eugene and Jordan, Richard},
  institution   = {European Co-operation in the Field of Scientific and Technical Research, Brussels, Belgium},
  year          = {2009},
  url           = {http://cost345.zag.si/Reports/COST_345_Summary_Document.pdf},
}

@article{shahrivar2025ai_shm,
  title     = {AI-based bridge maintenance management: a comprehensive review},
  author    = {Shahrivar, Farham and Sidiq, Amir and Mahmoodian, Mojtaba and Jayasinghe, Sanduni and Sun, Zhiyan and Setunge, Sujeeva},
  journal   = {Artificial Intelligence Review},
  volume    = {58},
  number    = {5},
  pages     = {135},
  year      = {2025},
  publisher = {Springer},
  doi       = {10.1007/s10462-025-11144-7}
}

@article{isac2025SHM_trends,
  title     = {The Current Status of Structural Monitoring: A Bibliometric Literature Review},
  author    = {Isac, Mihai Dorin and C{\^\i}mpean, Cosmina and Manea, Daniela Lucia},
  journal   = {Buildings},
  volume    = {15},
  number    = {5},
  pages     = {739},
  year      = {2025},
  publisher = {MDPI},
  doi       = {10.3390/buildings15050739}
}

@article{ali2022AI_inspection_trends,
  title     = {Bibliometric analysis and review of deep learning-based crack detection literature published between 2010 and 2022},
  author    = {Ali, Luqman and Alnajjar, Fady and Khan, Wasif and Serhani, Mohamed Adel and Al Jassmi, Hamad},
  journal   = {Buildings},
  volume    = {12},
  number    = {4},
  pages     = {432},
  year      = {2022},
  publisher = {MDPI},
  doi       = {10.3390/buildings12040432},
}

@article{bridges2004d1,
  title     = {D1. 2 European Railway Bridge Demography},
  author    = {Bell, Brian},
  journal   = {Sustainable Bridges--Assessment for Future Traffic Demands and Longer Lives--a project within EU FP6},
  year      = {2004},
  url       = {https://www.diva-portal.org/smash/get/diva2:1330533/FULLTEXT01.pdf}
}

@techreport{MILHDBK1823A,
  title        = {Nondestructive Evaluation System Reliability Assessment},
  number       = {MIL-HDBK-1823A},
  institution  = {U.S. Department of Defense},
  address      = {Wright-Patterson Air Force Base, OH},
  year         = {2009},
  month        = apr,
  note         = {Handbook; supersedes MIL-HDBK-1823 (2004)},
  url          = {https://statistical-engineering.com/wp-content/uploads/2017/10/MIL-HDBK-1823A2009.pdf}
}

@techreport{ENIQ_POD_BestPractices_2011,
  author       = {Gandossi, Luca and Annis, Charles},
  title        = {Probability of {D}etection {C}urves: {S}tatistical {B}est-{P}ractices},
  type         = {ENIQ TGR Technical Document},
  number       = {ENIQ Report No. 41; EUR 24429 EN},
  institution  = {European Commission, Joint Research Centre (JRC), Institute for Energy},
  year         = {2011},
  month        = jan,
  doi          = {10.2790/21826},
  isbn         = {978-92-79-16105-6},
  issn         = {1018-5593},
  note         = {Petten, The Netherlands}
}

@techreport{Campbell2019pod,
  title         = {Probability of {D}etection {S}tudy for {V}isual {I}nspection of {S}teel {B}ridges: {V}olume 2—{F}ull {P}roject {R}eport},
  author        = {Campbell, Leslie E. and Snyder, Luke R. and Whitehead, Julie M. and Connor, Robert J. and Lloyd, Jason B.},
  year          = {2019},
  institution   = {Indiana Dept. of Transportation},
  doi           = {10.5703/1288284317104}
}

@article{amirkhani2024ReviewDLInspect,
  title     = {Visual concrete bridge defect classification and detection using deep learning: A systematic review},
  author    = {Amirkhani, Dariush and Allili, Mohand Sa{\"\i}d and Hebbache, Loucif and Hammouche, Nadir and Lapointe, Jean-Fran{\c{c}}ois},
  journal   = {Transactions on Intelligent Transportation Systems},
  volume    = {25},
  number    = {9},
  pages     = {10483--10505},
  year      = {2024},
  publisher = {IEEE},
  doi       = {10.1109/TITS.2024.3365296}
}

@article{panigati2025ReviewDronesInspect,
  title     = {Drone-based bridge inspections: Current practices and future directions},
  author    = {Panigati, Tommaso and Zini, Mattia and Striccoli, Domenico and Giordano, Pier Francesco and Tonelli, Daniel and Limongelli, Maria Pina and Zonta, Daniele},
  journal   = {Automation in Construction},
  volume    = {173},
  pages     = {106101},
  year      = {2025},
  publisher = {Elsevier},
  doi       = {10.1016/j.autcon.2025.106101},
}

@Article{KimAIvsMAN12022,
  AUTHOR            = {Kim, In-Ho and Yoon, Sungsik and Lee, Jin Hwan and Jung, Sungwook and Cho, Soojin and Jung, Hyung-Jo},
  TITLE             = {A Comparative Study of Bridge Inspection and Condition Assessment between Manpower and a UAS},
  JOURNAL           = {Drones},
  VOLUME            = {6},
  YEAR              = {2022},
  NUMBER            = {11},
  ARTICLE-NUMBER    = {355},
  ISSN              = {2504-446X},
  DOI               = {10.3390/drones6110355},
  publisher         = {MDPI}
}

@article{wojcikAIvsMAN12020,
  title     = {Asesment of state-of-the-art methods for bridge inspection: case study},
  author    = {W{\'o}jcik, Bartosz and {\.Z}arski, Mateusz},
  journal   = {Archives of Civil Engineering},
  volume    = {66},
  number    = {4},
  year      = {2020},
  pages     = {343--362},
  doi       = {10.24425/ace.2020.135225}
}

@article{liu2019deepcrack,
  title     = {DeepCrack: A deep hierarchical feature learning architecture for crack segmentation},
  author    = {Liu, Yahui and Yao, Jian and Lu, Xiaohu and Xie, Renping and Li, Li},
  journal   = {Neurocomputing},
  volume    = {338},
  pages     = {139--153},
  year      = {2019},
  publisher = {Elsevier},
  doi       = {10.1016/j.neucom.2019.01.036}
}

@article{kim2025exploratory,
  title     = {An Exploratory Study on Crack Detection in Concrete through Human-Robot Collaboration},
  author    = {Kim, Junyeon and Ruan, Tianshu and Contreras, Cesar Alan and Chiou, Manolis},
  journal   = {arXiv preprint},
  year      = {2025},
  doi       = {10.48550/arXiv.2508.11404}
}

@techreport{DNVGL_RP_C210_2015,
  title        = {Probabilistic Methods for Planning of Inspection for Fatigue Cracks in Offshore Structures},
  institution  = {{DNV GL}},
  number       = {DNVGL-RP-C210},
  type         = {Recommended Practice},
  year         = {2015},
  month        = nov,
  url          = {https://www.dnv.com/energy/standards-guidelines/dnv-rp-c210-probabilistic-methods-for-planning-of-inspection-for-fatigue-cracks-in-offshore-structures/},
  urldate      = {2026-02-13}
}

@techreport{HSE2006ProbabilityCurves,
  title        = {Probability of Detection (PoD) curves},
  author       = {Georgiou, George A.},
  institution  = {{UK Health and Safety Executive}},
  number       = {454},
  type         = {Recommended Practice},
  year         = {2006},
  url          = {https://webarchive.nationalarchives.gov.uk/ukgwa/20241208023035/https://www.hse.gov.uk/research/rrhtm/rr454.htm},
  urldate      = {2026-06-25}
}

@inproceedings{Deng2009ImageNet,
  title         = {Imagenet: A large-scale hierarchical image database},
  author        = {Deng, Jia and Dong, Wei and Socher, Richard and Li, Li-Jia and Li, Kai and Fei-Fei, Li},
  booktitle     = {2009 IEEE conference on computer vision and pattern recognition},
  pages         = {248--255},
  year          = {2009},
  organization  = {IEEE},
  doi           = {https://doi.org/10.1109/CVPR.2009.5206848}
}

@misc{CSB_dataset,
  author        = {Kompanets, Andrii and Leonetti, Davide and Duits, Remco and Snijder, H.H. Bert},
  title         = {Cracks in {S}teel {B}ridges {(CSB)} dataset},
  year          = {2024},
  howpublished  = {4TU.ResearchData},
  version       = {1},
  doi           = {10.4121/6162a9b6-2a20-4600-8207-e9dcd53a264a}
}

@article{JP2,
  title     = {Two-stage fatigue crack detection framework with crack-preserving downsampler},
  author    = {Kompanets, Andrii and Duits, Remco and Leonetti, Davide and Snijder, H.H. Bert},
  journal   = {International Journal of Fatigue},
  volume    = {109179},
  pages     = {1--15},
  year      = {2025},
  publisher = {Elsevier},
  doi       = {10.1016/j.ijfatigue.2025.109179}
}

@ARTICLE{Ren2017FasterRCNN,
  author    = {Ren, Shaoqing and He, Kaiming and Girshick, Ross and Sun, Jian},
  journal   = {Transactions on Pattern Analysis and Machine Intelligence}, 
  publisher = {IEEE},
  title     = {Faster R-CNN: Towards Real-Time Object Detection with Region Proposal Networks}, 
  year      = {2017},
  volume    = {39},
  number    = {6},
  pages     = {1137--1149},
  doi       = {10.1109/TPAMI.2016.2577031}
}

@inproceedings{xie2017ResNext,
  title     = {Aggregated Residual Transformations for Deep Neural Networks}, 
  author    = {Saining Xie and Ross Girshick and Piotr Dollár and Zhuowen Tu and Kaiming He},
  year      = {2017},
  booktitle = {Conference on Computer Vision and Pattern Recognition (CVPR)},
  publisher = {IEEE},
  doi       = {10.1109/CVPR.2017.634},
  pages     = {5987--5995}
}

@inproceedings{Loshchilov2017AdamW,
  title     = {{D}ecoupled {W}eight {D}ecay {R}egularization},
  author    = {Loshchilov, Ilya and Hutter, Frank},
  booktitle = {International Conference on Learning Representations (ICLR)},
  year      = {2019},
  url       = {https://openreview.net/forum?id=Bkg6RiCqY7},
  pages     = {1--18}
}

@inproceedings{loshchilov2016CosineAnnealing,
  title     = {{Sgdr}: {S}tochastic {G}radient {D}escent with {W}arm {R}estarts},
  author    = {Loshchilov, Ilya and Hutter, Frank},
  booktitle = {International Conference on Learning Representations (ICLR)},
  year      = {2017},
  url       = {https://openreview.net/forum?id=Skq89Scxx},
  pages     = {1--16}
}

@article{mmdetection,
  title     = {{MMDetection}: Open MMLab Detection Toolbox and Benchmark},
  author    = {Chen, Kai and Wang, Jiaqi and Pang, Jiangmiao and Cao, Yuhang and
             Xiong, Yu and Li, Xiaoxiao and Sun, Shuyang and Feng, Wansen and
             Liu, Ziwei and Xu, Jiarui and Zhang, Zheng and Cheng, Dazhi and
             Zhu, Chenchen and Cheng, Tianheng and Zhao, Qijie and Li, Buyu and
             Lu, Xin and Zhu, Rui and Wu, Yue and Dai, Jifeng and Wang, Jingdong
             and Shi, Jianping and Ouyang, Wanli and Loy, Chen Change and Lin, Dahua},
  journal   = {arXiv preprint},
  year      = {2019},
  doi       = {10.48550/arXiv.1906.07155}
}

@article{righiniotis2006comparativePoDs,
  title     = {A comparative study of fatigue inspection methods},
  author    = {Righiniotis, Timothy D.},
  journal   = {Journal of Constructional Steel Research},
  volume    = {62},
  number    = {4},
  pages     = {352--358},
  year      = {2006},
  publisher = {Elsevier},
  doi       = {10.1016/j.jcsr.2005.08.008}
}

@article{Kullback1951KLDiv,
  author    = {S. Kullback and R. A. Leibler},
  title     = {On {I}nformation and {S}ufficiency},
  volume    = {22},
  journal   = {The Annals of Mathematical Statistics},
  number    = {1},
  publisher = {Institute of Mathematical Statistics},
  pages     = {79 -- 86},
  year      = {1951},
  doi       = {10.1214/aoms/1177729694},
}

@article{kaiser2009IFOVHumaneye,
  title     = {{Prospective evaluation of visual acuity assessment: a comparison of snellen versus ETDRS charts in clinical practice (An AOS Thesis)}},
  author    = {Kaiser, Peter K},
  journal   = {Transactions of the American Ophthalmological Society},
  volume    = {107},
  pages     = {311},
  year      = {2009},
  url       = {https://europepmc.org/articles/PMC2814576}
}

@article{JP1,
  title     = {Loss function inversion for improved crack segmentation in steel bridges using a CNN framework},
  author    = {Andrii Kompanets and Remco Duits and Gautam Pai and Davide Leonetti and H.H. (Bert) Snijder},
  journal   = {Automation in Construction},
  volume    = {170},
  pages     = {105896},
  year      = {2025},
  issn      = {0926-5805},
  doi       = {https://doi.org/10.1016/j.autcon.2024.105896},
}

@inproceedings{Leonetti2021InfluenceModels,
  author    = {Leonetti, Davide and Maljaars, Johan and Snijder, H.H. (Bert)},
  title     = {Influence of inspection on the safety of fatigue loaded welded cruciform steel joints – comparison of simplified and advanced crack growth models},
  booktitle = {IABSE CONGRESS GHENT 2021: Structural Engineering for Future Societal Needs},
  year      = {2021},
  editor    = {Snijder, H.H. (Bert) and De Pauw, Bart and {v}an Alphen, Sander and Mengeot, Pierre},
  pages     = {1546--1554},
  publisher = {International Association for Bridge and Structural Engineering},
  url       = {https://research.tue.nl/en/publications/influence-of-inspection-on-the-safety-of-fatigue-loaded-welded-cr/}
}

@article{Maljaars2014ProbabilisticLocations,
  author    = {Maljaars, Johan and Vrouwenvelder, A.C.W.M.},
  title     = {Probabilistic fatigue life updating accounting for inspections of multiple critical locations},
  journal   = {International Journal of Fatigue},
  publisher = {Elsevier},
  year      = {2014},
  doi       = {10.1016/j.ijfatigue.2014.06.011}
}

@inproceedings{Hashemi2019ValueBridges,
  author    = {Hashemi, Seyed and Maljaars, Johan and Snijder, H.H. (Bert)},
  title     = {Added value of regular in-service visual inspection to the fatigue reliability of structural details in steel bridges},
  booktitle = {5th International Conference on Smart Monitoring, Assessment and Rehabilitation of Civil Structures},
  year      = {2019},
  pages     = {1--8},
  isbn      = {978-3-947971-07-7},
  url       = {https://research.tue.nl/en/publications/added-value-of-regular-in-service-visual-inspection-to-the-fatigu/}
  }

@article{scikit-learn,
  title     = {Scikit-learn: Machine Learning in {P}ython},
  author    = {Pedregosa, F. and Varoquaux, G. and Gramfort, A. and Michel, V.
          and Thirion, B. and Grisel, O. and Blondel, M. and Prettenhofer, P.
          and Weiss, R. and Dubourg, V. and Vanderplas, J. and Passos, A. and
          Cournapeau, D. and Brucher, M. and Perrot, M. and Duchesnay, E.},
  journal   = {Journal of Machine Learning Research},
  volume    = {12},
  pages     = {2825--2830},
  year      = {2011},
  url       = {http://jmlr.org/papers/v12/pedregosa11a.html}
}

@misc{Sherry2026PoDCurveComparison,
  author       = {Sherry, Finn M.},
  title        = {{finnsherry/PoDCurveComparison}},
  month        = aug,
  year         = 2026,
  publisher    = {Zenodo},
  version      = {v0.0.0-alpha},
  doi          = {10.5281/zenodo.21771171},
  url          = {https://github.com/finnsherry/PoDCurveComparison},
}
}

\appendix
\section{Results of the crack update simulation}\label{app:A}
\begin{figure*}
\centering
\begin{subfigure}{0.49\textwidth}
    \includegraphics[width=\textwidth]{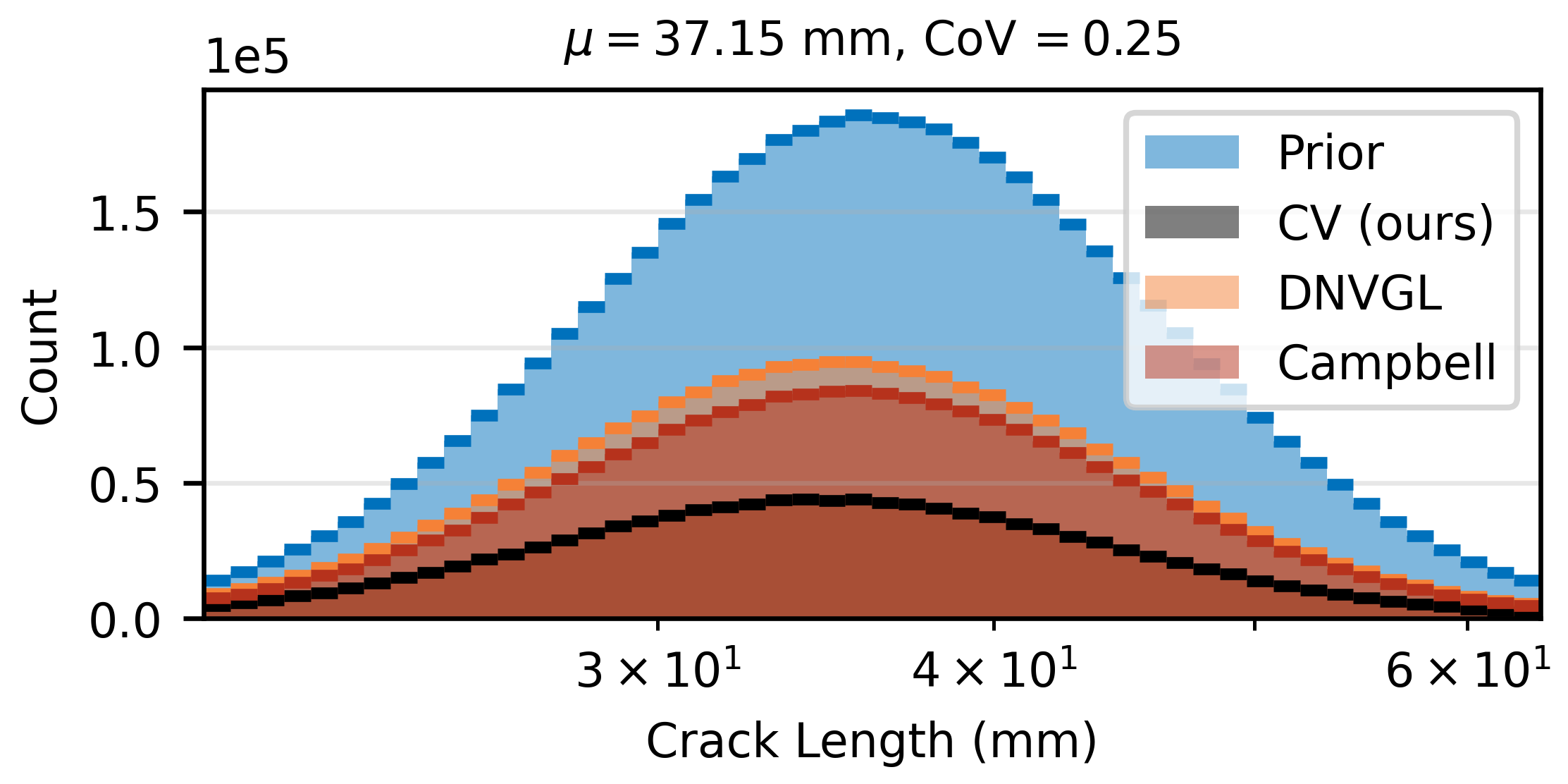}
    \caption{}
    \label{fig:Histograms:mu37.15cov0.25_hist}
\end{subfigure}
\begin{subfigure}{0.49\textwidth}
    \includegraphics[width=\textwidth]{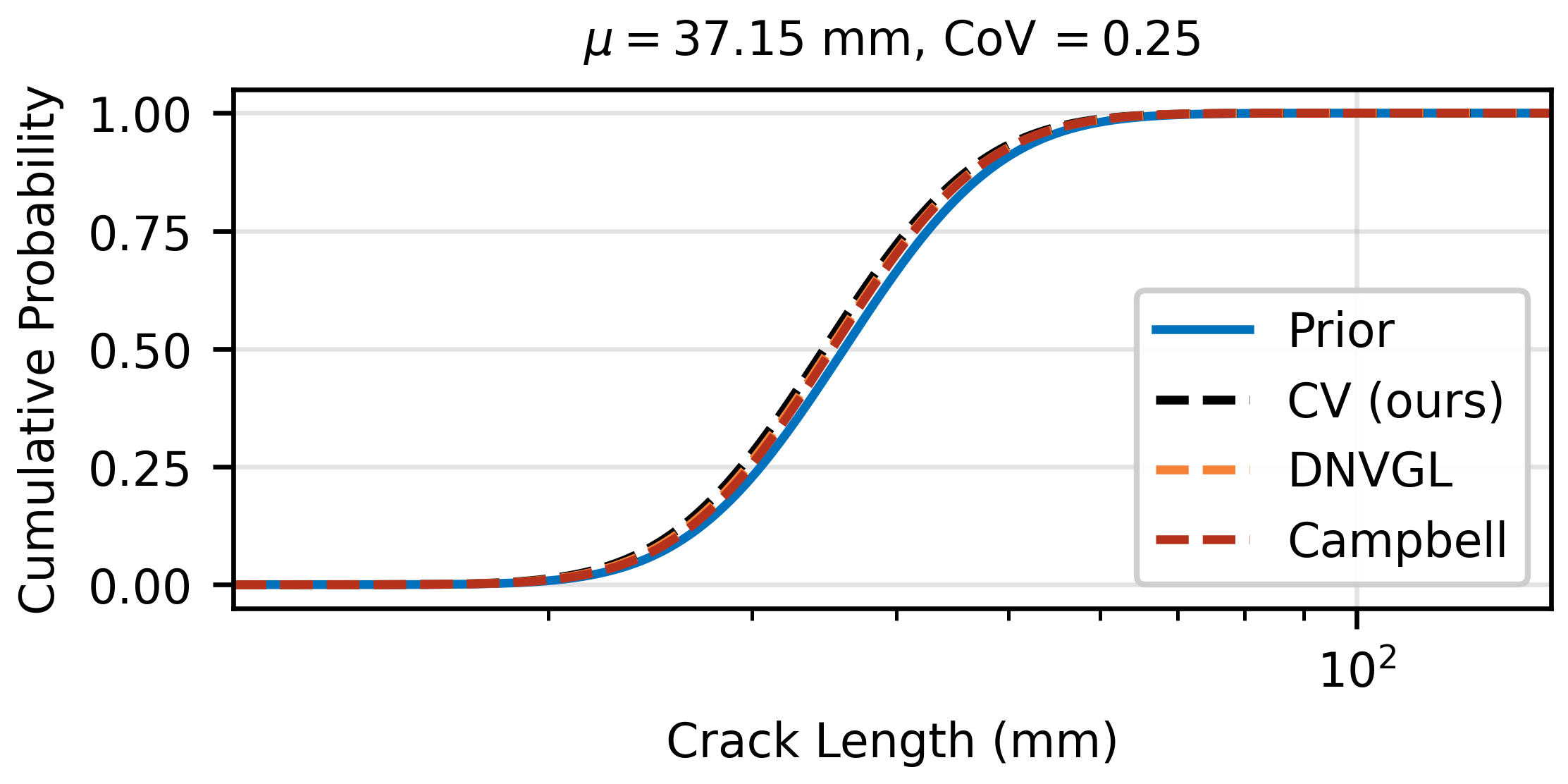}
    \caption{}
    \label{fig:Histograms:mu37.15cov0.25_cdf}
\end{subfigure}
\caption{Histogram and CDF for prior crack-length distribution with $\mu = 37.15$ mm and $\CoV = 0.25$.}
\label{fig:Histograms_mu_37.15_cov_0.25}
\end{figure*}

\begin{figure*}
\centering
\begin{subfigure}{0.49\textwidth}
    \includegraphics[width=\textwidth]{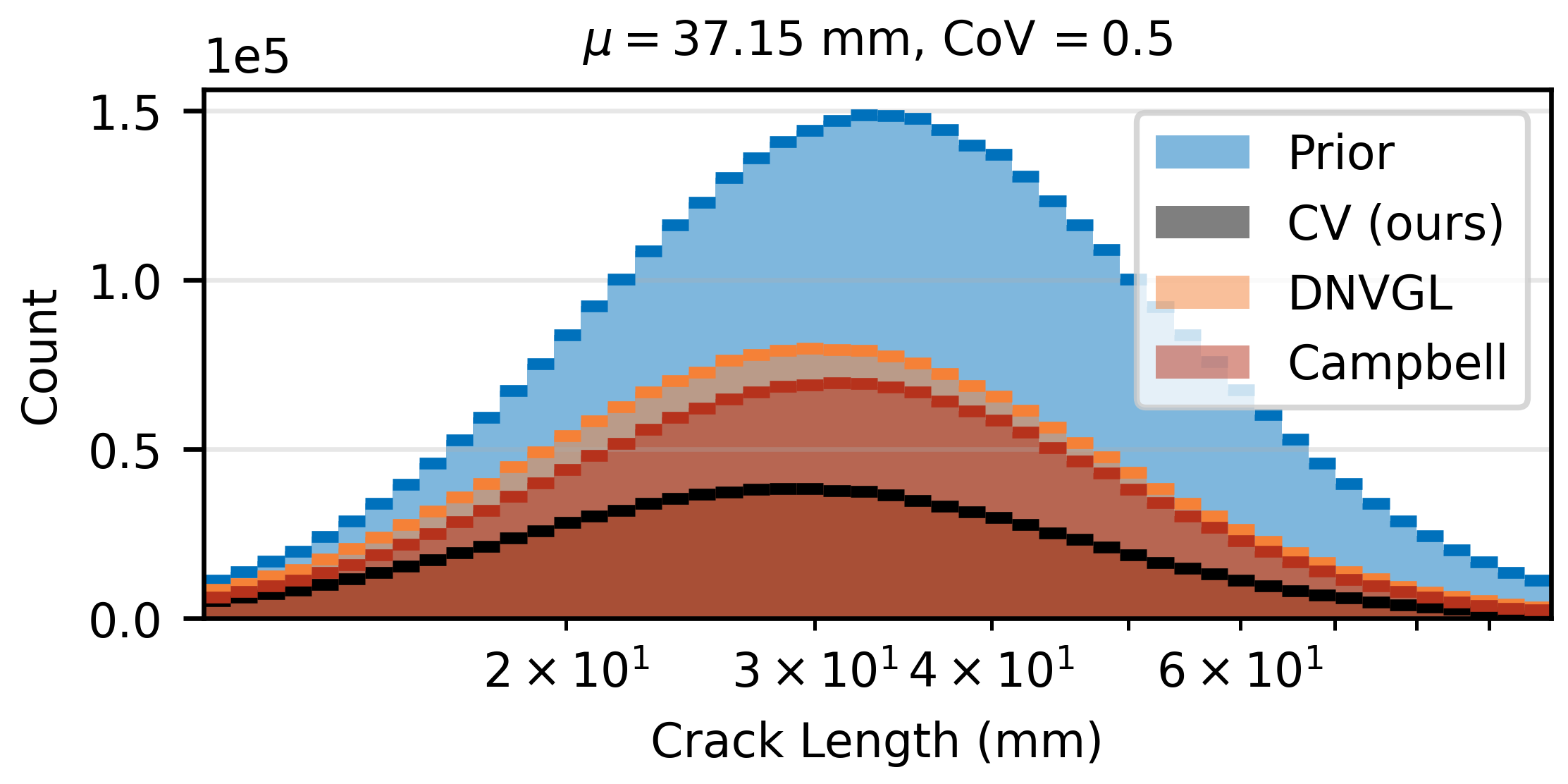}
    \caption{}
    \label{fig:Histograms:mu37.15cov0.5_hist}
\end{subfigure}
\begin{subfigure}{0.49\textwidth}
    \includegraphics[width=\textwidth]{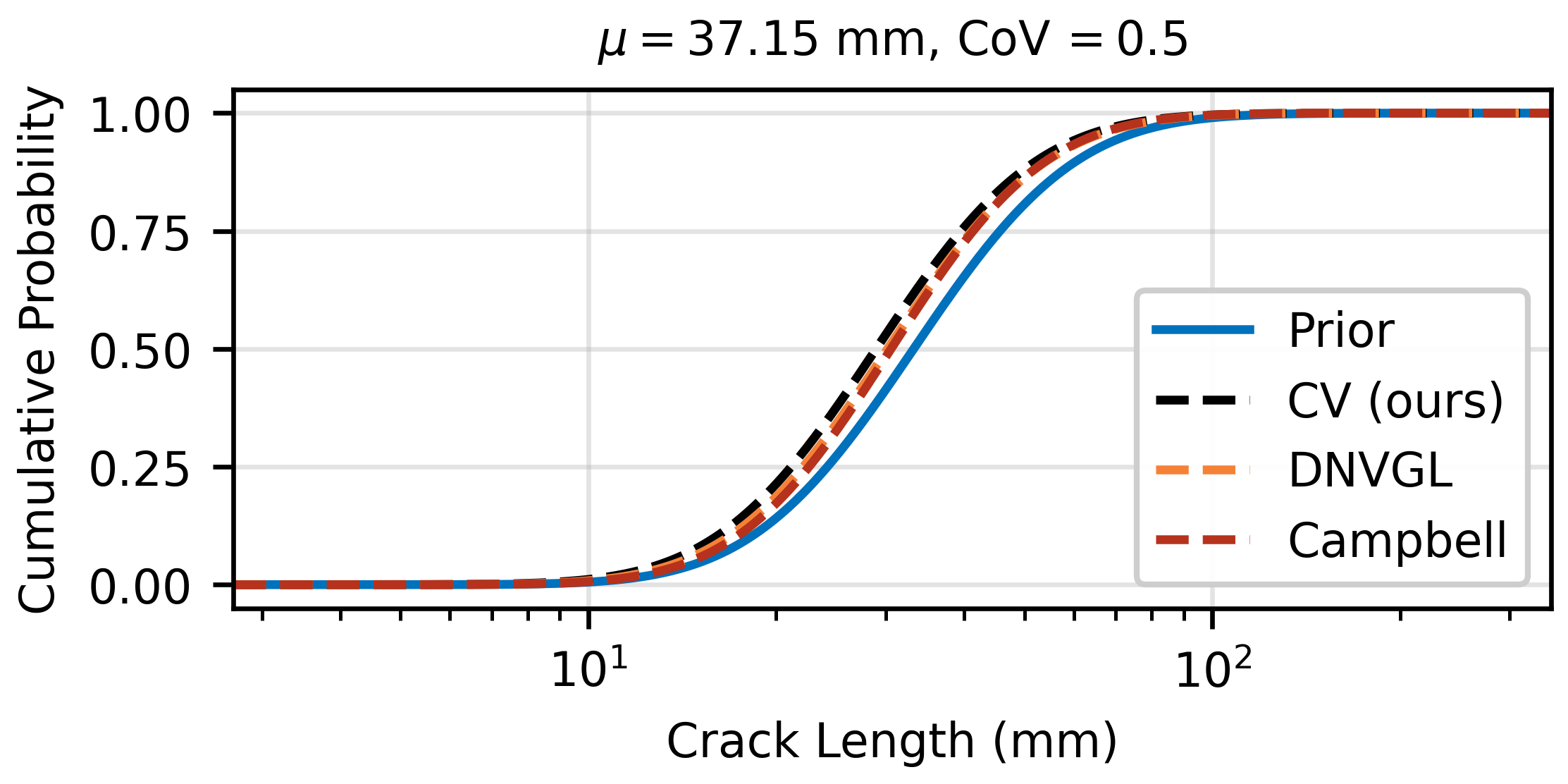}
    \caption{}
    \label{fig:Histograms:mu37.15cov0.5_cdf}
\end{subfigure}
\caption{Histogram and CDF for prior crack-length distribution with $\mu = 37.15$ mm and $\CoV = 0.5$.}
\label{fig:Histograms_mu_37.15_cov_0.5}
\end{figure*}

\begin{figure*}
\centering
\begin{subfigure}{0.49\textwidth}
    \includegraphics[width=\textwidth]{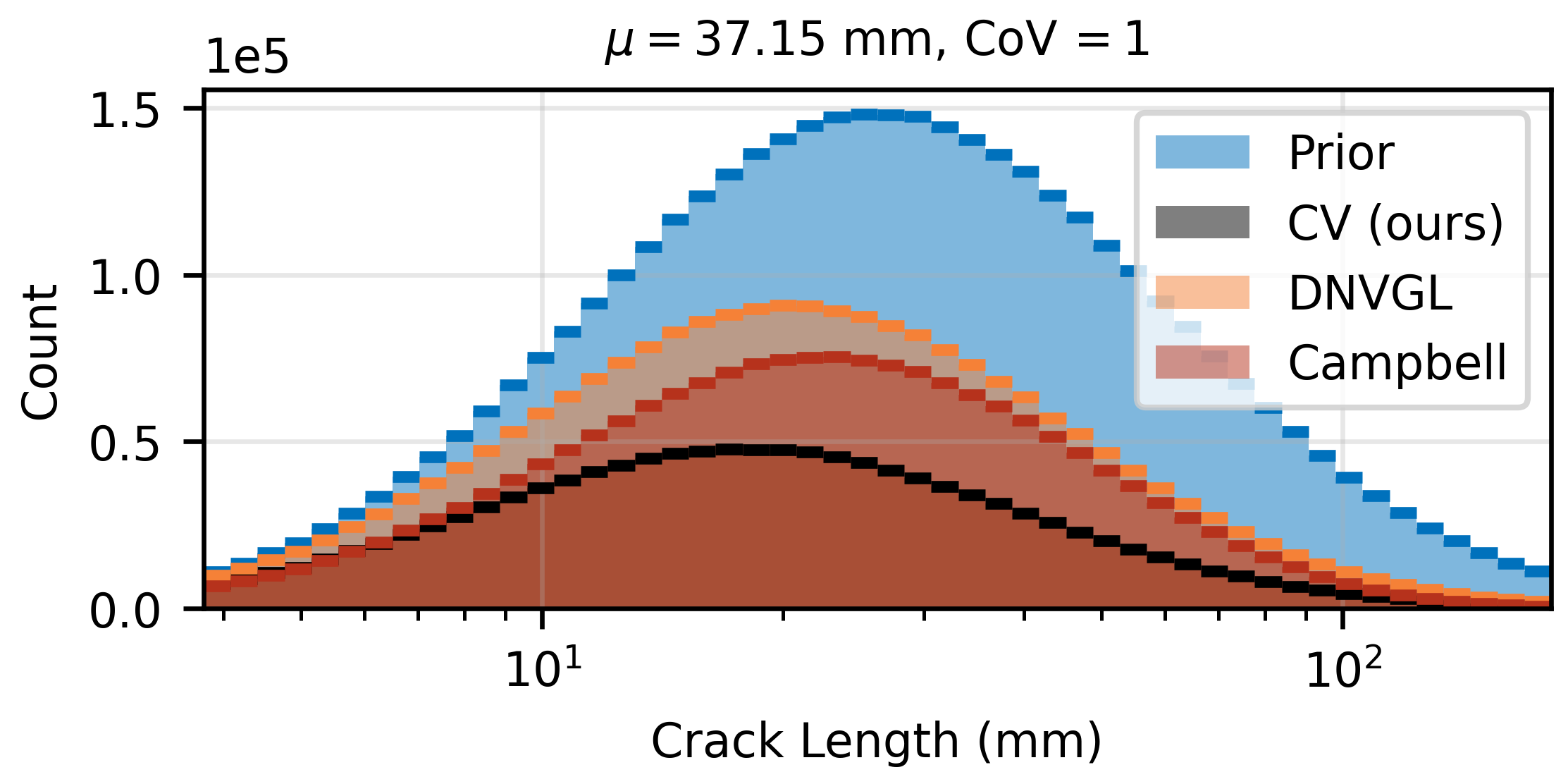}
    \caption{}
    \label{fig:Histograms:mu37.15cov1_hist}
\end{subfigure}
\begin{subfigure}{0.49\textwidth}
    \includegraphics[width=\textwidth]{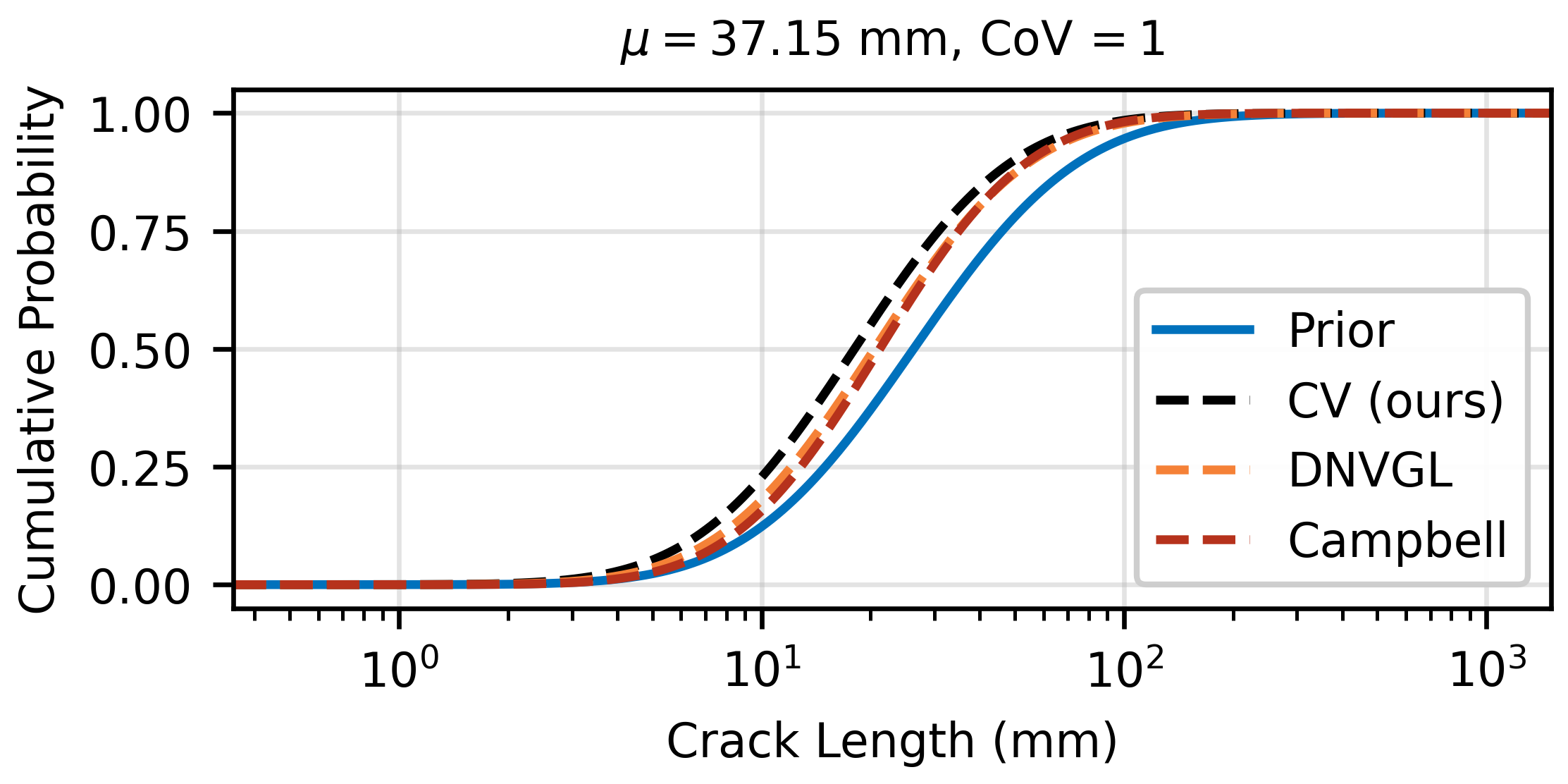}
    \caption{}
    \label{fig:Histograms:mu37.15cov1_cdf}
\end{subfigure}
\caption{Histogram and CDF for prior crack-length distribution with $\mu = 37.15$ mm and $\CoV = 1$.}
\label{fig:Histograms_mu_37.15_cov_1}
\end{figure*}

\begin{figure*}
\centering
\begin{subfigure}{0.49\textwidth}
    \includegraphics[width=\textwidth]{Figures/Update_Crack_Length/update_crack_length_histogram_37.15_0.25.png}
    \caption{}
    \label{fig:Histograms:mu37.15cov2_hist}
\end{subfigure}
\begin{subfigure}{0.49\textwidth}
    \includegraphics[width=\textwidth]{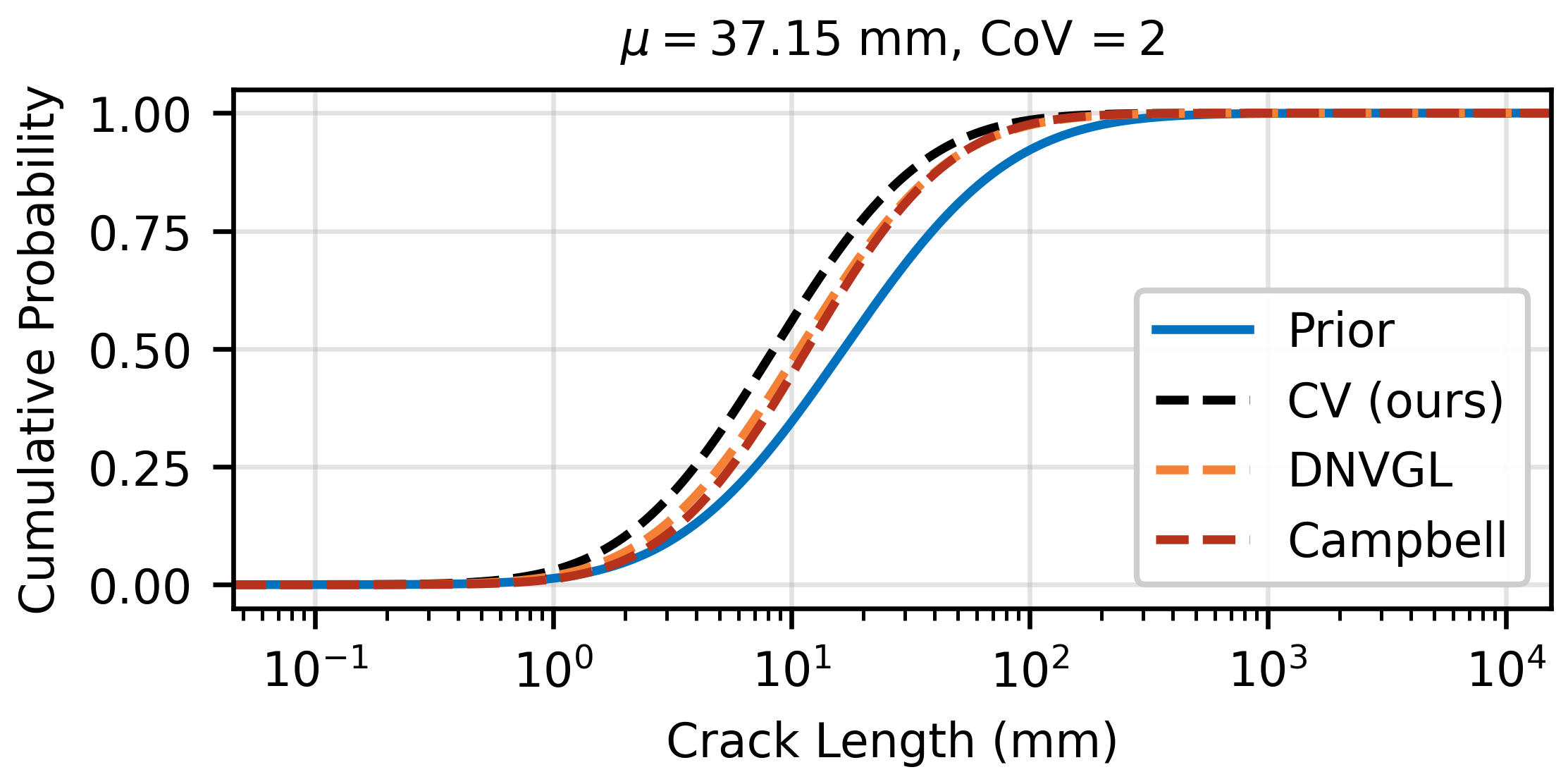}
    \caption{}
    \label{fig:Histograms:mu37.15cov2_cdf}
\end{subfigure}
\caption{Histogram and CDF for prior crack-length distribution with $\mu = 37.15$ mm and $\CoV = 2$.}
\label{fig:Histograms_mu_37.15_cov_2}
\end{figure*}

\begin{figure*}
\centering
\begin{subfigure}{0.49\textwidth}
    \includegraphics[width=\textwidth]{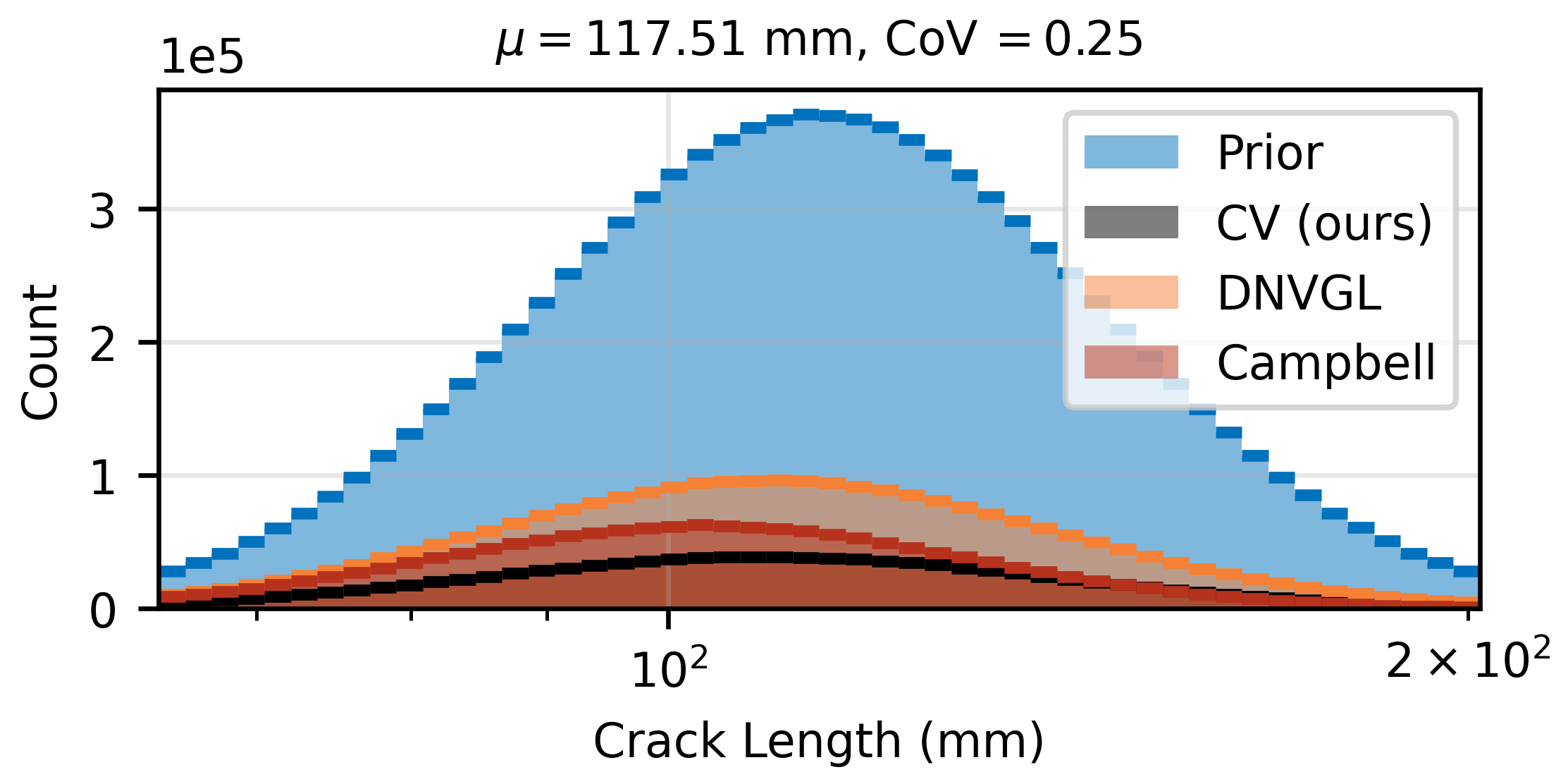}
    \caption{}
    \label{fig:Histograms:mu117.51cov0.25_hist}
\end{subfigure}
\begin{subfigure}{0.49\textwidth}
    \includegraphics[width=\textwidth]{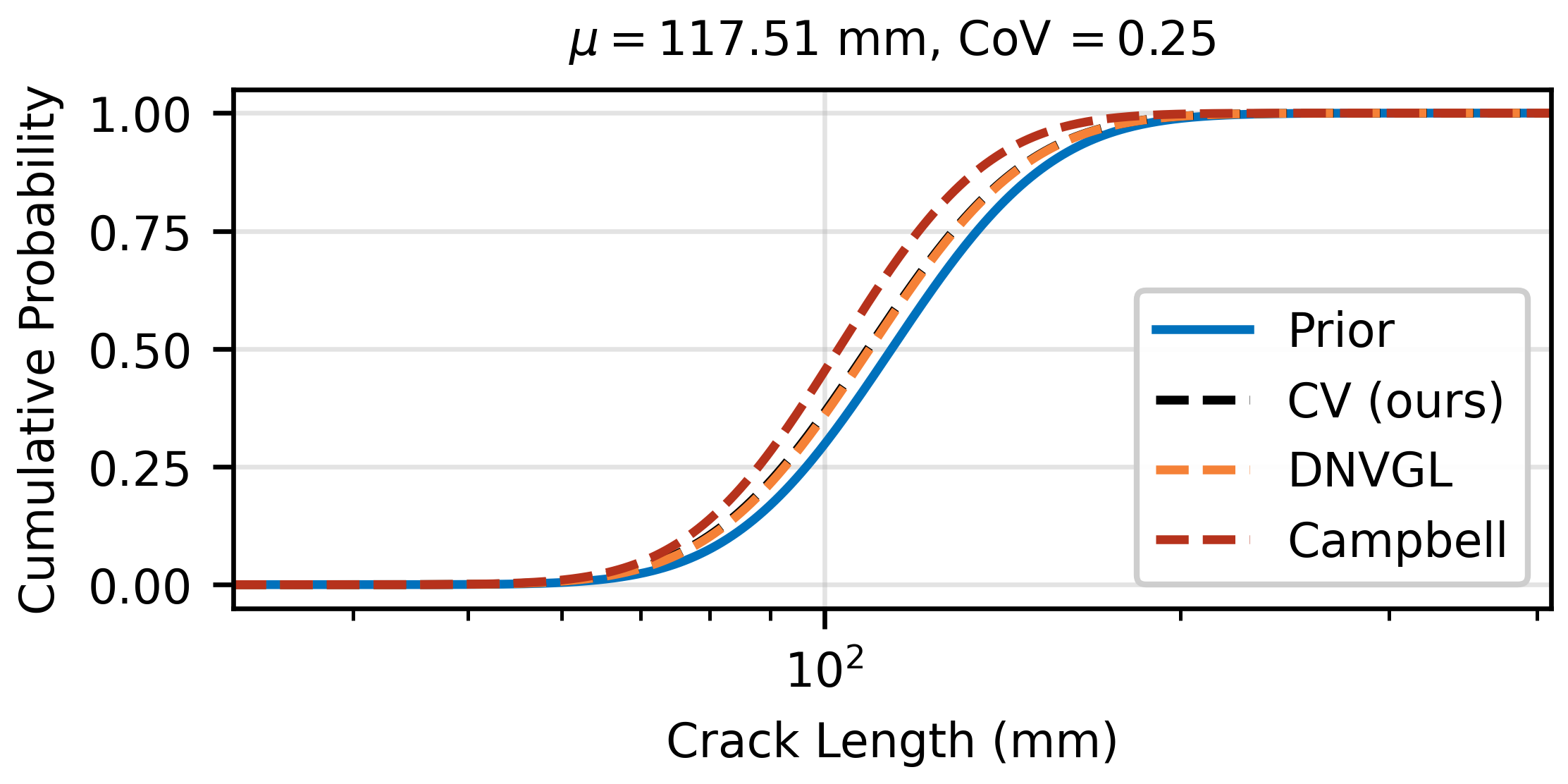}
    \caption{}
    \label{fig:Histograms:mu117.51cov0.25_cdf}
\end{subfigure}
\caption{Histogram and CDF for prior crack-length distribution with $\mu = 117.51$ mm and $\CoV = 0.25$.}
\label{fig:Histograms_mu_117.51_cov_0.25}
\end{figure*}

\begin{figure*}
\centering
\begin{subfigure}{0.49\textwidth}
    \includegraphics[width=\textwidth]{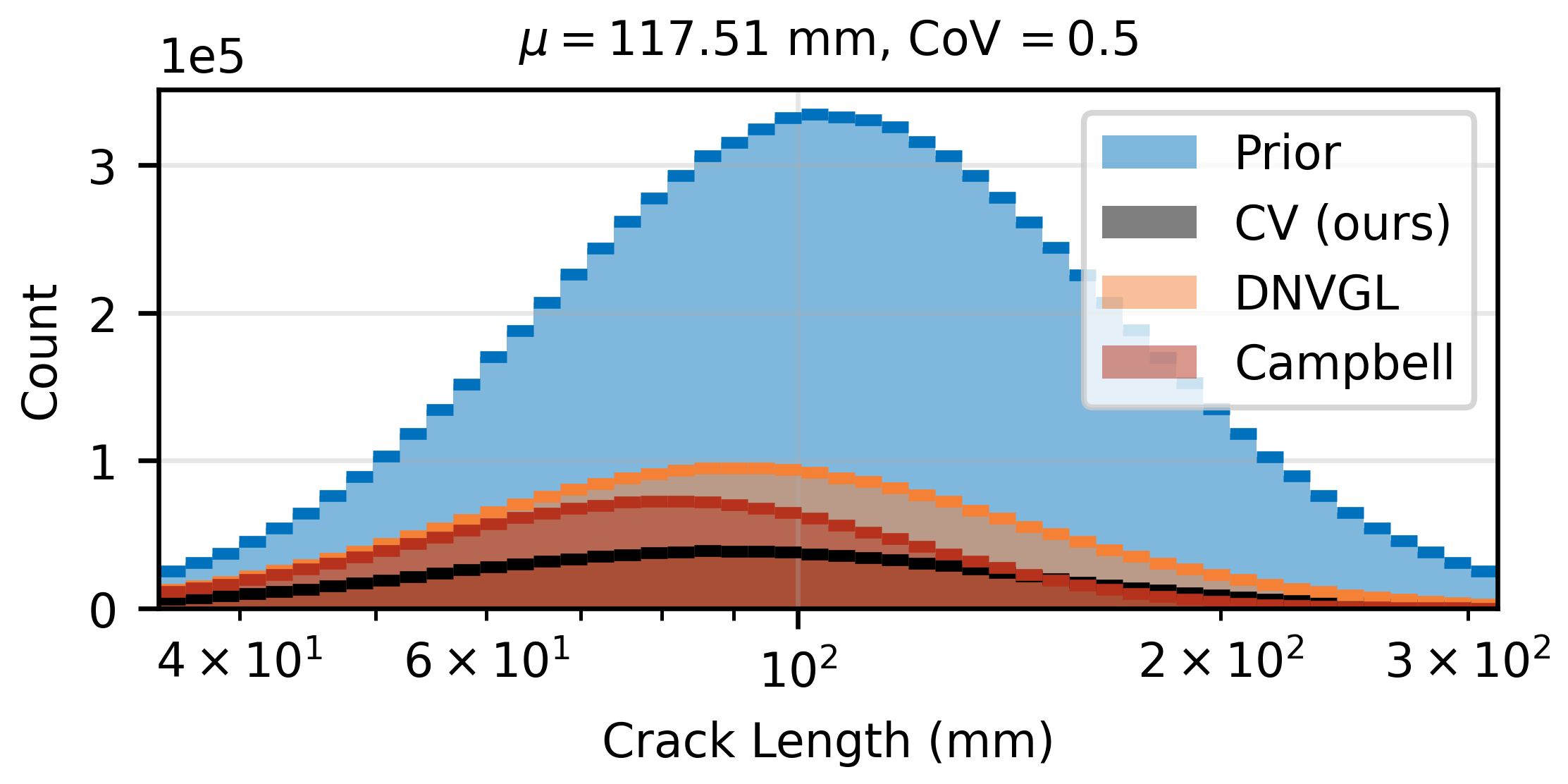}
    \caption{}
    \label{fig:Histograms:mu117.51cov0.5_hist}
\end{subfigure}
\begin{subfigure}{0.49\textwidth}
    \includegraphics[width=\textwidth]{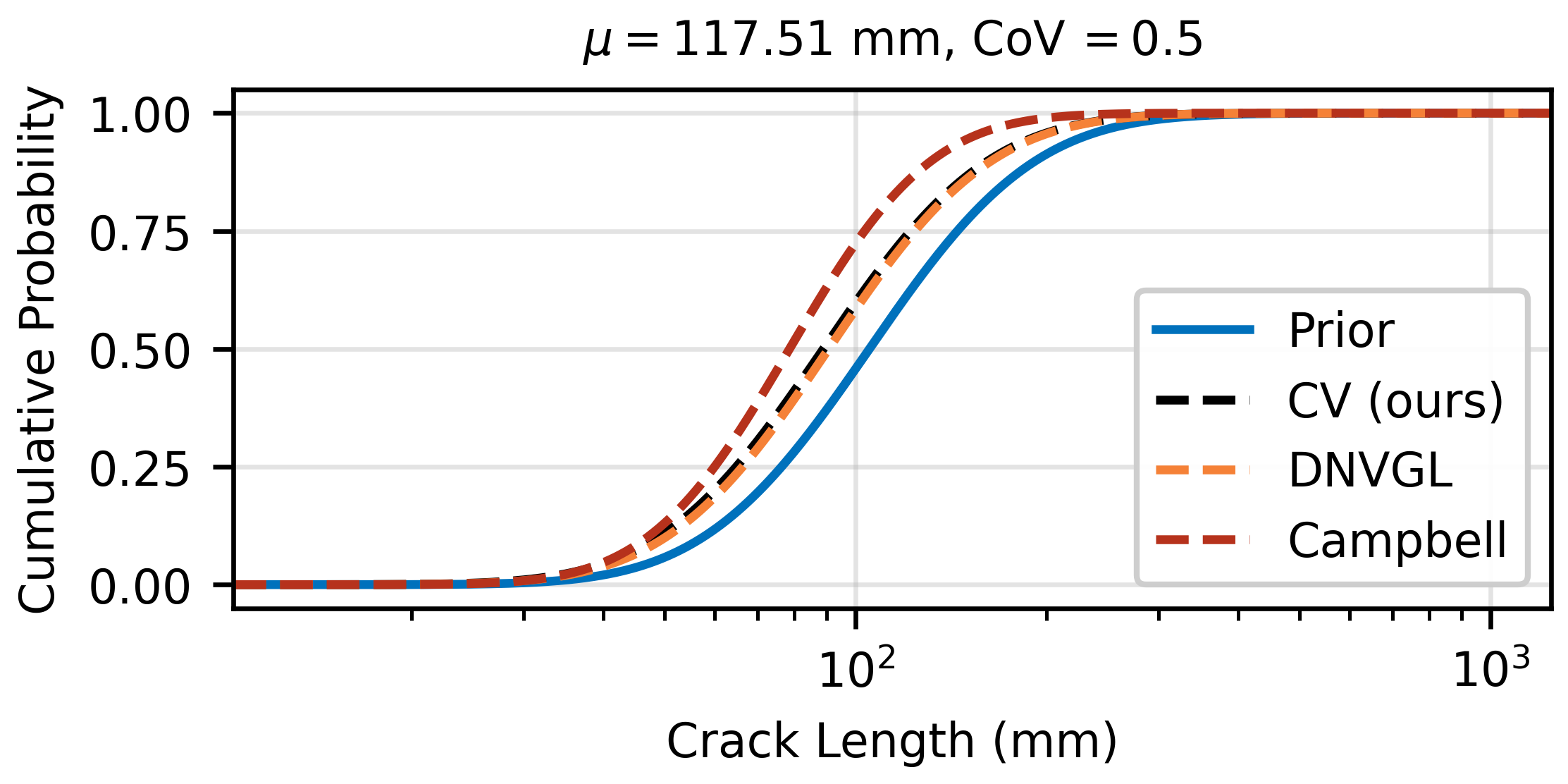}
    \caption{}
    \label{fig:Histograms:mu117.51cov0.5_cdf}
\end{subfigure}
\caption{Histogram and CDF for prior crack-length distribution with $\mu = 117.51$ mm and $\CoV = 0.5$.}
\label{fig:Histograms_mu_117.51_cov_0.5}
\end{figure*}

\begin{figure*}
\centering
\begin{subfigure}{0.49\textwidth}
    \includegraphics[width=\textwidth]{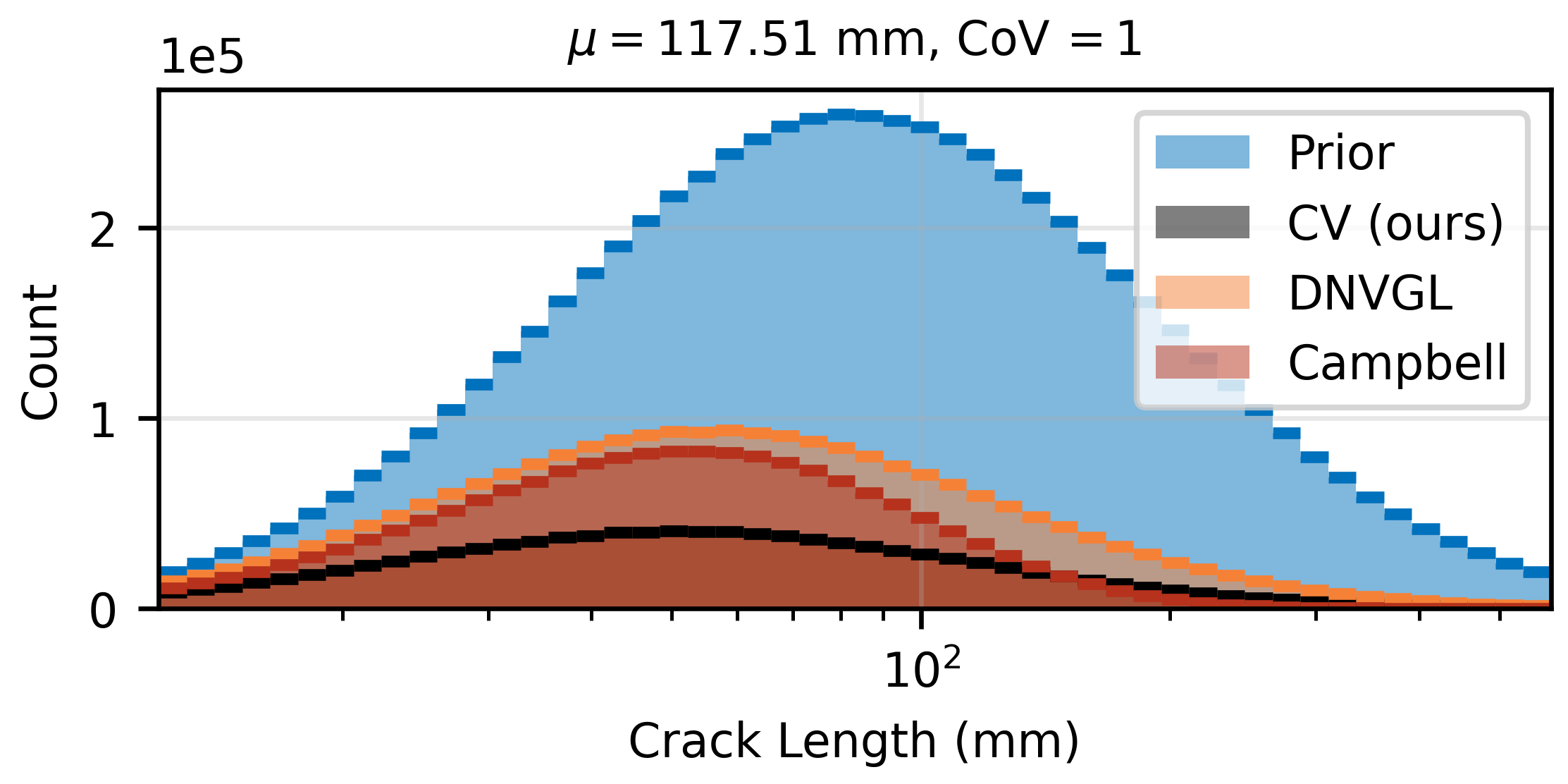}
    \caption{}
    \label{fig:Histograms:mu117.51cov1_hist}
\end{subfigure}
\begin{subfigure}{0.49\textwidth}
    \includegraphics[width=\textwidth]{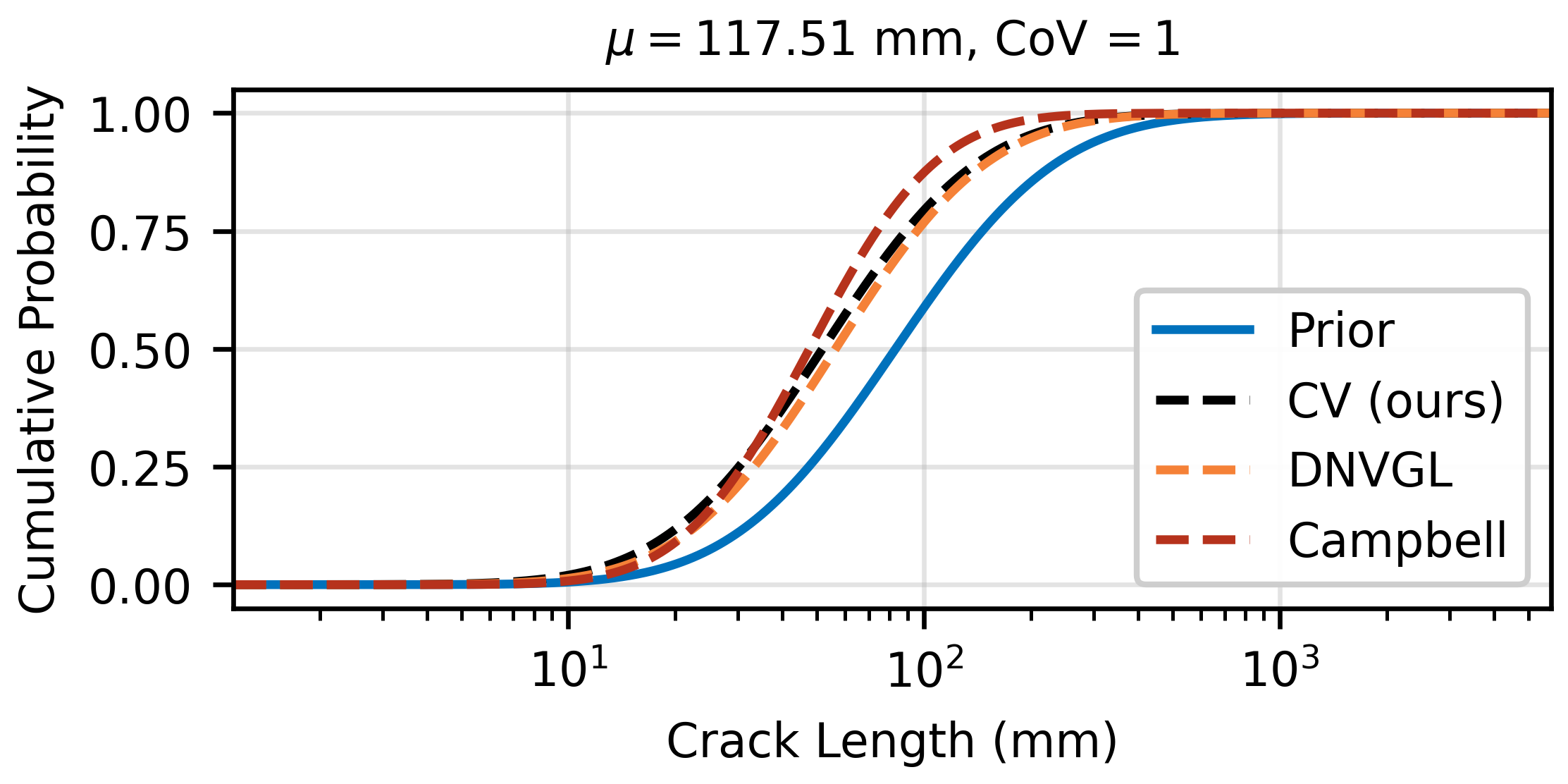}
    \caption{}
    \label{fig:Histograms:mu117.51cov1_cdf}
\end{subfigure}
\caption{Histogram and CDF for prior crack-length distribution with $\mu = 117.51$ mm and $\CoV = 1$.}
\label{fig:Histograms_mu_117.51_cov_1}
\end{figure*}

\begin{figure*}
\centering
\begin{subfigure}{0.49\textwidth}
    \includegraphics[width=\textwidth]{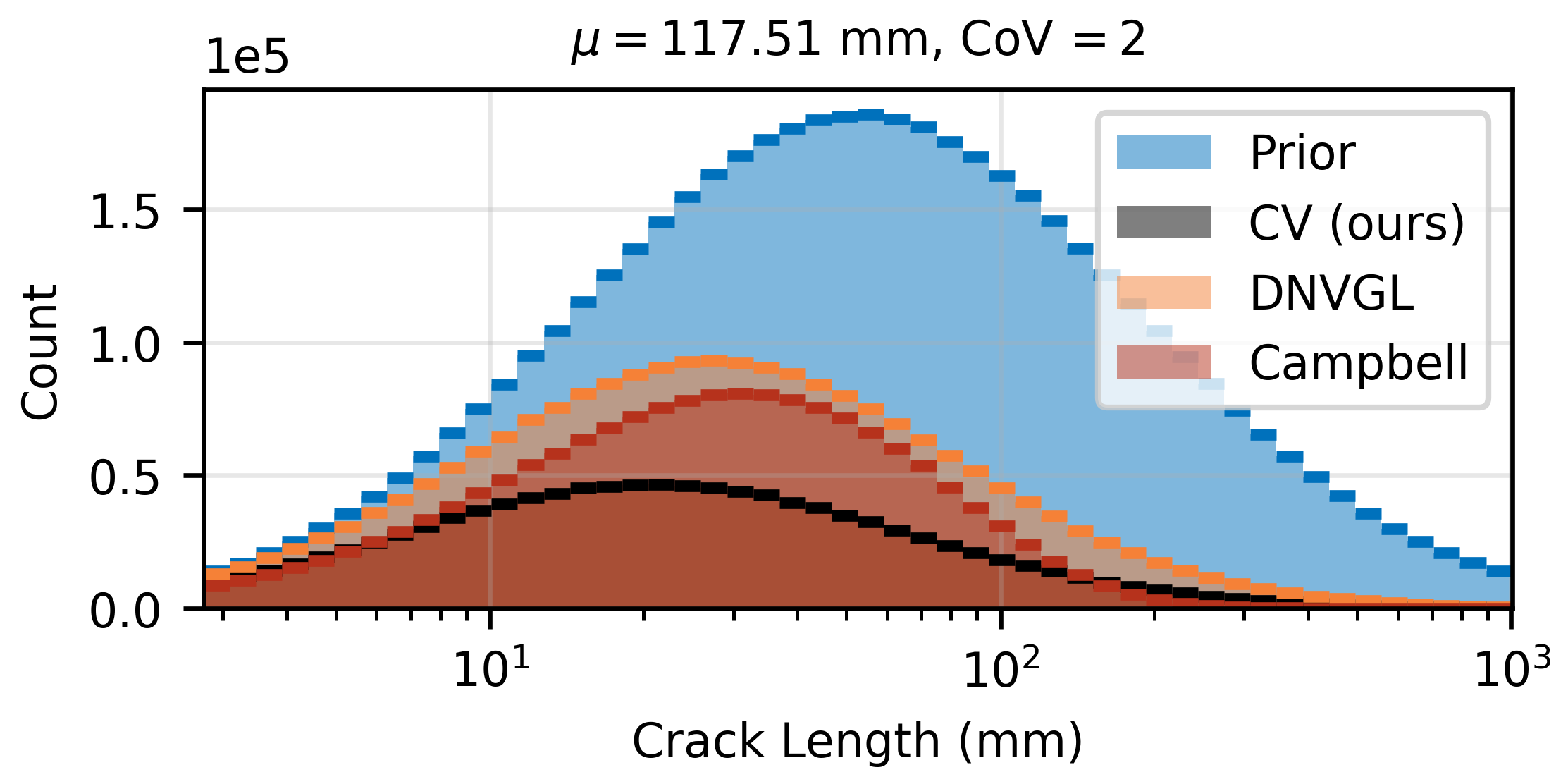}
    \caption{}
    \label{fig:Histograms:mu117.51cov2_hist}
\end{subfigure}
\begin{subfigure}{0.49\textwidth}
    \includegraphics[width=\textwidth]{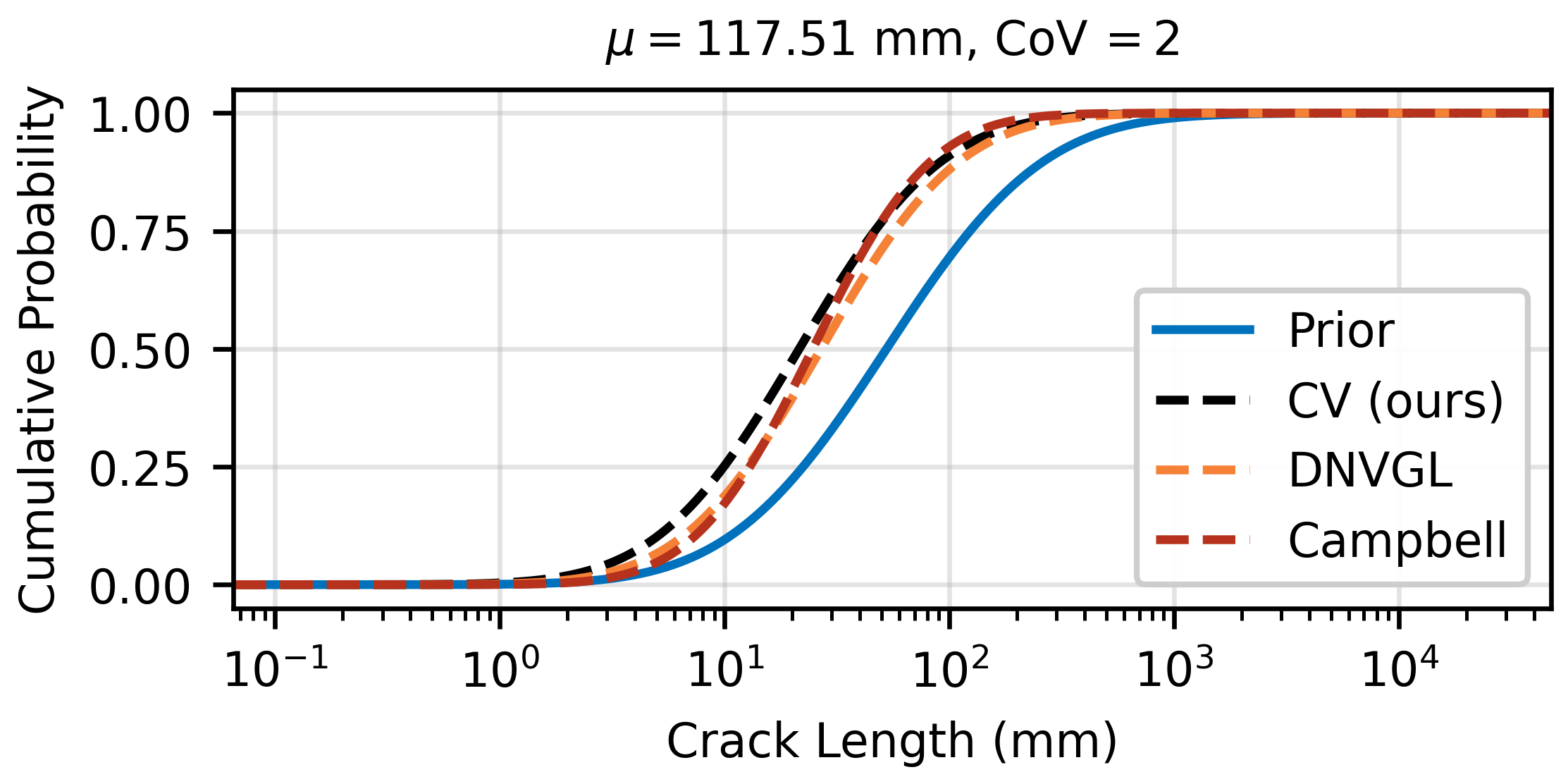}
    \caption{}
    \label{fig:Histograms:mu117.51cov2_cdf}
\end{subfigure}
\caption{Histogram and CDF for prior crack-length distribution with $\mu = 117.51$ mm and $\CoV = 2$.}
\label{fig:Histograms_mu_117.51_cov_2}
\end{figure*}

\begin{figure*}
\centering
\begin{subfigure}{0.49\textwidth}
    \includegraphics[width=\textwidth]{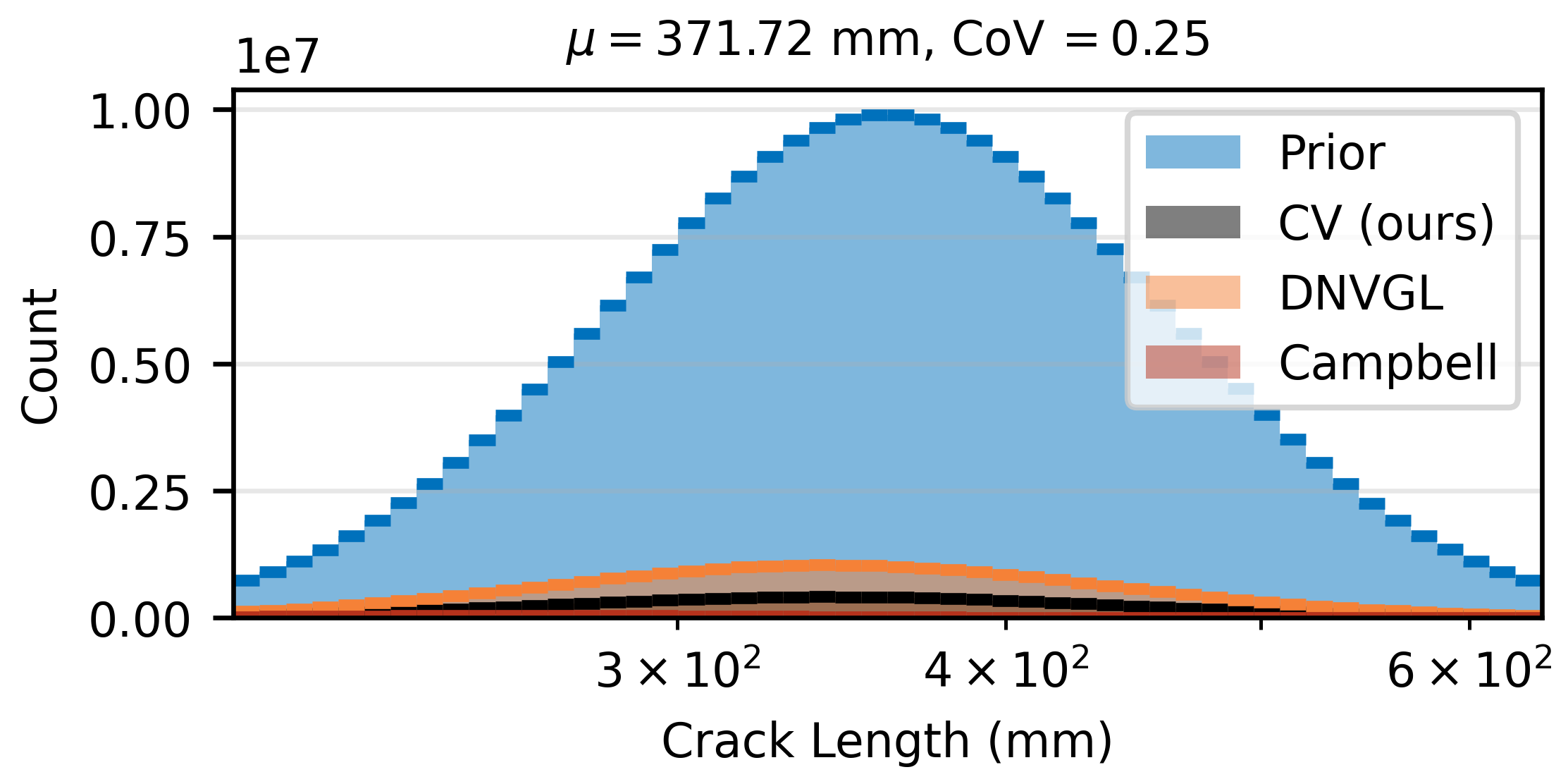}
    \caption{}
    \label{fig:Histograms:mu371.72cov0.25_hist}
\end{subfigure}
\begin{subfigure}{0.49\textwidth}
    \includegraphics[width=\textwidth]{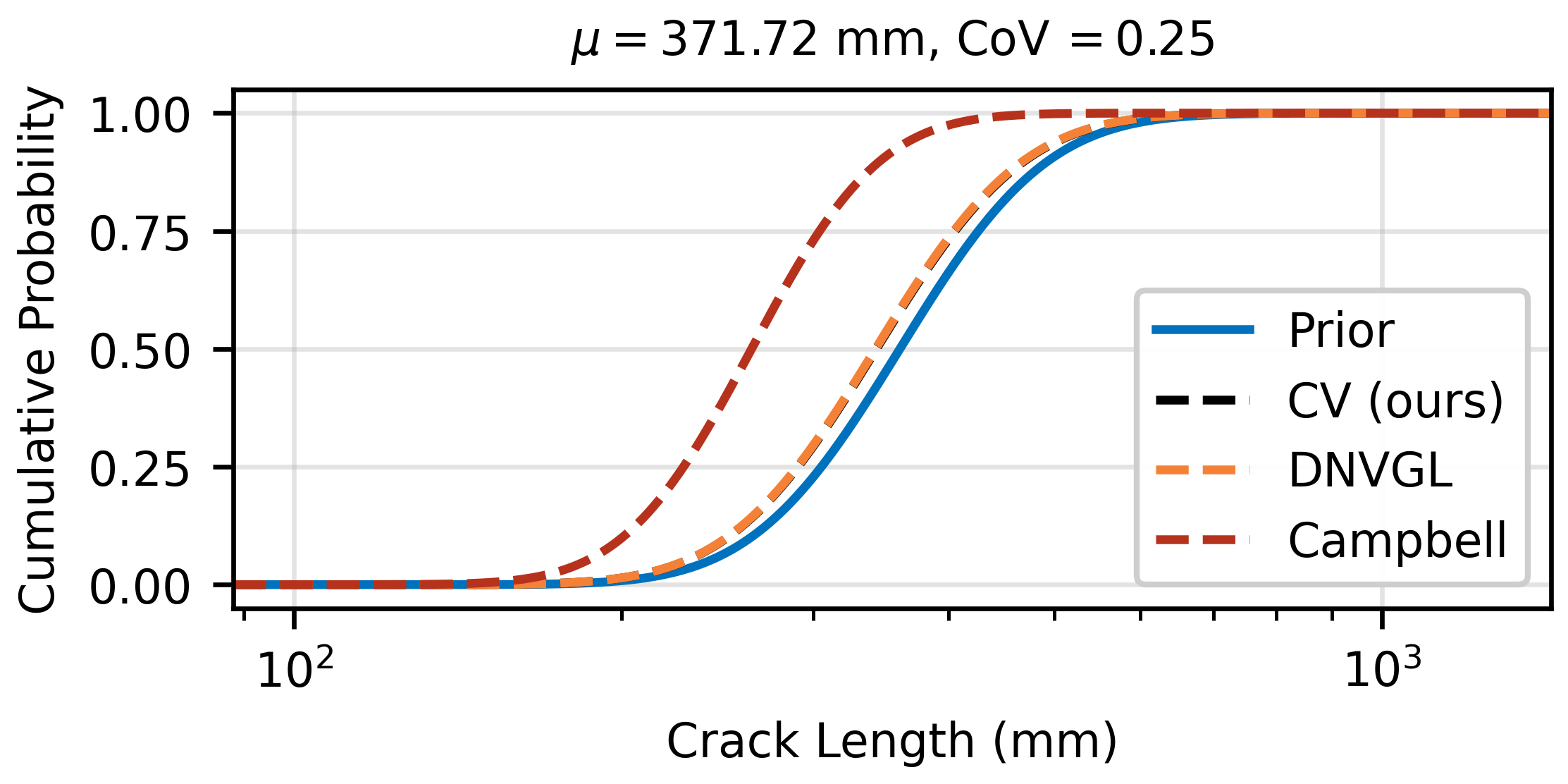}
    \caption{}
    \label{fig:Histograms:mu371.72cov0.25_cdf}
\end{subfigure}
\caption{Histogram and CDF for prior crack-length distribution with $\mu = 371.72$ mm and $\CoV = 0.25$.}
\label{fig:Histograms_mu_371.72_cov_0.25}
\end{figure*}

\begin{figure*}
\centering
\begin{subfigure}{0.49\textwidth}
    \includegraphics[width=\textwidth]{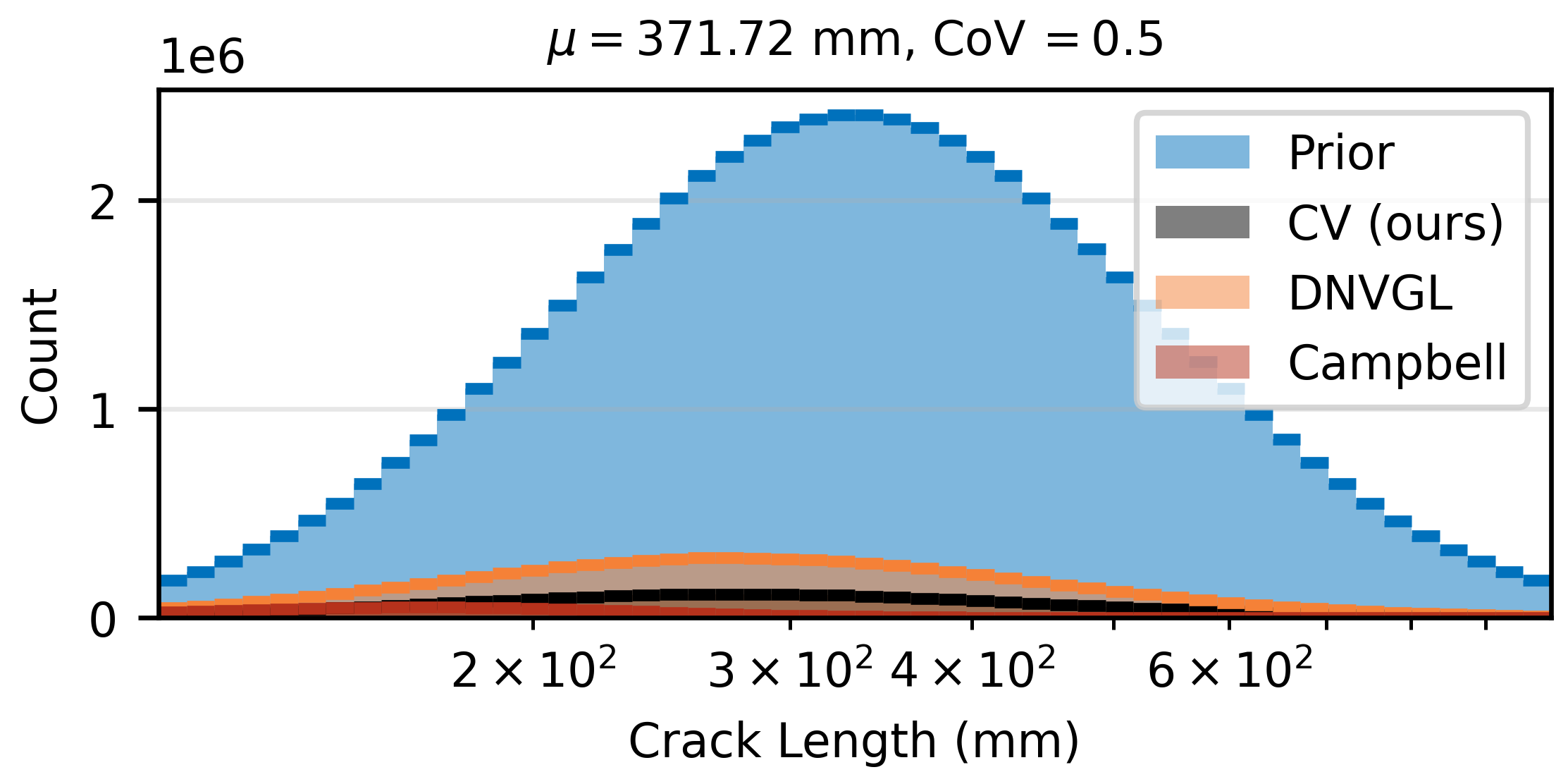}
    \caption{}
    \label{fig:Histograms:mu371.72cov0.5_hist}
\end{subfigure}
\begin{subfigure}{0.49\textwidth}
    \includegraphics[width=\textwidth]{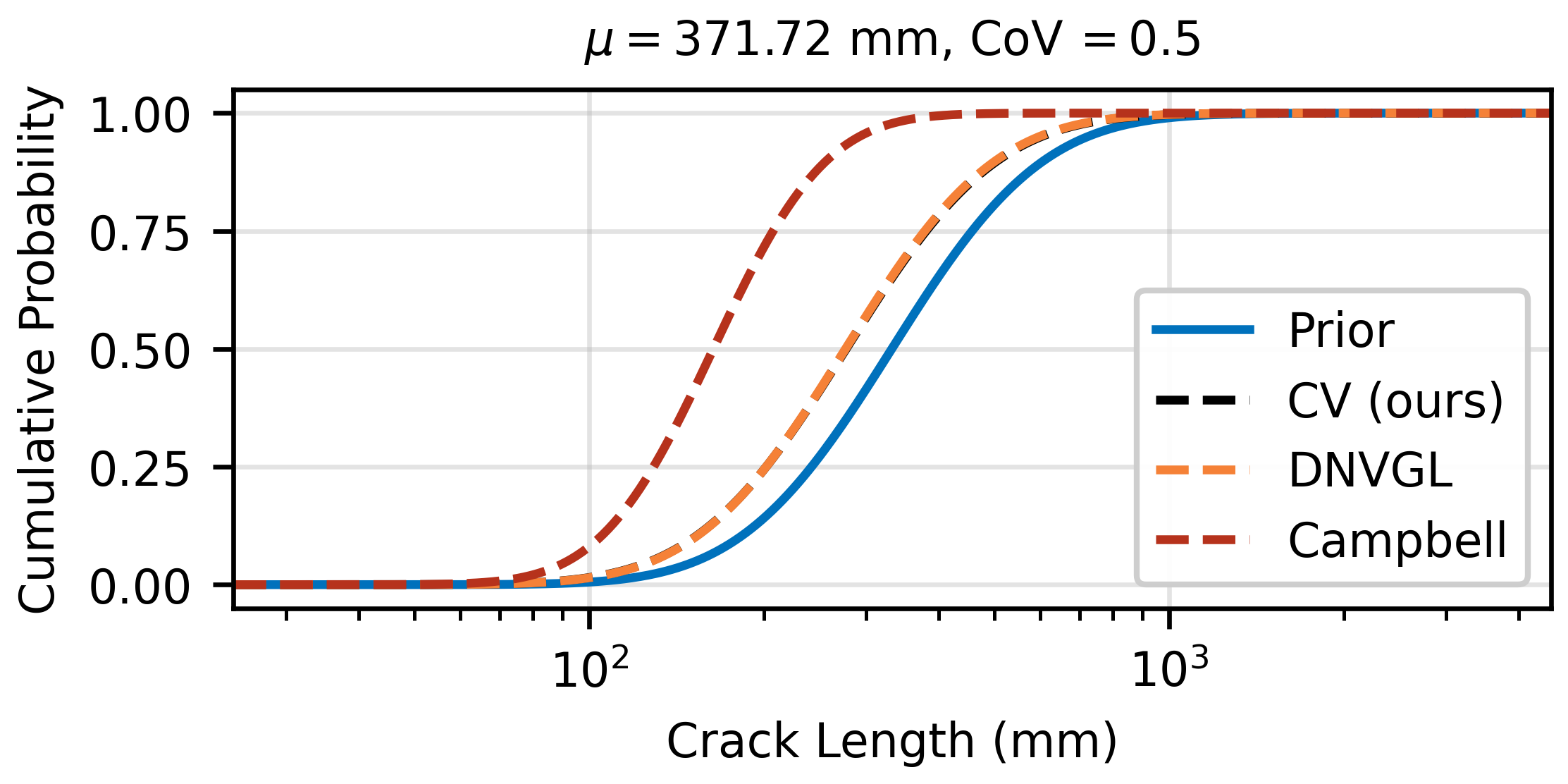}
    \caption{}
    \label{fig:Histograms:mu371.72cov0.5_cdf}
\end{subfigure}
\caption{Histogram and CDF for prior crack-length distribution with $\mu = 371.72$ mm and $\CoV = 0.5$.}
\label{fig:Histograms_mu_371.72_cov_0.5}
\end{figure*}

\begin{figure*}
\centering
\begin{subfigure}{0.49\textwidth}
    \includegraphics[width=\textwidth]{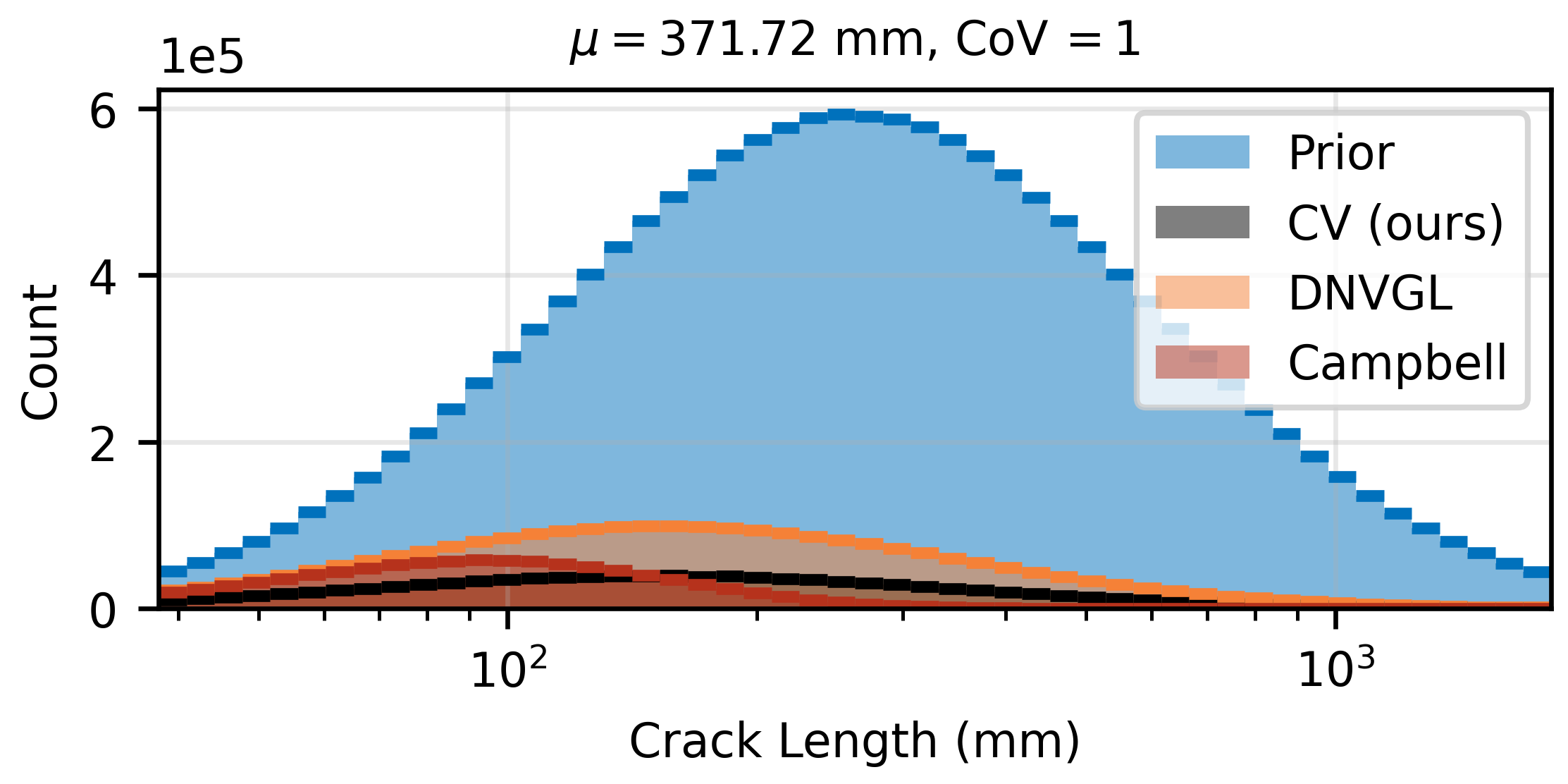}
    \caption{}
    \label{fig:Histograms:mu371.72cov1_hist}
\end{subfigure}
\begin{subfigure}{0.49\textwidth}
    \includegraphics[width=\textwidth]{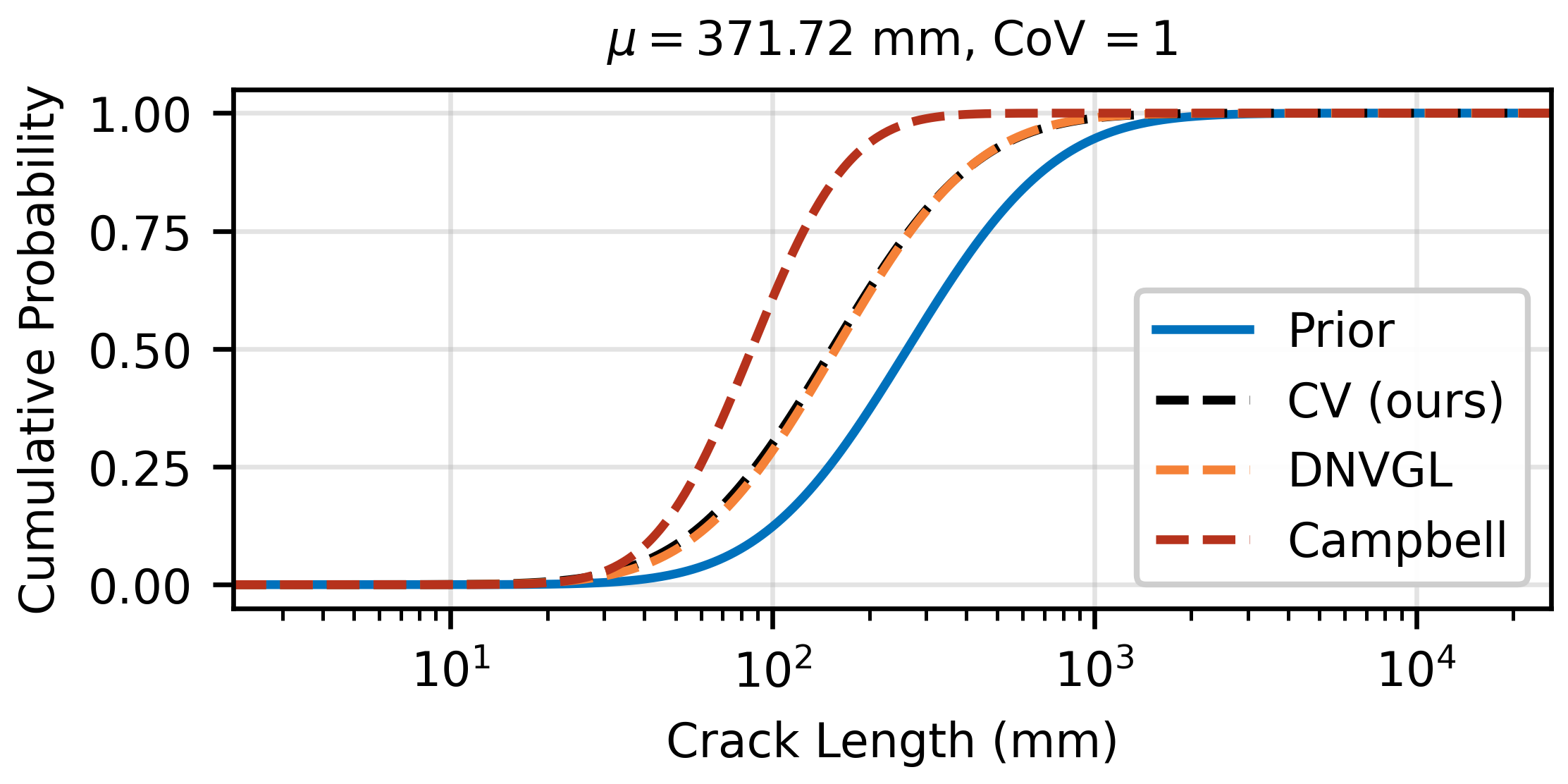}
    \caption{}
    \label{fig:Histograms:mu371.72cov1_cdf}
\end{subfigure}
\caption{Histogram and CDF for prior crack-length distribution with $\mu = 371.72$ mm and $\CoV = 1$.}
\label{fig:Histograms_mu_371.72_cov_1}
\end{figure*}

\begin{figure*}
\centering
\begin{subfigure}{0.49\textwidth}
    \includegraphics[width=\textwidth]{Figures/Update_Crack_Length/update_crack_length_histogram_371.72_0.25.png}
    \caption{}
    \label{fig:Histograms:mu371.72cov2_hist}
\end{subfigure}
\begin{subfigure}{0.49\textwidth}
    \includegraphics[width=\textwidth]{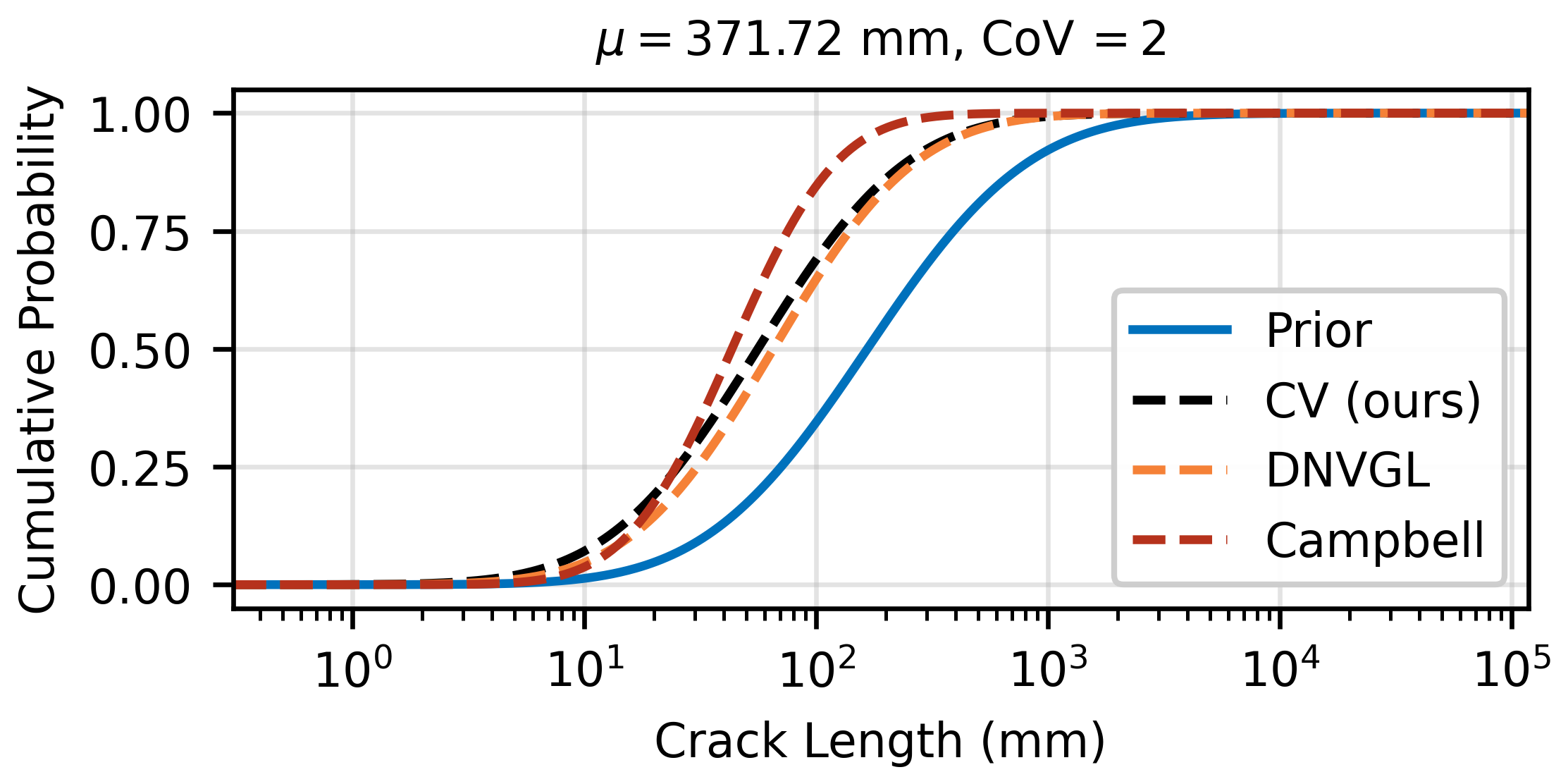}
    \caption{}
    \label{fig:Histograms:mu371.72cov2_cdf}
\end{subfigure}
\caption{Histogram and CDF for prior crack-length distribution with $\mu = 371.72$ mm and $\CoV = 2$.}
\label{fig:Histograms_mu_371.72_cov_2}
\end{figure*}

\end{document}